\documentclass[preprint,12pt]{elsarticle}
\usepackage[top=1in,bottom=1in,left=1in,right=1in]{geometry}
\usepackage{amssymb}
\usepackage{amsmath}
\usepackage{amsbsy}
\usepackage{mathtools}
\usepackage{xcolor}
\usepackage{subfigure}

\biboptions{sort&compress}

\begin{document}

\begin{frontmatter}

\title{Uncertainty quantification for noisy inputs-outputs in physics-informed neural networks and neural operators}

\author[brown]{Zongren Zou}
\author[hust]{Xuhui Meng \fnref{1}}
\author[brown,eng]{George Em Karniadakis}

\fntext[1]{Corresponding author: xuhui\_meng@hust.edu.cn (Xuhui Meng).}
\address[brown]{Division of Applied Mathematics, Brown University, Providence, RI 02906, USA}
\address[hust]{Institute of Interdisciplinary Research for Mathematics and Applied Science, School of Mathematics and Statistics, Huazhong University of Science and Technology, Wuhan 430074, China}
\address[eng]{School of Engineering, Brown University, Providence, RI 02906, USA}

\begin{abstract}
Uncertainty quantification (UQ) in scientific machine learning (SciML) becomes increasingly critical as neural networks (NNs) are being widely adopted in addressing complex problems across various scientific disciplines.
Representative SciML models are physics-informed neural networks (PINNs) and neural operators (NOs).
While UQ in SciML has been increasingly investigated in recent years, very few works have focused on addressing the uncertainty caused by the noisy inputs, such as spatial-temporal coordinates in PINNs and input functions in NOs. The presence of noise in the inputs of the models can pose significantly more challenges compared to noise in the outputs of the models, primarily due to the inherent nonlinearity of most SciML  algorithms.
As a result, UQ for noisy inputs becomes a crucial factor for reliable and trustworthy deployment of these models in applications involving physical knowledge.
To this end, we introduce a Bayesian approach to quantify uncertainty arising from noisy inputs-outputs in PINNs and NOs.
We show that this approach can be seamlessly integrated into  PINNs and NOs, when they are employed to encode the physical information. 
PINNs incorporate physics by including physics-informed terms via automatic differentiation, either in the loss function or the likelihood, and often take as input the spatial-temporal coordinate. Therefore, the present method equips PINNs with the capability to address problems where the observed coordinate is subject to noise. 
On the other hand, pretrained NOs are also commonly employed as equation-free surrogates in solving differential equations and Bayesian inverse problems, in which they take functions as inputs. The proposed approach enables them to handle noisy measurements for both input and output functions with UQ. 
We present a series of numerical examples to demonstrate the consequences of ignoring the noise in the inputs and the effectiveness of our approach in addressing noisy inputs-outputs with UQ when PINNs and pretrained NOs are employed for physics-informed learning.
\end{abstract}


\end{frontmatter}

\newpage

\section{Introduction}\label{sec:1}

Recent years have witnessed great progress in scientific machine learning (SciML) in which the data and physics  are combined seamlessly under modern machine learning (ML) frameworks to address complex scientific modeling, computation and model discovery \cite{karniadakis2021physics}. 
As SciML is emerging as a revolutionary paradigm, its uncertainty quantification (UQ) becomes increasingly critical for trustworthy and reliable predictions. 
Numerous UQ approaches have been proposed to tackle diverse sources of uncertainty in SciML \cite{psaros2023uncertainty, zou2022neuraluq, yang2022scalable, lin2021accelerated, moya2023deeponet}. For instance Bayesian neural networks (BNNs) were employed to quantify uncertainty arising from noisy data \cite{yang2021b, zou2022neuraluq, linka2022bayesian} and physical models \cite{zou2023correcting} when solving ordinary/partial differential equations (ODEs/PDEs). Non-Bayesian methods, as reported in \cite{psaros2023uncertainty, zou2022neuraluq, yang2022multi, zhang2023discovering}, have also been used to solve such problems. In \cite{zhang2019quantifying, yang2020physics}, uncertainty stemming from solving stochastic differential equations was investigated, while in \cite{meng2022learning, zou2023hydra, YIN2023105424, winovich2019convpde}, researchers employed the deep generative models to quantify the uncertainties in solving ODEs/PDEs. A comprehensive review and a Python library for UQ in SciML can be found in \cite{psaros2023uncertainty} and \cite{zou2022neuraluq}, respectively. 

In the present study, our specific interest is to explore the uncertainties in the predictions of two popular scientific machine learning models, e.g. the physics-informed neural networks (PINNs) \cite{raissi2019physics} and neural operators (NOs) \cite{lu2021learning,li2020fourier}. The former encodes the physics in the form of differential equations via automatic differentiation, while the latter is capable of learning the solution operator or hidden physics given training data. Generally, PINNs take as input the spatial-temporal coordinates and outputs functions that approximate the solutions to PDEs, while NOs take the boundary conditions and/or source terms, etc. as inputs and outputs functions, which approximate the solutions to differential equations. Despite the plurality of aforementioned UQ methods in SciML, we note that most of the existing works assume that the inputs, e.g. spatial-temporal coordinates in PINNs, are clean and only take the uncertainties associated with the noisy measurements on the PINN outputs into consideration. However, in real-world applications the measurements on both the coordinates and the corresponding responses may be noisy if they are from sensors, e.g., particle image velocimetry (PIV) \cite{cai2019particle, cai2019dense}, estimating biomechanical properties of the aorta \cite{YIN2023105424}, hydrological modeling \cite{kavetski2006bayesian1, kavetski2006bayesian2}. As reported in \cite{dellaportas1995bayesian, fan1993nonparametric, gleser1981estimation, wright1999bayesian, tresp1993training, van2000learning, williams2006gaussian, atkinson1998statistical, schennach2016recent, girard2002gaussian, mchutchon2011gaussian, de2023convergence, patel2022error}, the measurement error in the inputs of models can present considerably greater challenges compared to noise in the outputs, primarily attributed to the inherent model nonlinearity. To the best of our knowledge,  quantifying uncertainty originating from noisy inputs of PINNs as well as NOs has  been largely ignored.   Specifically, we describe the problem of interest as follows \cite{dellaportas1995bayesian}:
\begin{subequations}\label{eq:noisy_model}
\begin{align}
    \tilde{y} &= \mathcal{H}(\chi) + \epsilon_o, \label{eq:noisy_output}\\
    \tilde{\chi} &= \chi + \epsilon_{in} \label{eq:noisy_input},
\end{align}
\end{subequations}
where $\mathcal{H}$ denotes an NN surrogate model, $\chi$ denotes the input, and $\tilde{y}$ and $\tilde{\chi}$ denote the observed target value of the output and input of $\mathcal{H}$, respectively. 
We denote the measurement error or noise as $\epsilon_o$ and $\epsilon_{in}$ for the output and input, respectively, making the regression of $\chi, \mathcal{H}$ a noisy input-output problem.

Studies on uncertainty arising from both noisy inputs and outputs can be traced back to the 1980s and 1990s within statistical research \cite{dellaportas1995bayesian, fan1993nonparametric, gleser1981estimation} and ML communities \cite{wright1999bayesian, tresp1993training, van2000learning}, respectively. Regression models developed to address this challenge are commonly referred to as \textit{errors-in-variables} or \textit{measurement error} models \cite{williams2006gaussian, atkinson1998statistical, schennach2016recent}. For example, Gaussian processes (GPs), which are a popular machine learning algorithm for regression problems, have been used in modeling uncertainty arising from noisy inputs as well as the outputs \cite{girard2002gaussian, mchutchon2011gaussian}.
However, the aforementioned work on UQ for noisy inputs-outputs has primarily focused on function approximation problems using NNs and GPs, without the involvement of physics information, e.g. ODEs/PDEs.
More recently, the effect of noisy inputs in operator learning has been studied, e.g., in \cite{de2023convergence} convergence rates were analyzed for learning linear operators with noisy inputs-outputs data and in \cite{patel2022error} an errors-in-variables modeling was adopted for operator learning. 
Nevertheless, they focused on analyzing and/or mitigating the effects of noisy inputs in NOs. How to quantify the uncertainties caused by both  noisy inputs and outputs on the SciML models, especially PINNs and NOs, remains unclear.

In this work, we present a Bayesian approach for quantifying uncertainty induced by noisy inputs-outputs in PINNs as well as NOs. Specifically, we assume that we have measurement errors or noise for (1) the spatial-temporal coordinates (i.e., inputs) as well as the output functions (i.e., outputs) in PINNs, and (2) the input and output functions in NOs, as illustrated in Fig. \ref{fig:1}. We like to quantify the uncertainties arising from both noisy inputs and outputs in PINNs and NOs with the proposed method. We also point out that in the current study we treat NOs as pretrained surrogates models and use them for certain downstream tasks (see Sec.~\ref{sec:2} for the task setup). 
Hence, the NOs are deterministic models, which is essentially different from training NOs with UQ as in \cite{yang2022scalable, lin2021accelerated, garg2022variational}.
This employment is in fact one of the major functionalities of NOs in the literature. To name a few examples, in \cite{li2020fourier, li2021physics}, Bayesian inverse problems were solved with pretrained FNOs where data of the solution are available; in \cite{meng2022learning, zou2022neuraluq, psaros2023uncertainty}, a combination of NNs and pretrained DeepONets was employed for mixed problems with partial and noisy data. 

The rest of this paper is organized as follows. In Sec.~\ref{sec:2}, we introduce the problem formulation, PINNs and NOs, and discuss how PINNs and NOs are employed to encode the physics information. In Sec.~\ref{sec:3}, we present the Bayesian framework for quantifying uncertainty arising from noisy inputs-outputs under the frameworks of PINNs and NOs. In Sec.~\ref{sec:4}, we conduct four computational examples to demonstrate the consequences of ignoring the noise in the inputs and the effectiveness of our approach in addressing noisy inputs-outputs with UQ. We summarize our work in Sec.~\ref{sec:5}.

\section{Encoding physics in scientific machine learning}\label{sec:2}
\begin{figure}[ht!]
    \centering
    \includegraphics[scale=.65]{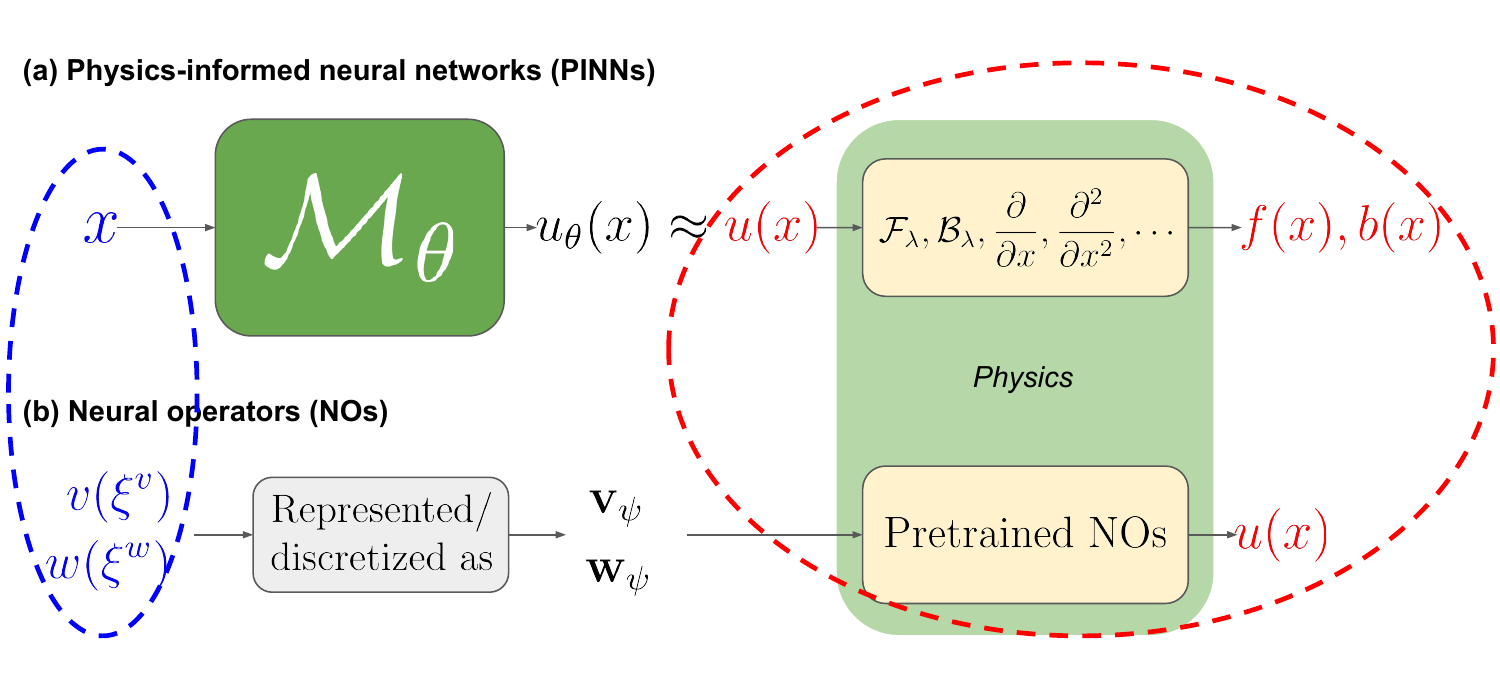}
    \caption{A schematic view of two prevalent ways of encoding physics in SciML. In the upper plot we present the framework of physics-informed neural networks (PINNs) \cite{raissi2019physics}, where the physics is encoded by modeling $u(x)$ using an NN, parameterized by $\theta$ and denoted by $\mathcal{M}_\theta$, and approximating $f, b$ via automatic differentiation and the ODE/PDE. In the lower plot we represent the physics by pretrained neural operators (NOs), which are trained offline to learn the solution operators of \eqref{eq:problem}, often based on clean and simulated data. The pretrained NOs effectively encode the underlying physics within the NN architecture and parameters \cite{kovachki2021neural}, while in the downstream tasks such as solving ODEs/PDEs and tackling hybrid problems, they remain fixed and serve as equation-free surrogates. 
    Here $\mathbf{v}_\psi$ and $\mathbf{w}_\psi$ denote discretizations of the input functions $v(\xi^v)$ and $w(\xi^w)$, respectively, which are parameterized by $\psi$.
    Note that in these two methods, the inputs and outputs (color-coded by {\color{blue}blue} and {\color{red}red}, respectively) differ, making it crucial to develop uncertainty quantification (UQ) approaches to handle noisy inputs-outputs in both scenarios.}
    \label{fig:1}
\end{figure}

Consider the following ODE/PDE:
\begin{subequations}\label{eq:problem}
\begin{align}
    \begin{split}
        \mathcal{F}_\lambda [u](x) &= f(x), x\in\Omega,
    \end{split}\\
    \begin{split}
        \mathcal{B}_\lambda [u](x) &= b(x), x\in\partial\Omega,
    \end{split}
\end{align}
\end{subequations}
where $\Omega$ denotes the domain, $\mathcal{F}$ the general differential operator, $\mathcal{B}$ the general boundary operator, $\lambda$ the parameter of the ODE/PDE, $u$, $f$ and $b$ the sought solution, source term and initial/boundary term, respectively, and $x$ the spatial-temporal coordinate. 
In this work we are interested in solving a class of problems using SciML, which is referred to as the \textit{hybrid} problem, i.e., inferring $u$, $f$, $b$, and/or $\lambda$ given data (denoted as $\mathcal{D}$) and the underlying physics.
Note that if $\lambda$ is known and $u$ is not, the hybrid problem is equivalent to solving a forward ODE/PDE problem.

There are in general two common ways to encode physics in solving ODEs/PDEs problem in SciML: taking into consideration the physics-informed losses as in PINNs \cite{raissi2019physics, yang2021b, tang2023adversarial, tang2023pinns} and other SciML models \cite{sirignano2018dgm, chen2021physics, chen2023leveraging, chen2023leveraging2}, which are often realized via automatic differentiation, and using pretrained NOs that represent the physics, e.g. DeepONets \cite{lu2021learning} and FNOs \cite{li2020fourier}. A schematic view of these two prevalent approaches is presented in Fig.~\ref{fig:1}.
As shown, the inputs of PINNs are spatial-temporal coordinates while they are functions in NOs.
We briefly review the PINNs as well as NOs in what follows, and interested readers are directed to \cite{raissi2019physics, lu2021deepxde} and \cite{lu2021learning, li2020fourier, wang2021learning, li2021physics} for more details of PINNs and neural operators, respectively. 

\subsection{Physics-informed neural networks (PINNs)}

The PINN method, originally proposed in \cite{raissi2019physics}, addresses the hybrid problem by modeling $u(x)$ (and $\lambda$ if $\lambda$ is unknown) with an NN parameterized by $\theta$, denoted as $u_\theta(x)$ (and $\lambda_\theta(x)$ if $\lambda$ is unknown), and then modeling $f(x)$ and $b(x)$ with $\mathcal{F}_{\lambda}[u_\theta](x)$ and $\mathcal{B}_{\lambda}[u_\theta](x)$ via automatic differentiation, respectively. 
The physics is then explicitly encoded in the training process by adding physics-informed terms involving $\mathcal{F}_{\lambda}[u_\theta](x)$ and $\mathcal{B}_{\lambda}[u_\theta](x)$. The optimal $\theta$ can be obtained by minimizing the following loss function:
\begin{equation}\label{eq:loss_PINN}
\begin{split}
    \mathcal{L}(\theta) = \frac{w_u}{N_u}\sum_{i=1}^{N_u} ||u_\theta(x^u_i) - u_i||^2 + \mathcal{L}_{PDE} + \frac{w_f}{N_f}\sum_{i=1}^{N_u} ||\mathcal{F}[u_\theta](x^f_i) - f_i||^2\\
    + \frac{w_b}{N_b}\sum_{i=1}^{N_b} ||\mathcal{B}[u_\theta](x^b_i) - b_i||^2 + \frac{w_\lambda}{N_\lambda}\sum_{i=1}^{N_\lambda} ||\lambda_\theta(x^\lambda_i) - b_i||^2,
\end{split}
\end{equation}
where $w_u, w_f, w_b, w_\lambda$ are belief weights for balancing different terms, and $||\cdot||$ is the $\ell^2$-norm for finite-dimensional vector, and $\{x_i^u, u_i\}_{i=1}^{N_u}$, $\{x_i^f, f_i\}_{i=1}^{N_f}$, $\{x_i^b, b_i\}_{i=1}^{N_b}$, $\{x_i^\lambda, \lambda_i\}_{i=1}^{N_\lambda}$ are observational data.

\subsection{Neural operators (NOs)}

\begin{figure}[ht!]
    \centering
    \includegraphics[scale=.6]{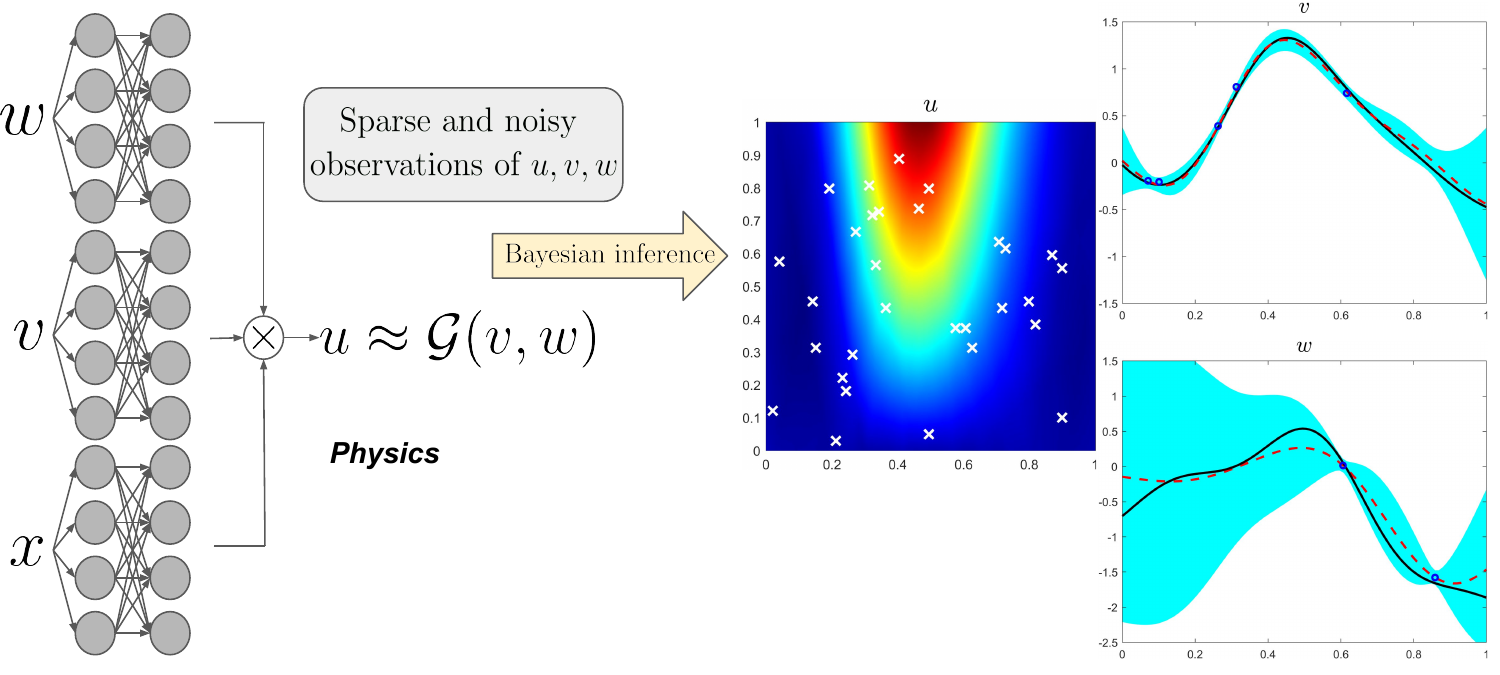}
    \caption{An example of using pretrained NOs to encode physics in the hybrid problem. In particular, we use a multi-input DeepONet \cite{jin2022mionet} to represent the physics which connects $u, v$ and $w$. Here, the multi-input DeepONet, denoted by $\mathcal{G}$, takes as input $w$ and $v$ and outputs $u$, while $\bigotimes$ denotes elementwise multiplication. In the hybrid problem, $u, v, w$ are reconstructed with UQ from their noisy measurements and the physics, represented by the pretrained multi-input DeepONet, enabling synergistic learning, as shown in the predicted mean and uncertainty. See Sec. \ref{sec:example_3} for details.}
    \label{fig:2}
\end{figure}

The NOs are developed as fast solvers to families of ODEs/PDEs.  Moreover, the pretrained NOs can also be employed as surrogates to represent the physics involved in addressing hybrid problems. This capability is regarded as one of the major functions of NOs in the literature \cite{li2020fourier, li2021physics, meng2022learning, psaros2023uncertainty, zou2022neuraluq}. In particular, the solution operator of \eqref{eq:problem} is first approximated with NNs, denoted by $\mathcal{G}$. The corresponding training procedure is often considered as offline and is done with simulated and clean data to ensure the physics is accurately encoded. Then the pretrained NO $\mathcal{G}$ is used as a differentiable equation-agnostic surrogate, which encodes the underlying physical knowledge between the input and output functions (see Sec.~\ref{sec:nos} for details).

In this study, we choose Deep operator networks (DeepONets) \cite{lu2021learning} and Fourier neural operators (FNO) \cite{li2020fourier} as two representative methods for operator learning. 
Although both of them have been further developed to physics-informed versions for higher accuracy and fewer data \cite{wang2021learning, li2021physics}, we intentionally omit the physics-informed learning in this paper for the sake of simplicity.
DeepONet typically consists of two sub-networks: one that takes as input the spatial-temporal coordinate $x$, termed the \textit{trunk net}, and another that takes as input the discretized function $v$, termed the \textit{branch net}. On the other hand, FNO utilizes the fast Fourier transform over the discretization of the input function. For a comprehensive comparison between DeepONets and FNOs, we refer the readers to \cite{lu2022comprehensive}. In addition to the vanilla version, we also consider the multi-input DeepONet \cite{jin2022mionet}, which maps multiple functions to a single function. We employ it as an equation-agnostic surrogate for hybrid problem involving more than two functions. 
An illustration of the architecture of a multi-input DeepONet and its associated hybrid problem is presented in Fig.~\ref{fig:2}; see Sec.~\ref{sec:example_3} for an example.

\section{Uncertainty quantification for noisy inputs-outputs}\label{sec:3}
In this section, we adopt the noisy input-output model formulated in \eqref{eq:noisy_model} in the hybrid problem, and present the methodology of addressing it with UQ.
Recall that we have the following model:
\begin{equation}\label{eq:noisy_model2}
\begin{split}
    \tilde{y} &= \mathcal{H}(\chi) + \epsilon_o,\\
    \tilde{\chi} &= \chi + \epsilon_{in}.
\end{split}
\end{equation}
Note that in the present study, $\mathcal{H}$ is (1) an NN in PINNs whose parameters are unknown, or (2) a pretrained NO whose parameters are known and fixed.  
Interpretations of inputs, outputs and NN model $\mathcal{H}$ are provided in Table \ref{tab:problems}. As shown, when the physics is encoded through PINNs and automatic differentiation, the input is the spatial-temporal coordinate and the outputs are the approximations of the solution $u$, the source term $f$, and/or the boundary term $b$; when the pretrained NOs are used to represent the physics implicitly, the inputs are functions, e.g. model parameter and source/boundary terms, and the output is the approximation of $u$. We note that the physics could be encoded by employing both approaches simultaneously, i.e. using pretrained NOs as surrogates and adding physics-informed terms in the downstream training, as in \cite{li2021physics, wang2021learning}. However, for the sake of simplicity in this study, we do not employ the physics-informed terms in NOs.

\begin{table}[ht!]
    \footnotesize
    \centering
    \begin{tabular}{c|c|c|c|c}
    \hline\hline
    \textbf{Methods of encoding physics} & Inputs & Outputs & Model ($\mathcal{H}$) &\textbf{Data} \\
    \hline
    PINNs & $x, t$ & $u/ f/b$ & $u_\theta/f_\theta/b_\theta$ & Measurements of $x, u, f, b$ \\
    \hline
    Pretrained NOs & $v$ & $u$ & $\mathcal{G}$ & Measurements of $v, u$\\
    \hline\hline
    \end{tabular}
    \caption{Interpretations of inputs, outputs and NN models $\mathcal{H}$ in PINNs and neural operators. We note that when solving hybrid problems with NOs, $\mathcal{G}$ is pretrained, often offline based on clean and simulated data, and therefore remain fixed in reconstructing $v$ and $u$ from their noisy measurements.}
    \label{tab:problems}
\end{table}

We then adopt the Bayesian framework \cite{dellaportas1995bayesian, wright1999bayesian} to address the noisy inputs-outputs in PINNs and pretrained NOs for solving hybrid problems. Specifically, we model the inputs $\chi$ as well as the parameter $\theta$ of the NN model $\mathcal{H}$ and establish likelihoods for measurements of both the inputs $\chi$ and outputs $\tilde{y}$ of $\mathcal{H}$ following \eqref{eq:noisy_model2}.
With the Bayes' theorem, we have the following:
\begin{equation}\label{eq:posterior}
    p(\theta, \chi|\mathcal{D}) \propto p(\mathcal{D}|\theta, \chi) p(\theta, \chi),
\end{equation}
where $p(\theta, \chi|\mathcal{D})$ is the posterior, $p(\mathcal{D}|\theta, \chi)$ is the likelihood, and $p(\theta, \chi)$ is the prior. 
Note that in \cite{yang2021b, meng2022learning, psaros2023uncertainty, zou2022neuraluq}, only the noise in the outputs is considered. Consequently, the posterior is exclusively concerned with the NN parameter $\theta$, i.e. $p(\theta|\mathcal{D}) \propto p(\mathcal{D}|\theta)p(\theta)$, and hence does not take into consideration the uncertainty induced by noisy inputs.

Similar as in \cite{yang2021b, meng2022learning, psaros2023uncertainty, zou2022neuraluq}, we assume independent and identically distributed (i.i.d.) distribution for data ($\mathcal{D} = \{\tilde{\chi}_i, \tilde{y}_i\}_{i=1}^N$, where $N$ denotes the number of data) and further assume $\epsilon_i$ and $\epsilon_o$ follow Gaussian distributions with mean zero and standard deviations $\sigma_i$ and $\sigma_o$, respectively. Then, by the independence between $\tilde{y}_i$ and $\tilde{\chi}_i$ conditional $\chi_i$, the likelihood can be expressed as follows:
\begin{equation}\label{eq:likelihood}
\begin{split}
    p(\mathcal{D}|\theta, \chi) &= \prod_{i=1}^N p(\tilde{y}_i|\theta, \chi_i) p(\tilde{\chi}_i|\chi_i) \\
    &= \prod_{i=1}^N \frac{1}{\sqrt{2\pi}\sigma_o}\exp(-\frac{||\tilde{y}_i - \mathcal{H}(\chi_i)||^2}{2\sigma_o^2})\frac{1}{\sqrt{2\pi}\sigma_{in}}\exp(-\frac{||\chi_i - \tilde{\chi}_i||^2}{2\sigma_{in}^2}),
\end{split}
\end{equation}
where $\chi_i, i=1,...,N$ denote the actual inputs where $\tilde{y}_i$ are observed, i.e. the exact values of $\tilde{\chi}_i$, and $||\cdot||$ denotes the Euclidean norm.
We note that, although not explicitly expressed, here $\mathcal{H}$ depends on $\theta$; see also Table \ref{tab:problems}. 
By the independence between the NN parameter $\theta$ and the actual input $\tilde{\chi}$, the prior can be expressed as:
\begin{equation}\label{eq:prior}
    p(\theta, \chi) = p(\theta) p(\chi).
\end{equation}
When $\mathcal{H}$ is an NN and no physics or neural operator is applied, the problem degenerates to a function approximation problem. A pedagogical example in addressing noisy inputs-outputs in function approximations is presented in \ref{sec:appendix_a}.

\subsection{PINNs}\label{sec:pinns}
Solving hybrid problems with noisy inputs-outputs under the PINN framework allows us to consider scenarios where both the spatial-temporal coordinate ($x$) and the value of functions ($u, f, b, \lambda$) are measured with noises. In this case, the posterior \eqref{eq:posterior} can be directly integrated into B-PINNs to handle problems involving ODEs/PDEs with noisy inputs-outputs. Given i.i.d. data of the sought solution $u$ ($\{\tilde{x}_i^u, \tilde{u}_i\}_{i=1}^{N_u}$), the source term $f$ ($\{\tilde{x}_i^f, \tilde{f}_i\}_{i=1}^{N_f}$), and/or the boundary term $b$ ($\{\tilde{x}_i^b, \tilde{b}_i\}_{i=1}^{N_b}$), the likelihood is expressed as follows:
\begin{equation}\label{eq:pinn_likelihood}
\begin{split}
    p(\mathcal{D} | \theta, \{x^u_i\}_{i=1}^{N_u}, &\{x^f_i\}_{i=1}^{N_f}, \{x^b_i\}_{i=1}^{N_b}) = \prod_{i=1}^{N_u} p(\tilde{u}_i|\theta, x^u_i)p(\tilde{x}^u_i|x^u_i)
    \\&\prod_{i=1}^{N_f} p(\tilde{f}_i|\theta, x^f_i) p(\tilde{x}^f_i|x^f_i) \prod_{i=1}^{N_b} p(\tilde{b}_i|\theta, x^b_i) p(\tilde{x}^b_i|x^b_i),
\end{split}
\end{equation}
where $x^u_i, x^f_i, x^b_i$ are the actual values of $\tilde{x}^u_i, \tilde{x}^f_i, \tilde{x}^b_i$, respectively, and
\begin{subequations}
    \begin{align}
        p(\tilde{u}_i|\theta, x^u_i) &= \frac{1}{\sqrt{2\pi}\sigma_o^u}\exp(-\frac{||u_\theta(x^u_i) - \tilde{u}_i||_2^2}{2{\sigma_o^u}^2}),\\
        p(\tilde{f}_i|\theta, x^f_i) &= \frac{1}{\sqrt{2\pi}\sigma_o^f}\exp(-\frac{||\mathcal{F}_\lambda[u_\theta](x^f_i) - \tilde{f}_i||_2^2}{2{\sigma_o^f}^2}), \\
        p(\tilde{b}_i|\theta, x^b_i) &= \frac{1}{\sqrt{2\pi}\sigma_o^b}\exp(-\frac{||\mathcal{B}_\lambda[u_\theta](x^b_i) - \tilde{b}_i||_2^2}{2{\sigma_o^b}^2}),
    \end{align}
\end{subequations}
in which $\sigma_o^u, \sigma_o^f, \sigma_o^b$ are the scales of the additive Gaussian noise in observing $u, f, b$, respectively. In addition, the likelihood for observing the inputs can be expressed as follows:
\begin{subequations}
    \begin{align}
        p(\tilde{x}_u^i|x_u^i) &= \frac{1}{\sqrt{2\pi}\sigma_{in}^u} \exp(-\frac{||x^u_i - \tilde{x}^u_i||^2}{2{\sigma_{in}^u}^2}),\\
        p(\tilde{x}_f^i|x_f^i) &= \frac{1}{\sqrt{2\pi}\sigma_{in}^f} \exp(-\frac{||x^f_i - \tilde{x}^f_i||^2}{2{\sigma_{in}^f}^2}), \\
        p(\tilde{x}_b^i|x_b^i) &= \frac{1}{\sqrt{2\pi}\sigma_{in}^b} \exp(-\frac{||x^b_i - \tilde{x}^b_i||^2}{2{\sigma_{in}^b}^2}),
    \end{align}
\end{subequations}
where $\sigma_{in}^u, \sigma_{in}^f, \sigma_{in}^b$ denote the scales of the additive Gaussian noise in observing the inputs. The prior reads as $p(\theta, x) = p(\theta) p(x)$.
We note that this (extended B-PINNs) framework is also capable of handling a combination of clean and noisy inputs; see Sec.~\ref{sec:example_1} for an example.

An alternative strategy for addressing noisy inputs-outputs is recasting the input noise as a heteroscedastic output noise, as discussed in \cite{williams2006gaussian, psaros2023uncertainty}. For example, \cite{mchutchon2011gaussian} explored the use of Gaussian processes to handle noisy inputs-outputs through a two-step procedure: (1) Gaussian processes regression is employed to fit the data and (2) then the input noise is substituted by an output noise via first-order Taylor expansion. Recall that our model for noisy inputs-outputs can be expressed as $\tilde{y} = \mathcal{H}(\tilde{\chi} - \sigma_{in}) + \sigma_o$ and hence the two-step approach proposed in \cite{mchutchon2011gaussian} can in fact be integrated into a one-step approach via automatic differentiation \cite{raissi2019physics}. We obtain the following approximation of the original model for one-dimensional $\tilde{\chi}$:
\begin{equation}\label{eq:recasting}
    \tilde{y} \approx \mathcal{H}(\tilde{\chi}) - \mathcal{H}^\prime(\tilde{\chi}) \epsilon_{in} + \epsilon_o.
\end{equation}
By adopting this approach, we do not need to model the actual input or the input noise. Instead, the original B-PINNs method can be directly applied with a heteroscedastic output noise, which follows a Gaussian distribution with mean zero and variance $(\mathcal{H}^\prime(\tilde{\chi}))^2 \sigma_{in}^2 + \sigma_o^2$. Note that this approach serves as an approximation of the original model for noisy inputs-outputs. It may not work as well as the original model, especially in solving ODEs/PDEs. An  empirical comparison is conducted in Sec.~\ref{sec:example_1}.

\subsection{NOs}\label{sec:nos}

When pretrained NOs are used to encode physics in hybrid problems, as discussed in Sec.~\ref{sec:2}, their inputs and outputs become functions. For the sake of simplicity in analyzing uncertainties, in this work we consider scenarios where only values of functions are noisy and corresponding coordinates are clean. As shown in Fig.~\ref{fig:1}, solving a hybrid problem with a pretrained NO $\mathcal{G}$ only requires regression of the input function, $v$, since the regressed output function can be obtained immediately by $\mathcal{G}(\mathbf{v})$ where $\mathbf{v}$ denotes its discretization. 
In this regard, we directly model the discretization of $v$ on a mesh, i.e. $\mathbf{v}_\psi = [v^\psi_1, ..., v^\psi_n]$ where $n$ denotes the mesh size. The data are $\mathcal{D}_u = \{x_i^u, \tilde{u}_i\}_{i=1}^{N_u}$ and $\mathcal{D}_v = \{\tilde{v}_i\}_{i\in I_v}$, where $I_v$ is the index set of the grid point on which $v$ is measured. 
That is, in the hybrid problem we assume that $v$ is observed on the subset of the same mesh used in training $\mathcal{G}$. Then, the posterior can be written as
\begin{equation}\label{eq:operator_posterior}
    p(\mathbf{v}_\psi | \mathcal{D})  \propto p(\mathcal{D}_u, \mathcal{D}_v | \mathbf{v}_\psi)p(\mathbf{v}_\psi) = p(\mathcal{D}_u | \mathbf{v}_\psi) p(\mathcal{D}_v | \mathbf{v}_\psi) p(\mathbf{v}_\psi),
\end{equation}
where the last equality comes from the independence of observing $\mathcal{D}_u$ and $\mathcal{D}_v$ conditional on $\mathbf{v}_\psi$.
Here, $p(\tilde{\mathbf{v}})$ is the prior and is either analytical known, e.g. a multivariate Gaussian, or can be estimated from the data used to train the NO \cite{lu2021learning, li2020fourier}. The likelihood is formulated as follows, assuming additive Gaussian noises with scales $\sigma_o$ and $\sigma_{in}$ in observing $u$ and $v$, respectively:
\begin{subequations}
    \begin{align}
        p(\mathcal{D}_u | \mathbf{v}_\psi) &= \prod_{i=1}^{N_u}\frac{1}{\sqrt{2\pi}\sigma_o}\exp(-\frac{||\mathcal{G}(\mathbf{v}_\psi)(x^u_i) - \tilde{u}_i||^2}{2{\sigma_o}^2}), \label{eq:operator_likelihood_1}\\
        p(\mathcal{D}_v | \mathbf{v}_\psi) &= \prod_{i\in I_v}\frac{1}{\sqrt{2\pi}\sigma_{in}}\exp(-\frac{||v^\psi_i - \tilde{v}_i||^2}{2{\sigma_{in}}^2})\label{eq:operator_likelihood_2}.
    \end{align}
\end{subequations}

\subsection{Posterior estimate}

In this study, we employ the Hamiltonian Monte Carlo (HMC) \cite{neal2011mcmc} to obtain samples from the posterior. For fast and stable implementations, a Python library termed NeuralUQ \cite{zou2022neuraluq} is used. 
When using PINNs to encode the physics, we primarily focus on the regression of the NN surrogate, and therefore in the testing stage the input $x$ is clean.
In this regard, the predicted mean $\hat{u}(x)$ estimated by $\frac{1}{M}\sum_{i=1}^M u_{\theta_i}(x)$ is used as the prediction and the predicted variance defined as $\frac{1}{M}\sum_{i=1}^M (u_{\theta_i}(x) - \hat{u})^2$ is used as the quantified uncertainty. Here, $\{\theta_i\}_{i=1}^M$ are the posterior samples of the NN parameter $\theta$ ($M$ denotes the number of posterior samples).
In NOs, we are interested in the reconstruction of the input and output functions. The predicted means estimated by $\hat{\mathbf{v}} = \frac{1}{M}\sum_{i=1}^M\mathbf{v}_{\psi}^i$ and $\hat{u}(x) = \frac{1}{M}\sum_{i=1}^M \mathcal{G}(\mathbf{v}_{\psi}^i)(x)$ for $\mathbf{v}$ and $u$, respectively, are used as the predictions, where $\{\mathbf{v}_{\psi}^i\}_{i=1}^M$ are the posterior samples of $\mathbf{v}_\psi = [v^\psi_1, ..., v^\psi_n]$. Similarly, the predicted variances $\frac{1}{M}\sum_{i=1}^M(\mathbf{v}_{\psi}^i - \hat{\mathbf{v}} )^2, \frac{1}{M}\sum_{i=1}^M (\mathcal{G}(\mathbf{v}_{\psi}^i)(x) - \hat{u}(x))^2$ are used as the quantified uncertainties.

\section{Numerical examples}\label{sec:4}

In this section we present four computational examples: (1) a 1D nonlinear Poisson equation, (2) a 1D Burgers equation, (3) a 1D reaction-diffusion equation with heteroscedastic diffusion coefficient, and (4) a 120-dimensional parametric Darcy problem. In the first example, we employ the present method to quantify uncertainties for PINNs with noisy inputs and outputs, and in the rest three examples, we use the proposed approach to quantify uncertainties in two different neural operators, i.e., DeepONet and FNO, with noisy inputs/outputs. 

We note that the NO is pretrained offline with clean and sufficient data such that it approximates the solution operator accurately and therefore is able to represent the physics.
Furthermore, in the downstream hybrid problem, the test functions are in-distribution and therefore follow the physics encoded in the NOs beforehand.  In the literature of NOs \cite{lu2021learning, li2020fourier, kovachki2021neural}, data of the input functions for training and testing are often sampled from general distributions whose probabilistic density functions are analytically known, e.g. Gaussian random fields. 
We follow this setup in this work, and in hybrid problems with pretrained NOs, the distribution used to generate training data is employed as the prior of the discretization of the input functions, i.e. $p(\mathbf{v}_\psi)$ in \eqref{eq:operator_posterior}.
Details of data generation, the training of NOs, and hyperparameters in numerical experiments can be found in \ref{sec:appendix_b}. 

\subsection{A nonlinear 1D Poisson equation with PINNs}\label{sec:example_1}

Consider the following 1D nonlinear Poisson equation:
\begin{equation}\label{eq:poisson}
    \kappa \frac{\partial^2 u}{\partial x^2} - \lambda u^3 = f(x),~ x\in [0, 1],
\end{equation}
with the Dirichlet boundary condition $u(0) = u(1) = 1$ and $\kappa$ and $\lambda$ being non-negative constants. The data are generated by assuming $u(x) = \cos^3(2\pi x)$, $\kappa = 0.01$ and $\lambda=0.1$. In this example, we employ the B-PINN approach and consider noisy measurements of $u, f$ and $x$ with the present approach. 

\subsubsection{The forward problem}

\begin{table}[ht]
    \footnotesize
    \centering
    \begin{tabular}{c|c|c}
    \hline\hline
       & Error of $f$ (\%) & Error of $u$ (\%) \\
       \hline
       clean data  & $0.03$ & $0.00$\\
       \hline
       clean input and noisy output  & $7.16$ & $3.57$ \\
       \hline
       noisy input and clean output  & $13.50$ & $11.73$ \\
       \hline\hline
    \end{tabular}
    \caption{
    PINN for Poisson equation: Inference of $u$ and $f$ with different training data for $f$. The error is the relative $L_2$ error.
    }
    \label{tab:example_1_1}
\end{table}

To demonstrate the importance and necessity of respecting the noise in inputs, we start with a \textit{forward problem} where $\kappa$ and $\lambda$ are known and we solve \eqref{eq:poisson} given data on $(x, f)$. We first employ vanilla PINNs for the following three different scenarios: clean input and output data, clean input but noisy output data, and noisy input but clean output data. In all scenarios, the boundary condition is hard-encoded in the model, and we assume that 51 available measurements for $(x, f)$, which are equidistantly distributed on $[0, 1]$, are used to train PINNs. The measurement noise for both $x$ and $f$ is additive Gaussian noise with the scale $0.01$.
Results are shown in Table~\ref{tab:example_1_1} and Fig.~\ref{fig:example_1_1}, from which we can see noisy data impair the performance of PINNs significantly. In particular, even with the same scale, the noise in $x$ has larger damage upon the accuracy of PINNs in solving the equation than the noise in $f$.

\begin{figure}[h!]
    \centering
    \subfigure[Inference/fitting of $f$]{
    \includegraphics[scale=.4]{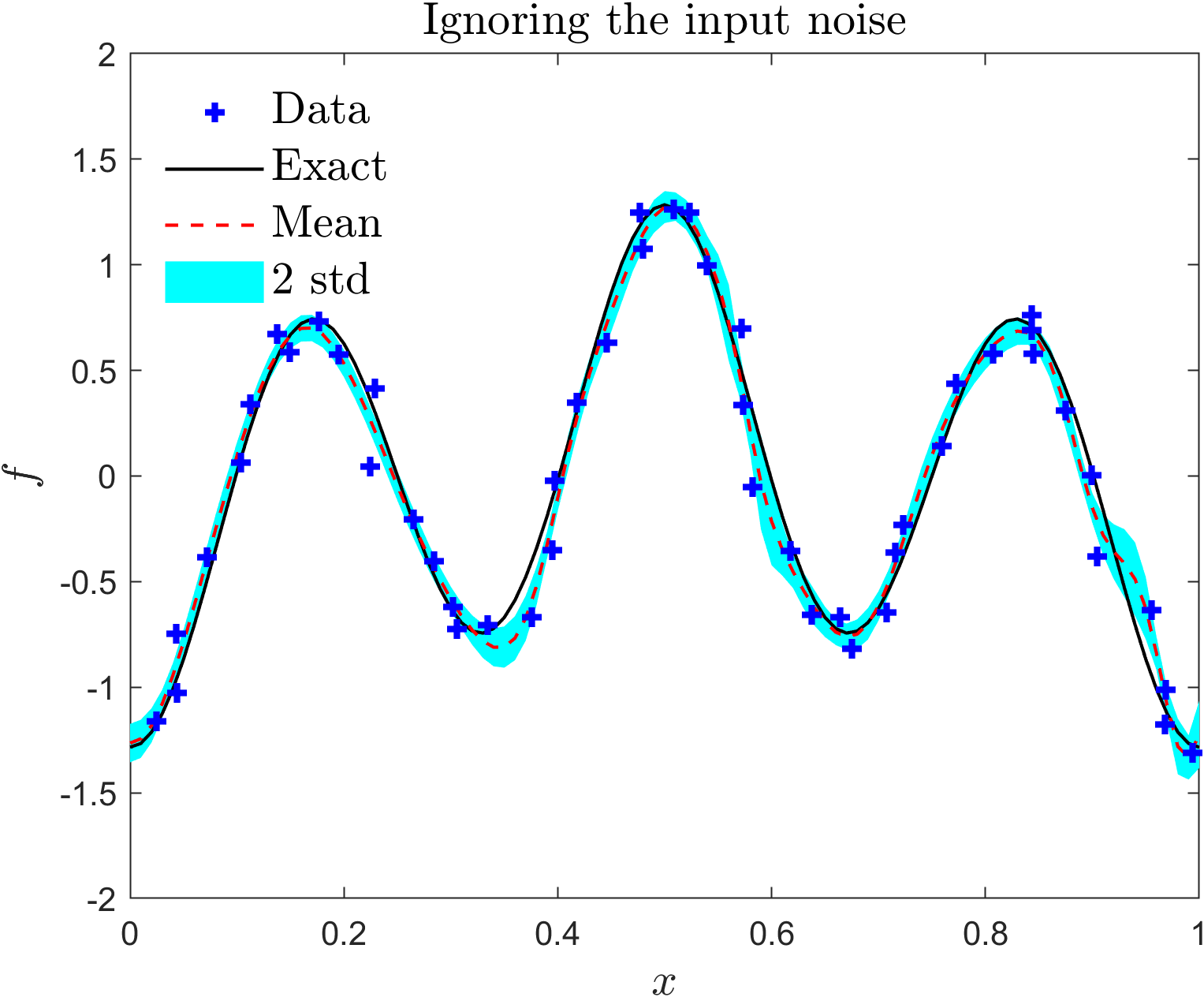}
    \includegraphics[scale=.4]{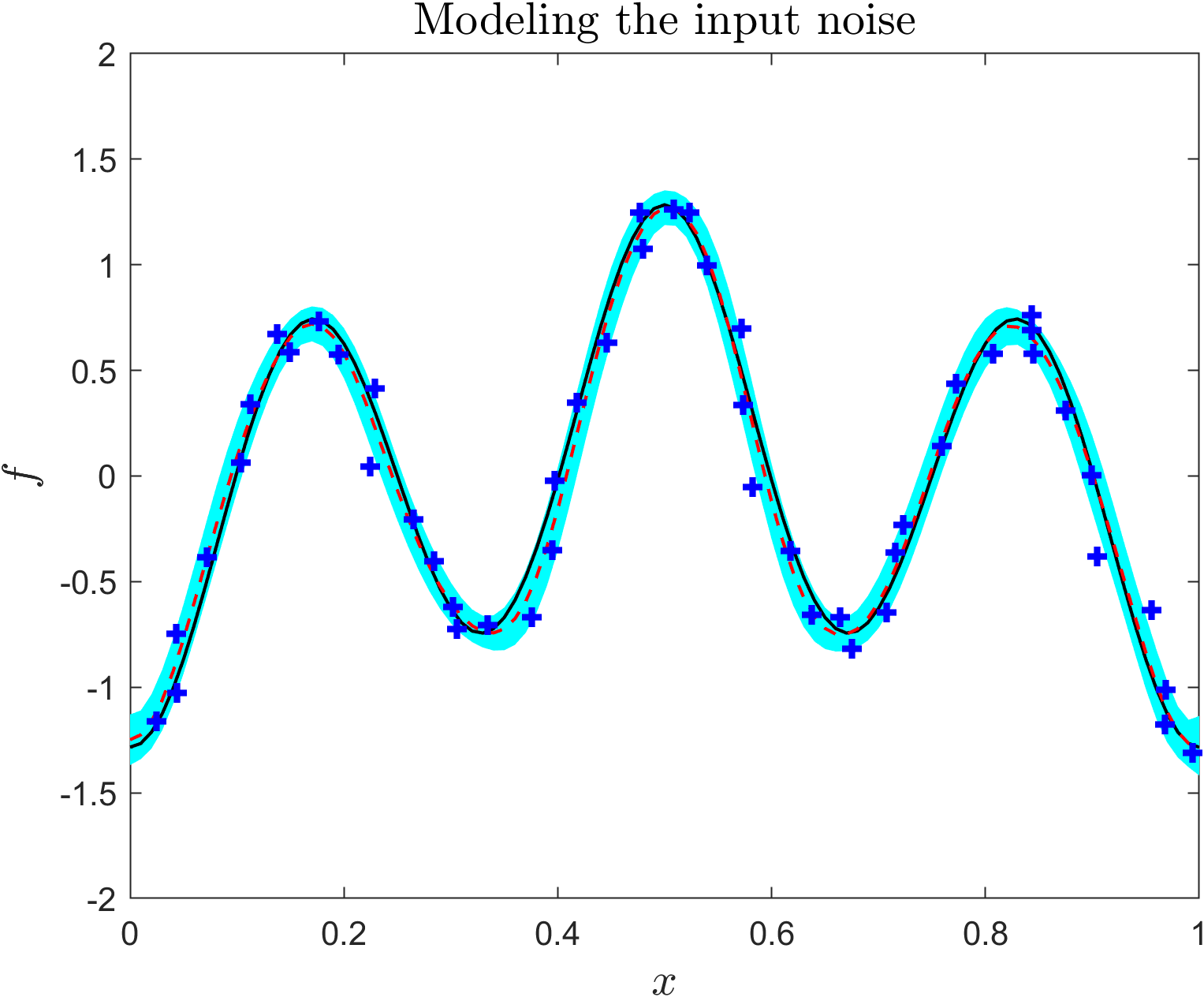}
    \includegraphics[scale=.4]{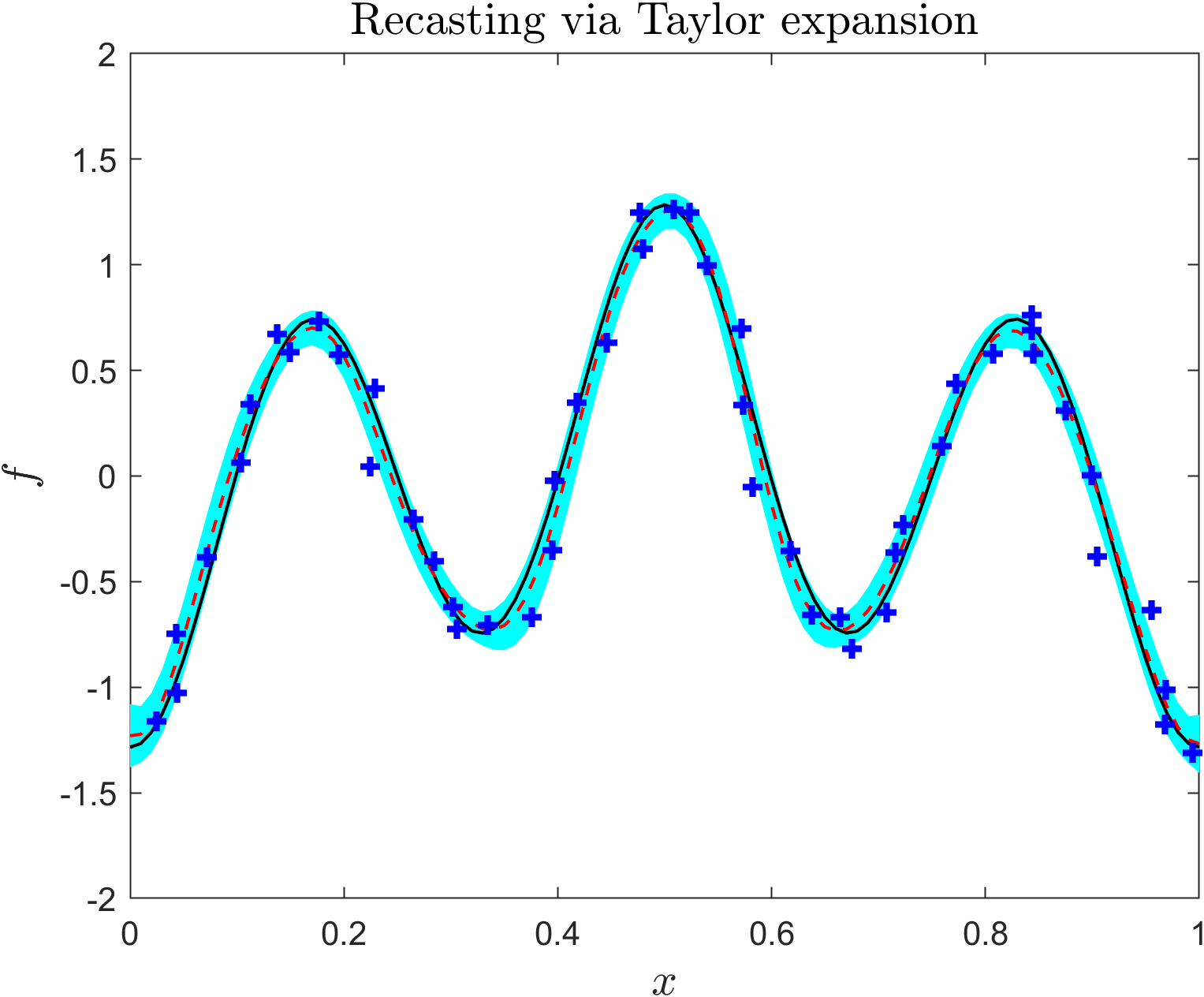}
    }
    \subfigure[Inference of $u$]{
    \includegraphics[scale=.4]{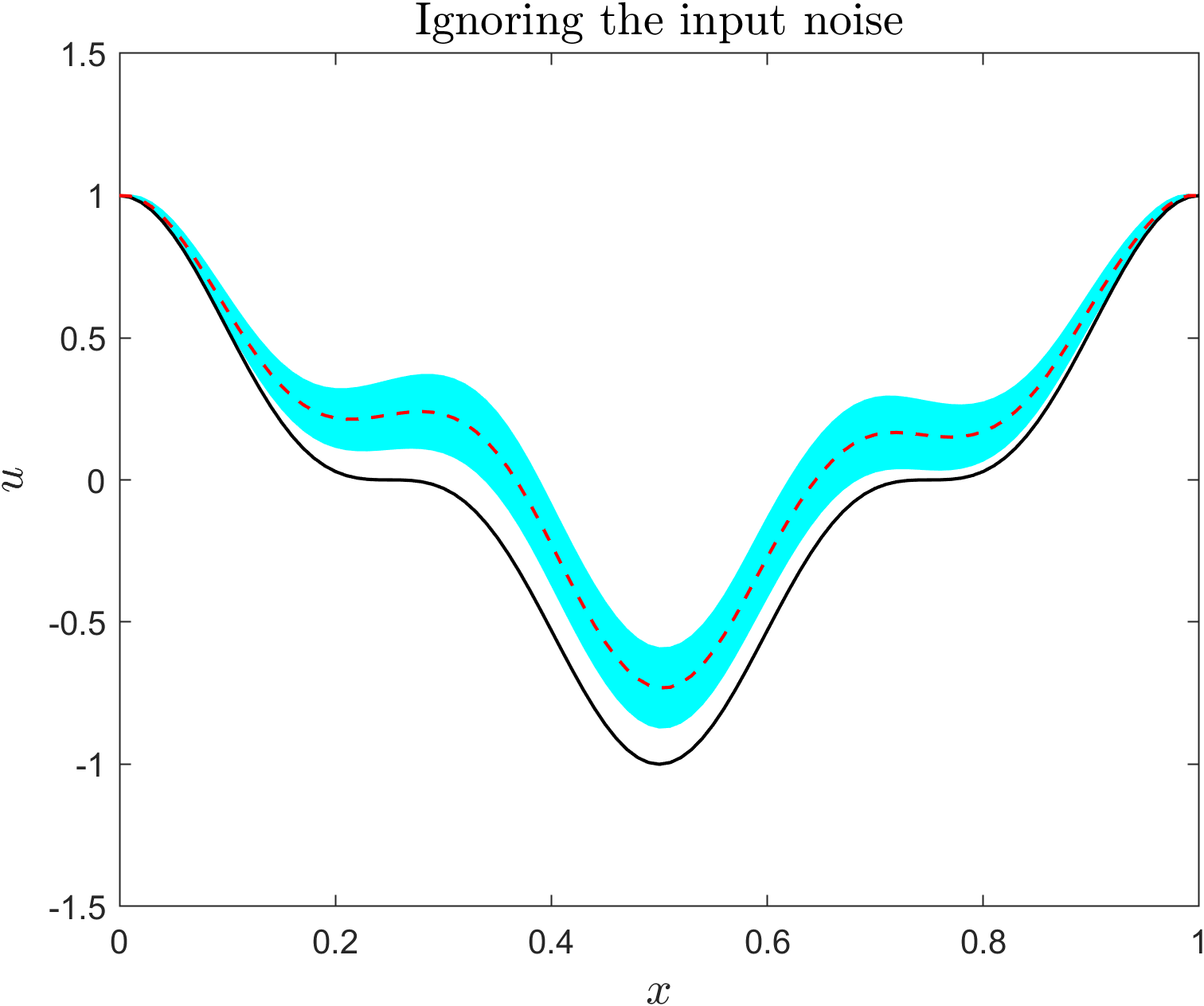}
    \includegraphics[scale=.4]{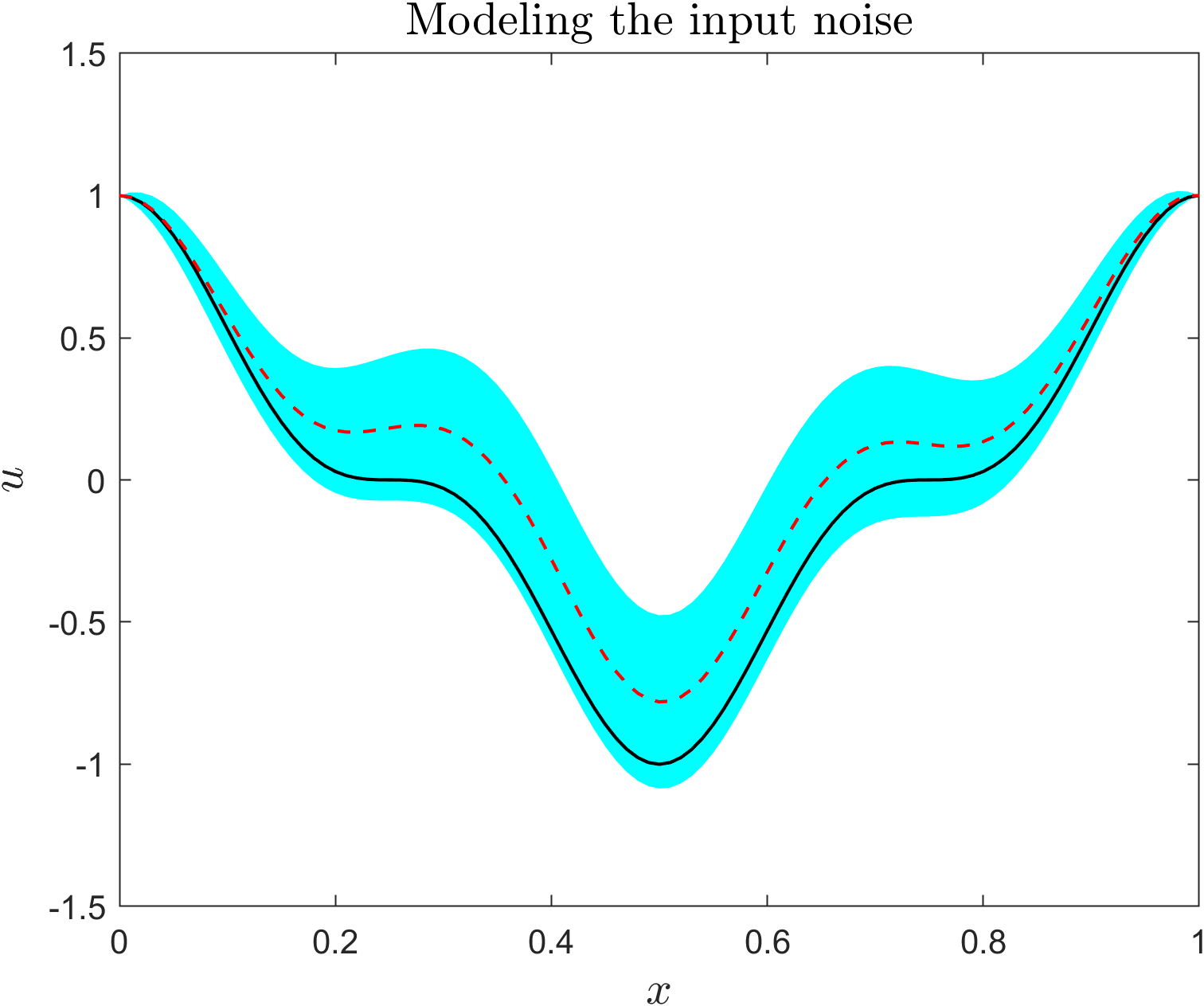}
    \includegraphics[scale=.4]{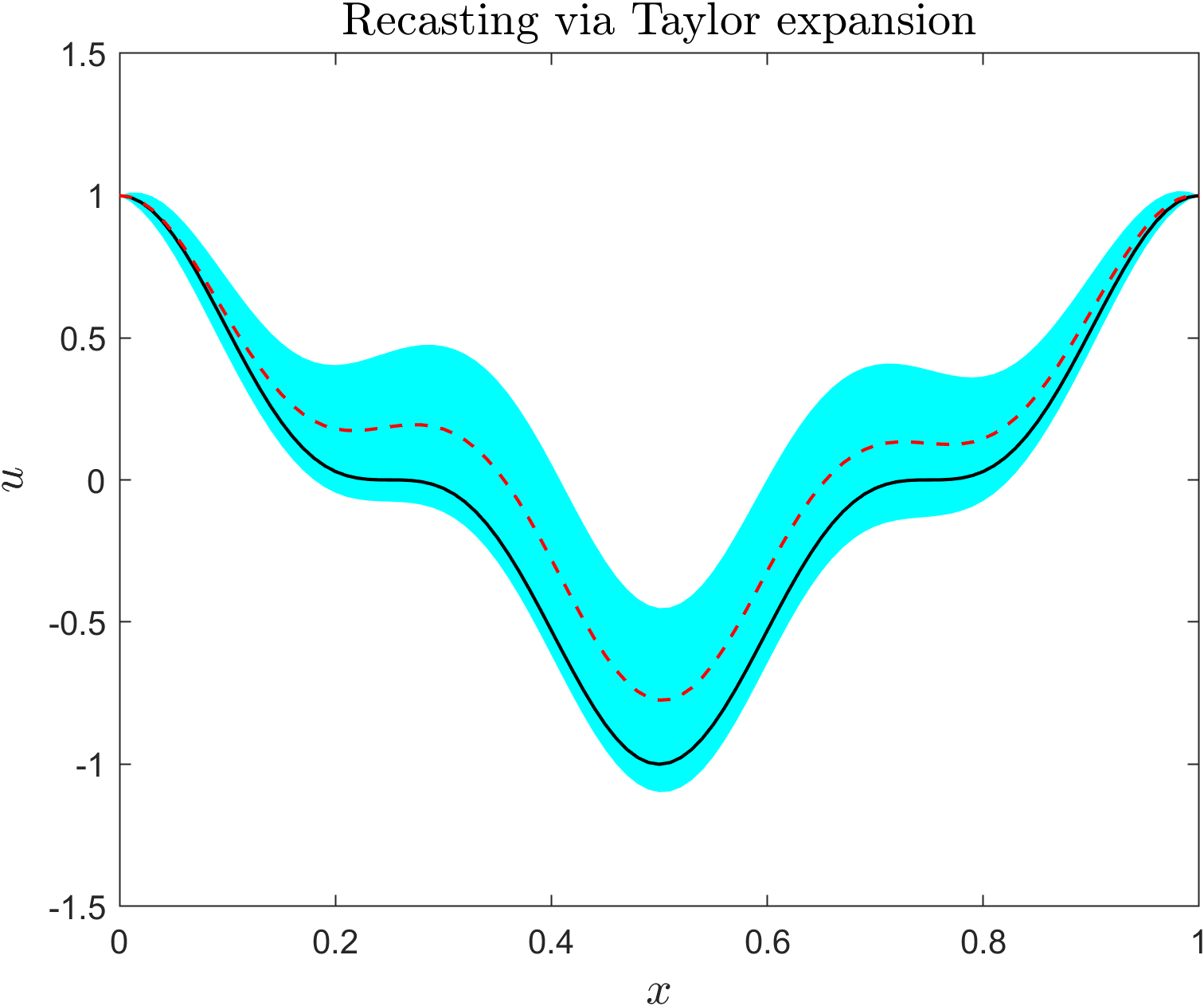}
    }
    \caption{Results from solving \eqref{eq:poisson} in the B-PINN framework based on noisy inputs-outputs data of $f$ using different approaches. The boundary condition is clean and hard-encoded in the modeling. From left to right are results from the original B-PINN method which ignores the noise in $x$, our approach which models the noise in $x$, and the recasting method which recasts the noise in $x$ as a heteroscedastic noise in $f$ via first-order Taylor expansion.}
    \label{fig:example_1_2}
\end{figure}

\begin{table}[ht]
    \footnotesize
    \centering
    \begin{tabular}{c|c|c|c}
    \hline\hline
       & Ignoring the input noise & Modeling the input noise & Recasting \\
       \hline
       Error of $f$  & $16.18\%$ & $8.16\%$ & $9.13\%$\\
       \hline
       Error of $u$  & $32.17\%$ & $28.17\%$ & $28.65\%$ \\
    \hline\hline
    \end{tabular}
    \caption{Errors of solving \eqref{eq:poisson} with noisy inputs-outputs data of $f$ using different methods.}
    \label{tab:example_1_2}
\end{table}

Now we solve \eqref{eq:poisson} with noisy inputs-outputs data of $f$, i.e., $(x, f)$.
In this case, $51$ data points of $f$ are sampled equidistantly on $[0, 1]$ and their inputs and outputs are corrupted by additive Gaussian noise with scales $0.01$ and $0.05$, respectively. The boundary condition is clean and hard-encoded in the modeling. 
We employ and compare three methods: the original B-PINN method and two of our approaches proposed in Sec.~\ref{sec:pinns}.
Results are presented in Fig.~\ref{fig:example_1_2} and Table~\ref{tab:example_1_2}. We can see that the original B-PINN method, which ignores the noise in $x$, is not able to provide accurate prediction (as shown in Table~\ref{tab:example_1_2}) and reasonable uncertainty of $u$ and $f$ (as shown in Fig.~\ref{fig:example_1_2}) that bounds the error between the exact and the prediction. 
Overfitting in predicting $f$ is observed from results of B-PINNs shown in Fig.~\ref{fig:example_1_2}(a), showing the consequence of the unawareness of the input noise.
Meanwhile, our approaches present much better results in terms of the predicted means and uncertainties: errors are small and bounded by the predicted uncertainties while the predicted uncertainties are reasonable and not over-confident.
We show in Fig.~\ref{fig:example_1_2}(b) that the consideration of the noise in the input with our approaches yields significantly larger quantified uncertainty in inferring $u$, indicating successful capturing of uncertainty induced by noisy inputs and demonstrating the effect of uncertain inputs in solving \eqref{eq:poisson} with noisy data.
Besides, as presented in Fig.~\ref{fig:example_1_2}(a), the uncertainty band of our approach, which is obtained by $[\mu-2\sigma, \mu+2\sigma]$ where $\mu$ is the predicted mean and $\sigma$ is the predicted standard deviation, is horizontally (along the $x$-axis) larger than the original B-PINNs, showing the effect of considering the input noise.
As reported in Table~\ref{tab:example_1_1}, modeling noises in both the coordinate and the value $f$ with our approaches results in significantly higher accuracy in fitting $f$ as well as inferring $u$, proving the effectiveness of the proposed approach in solving \eqref{eq:poisson} given noisy inputs-outputs data.

\subsubsection{The inverse problem}

\begin{figure}[h!]
    \centering
    \subfigure[Inference/fitting of $f$]{
    \includegraphics[scale=.4]{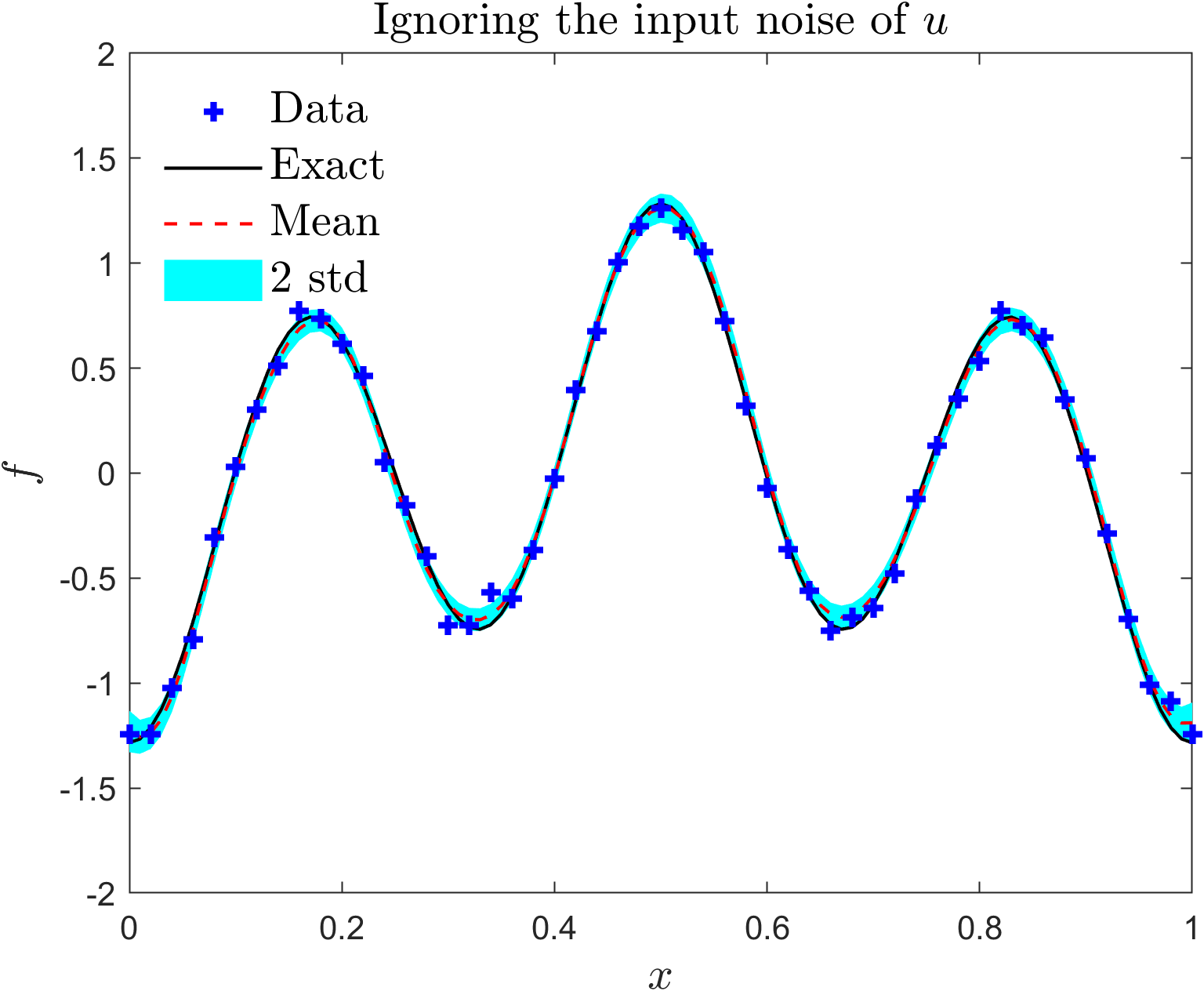}
    \includegraphics[scale=.4]{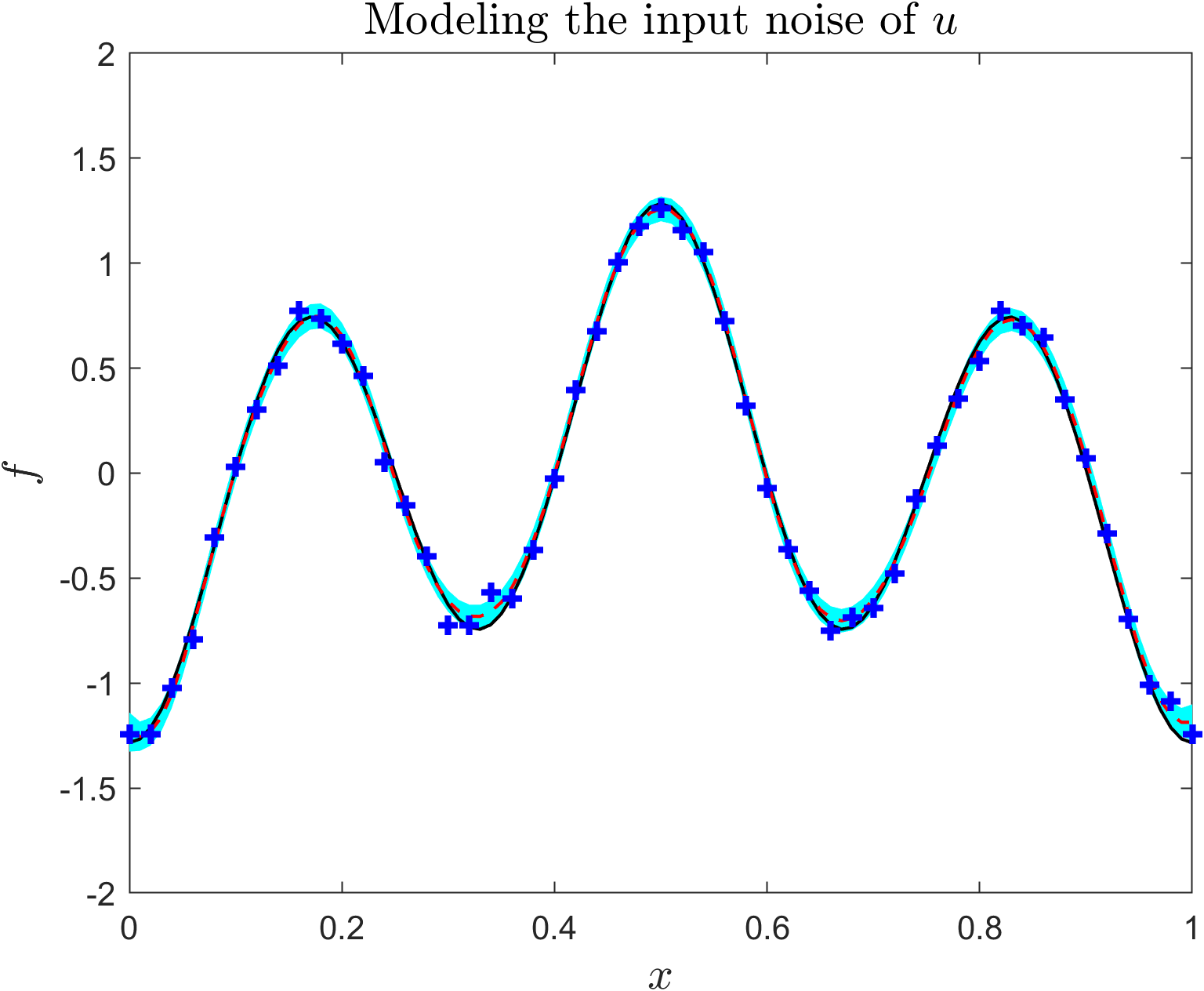}
    \includegraphics[scale=.4]{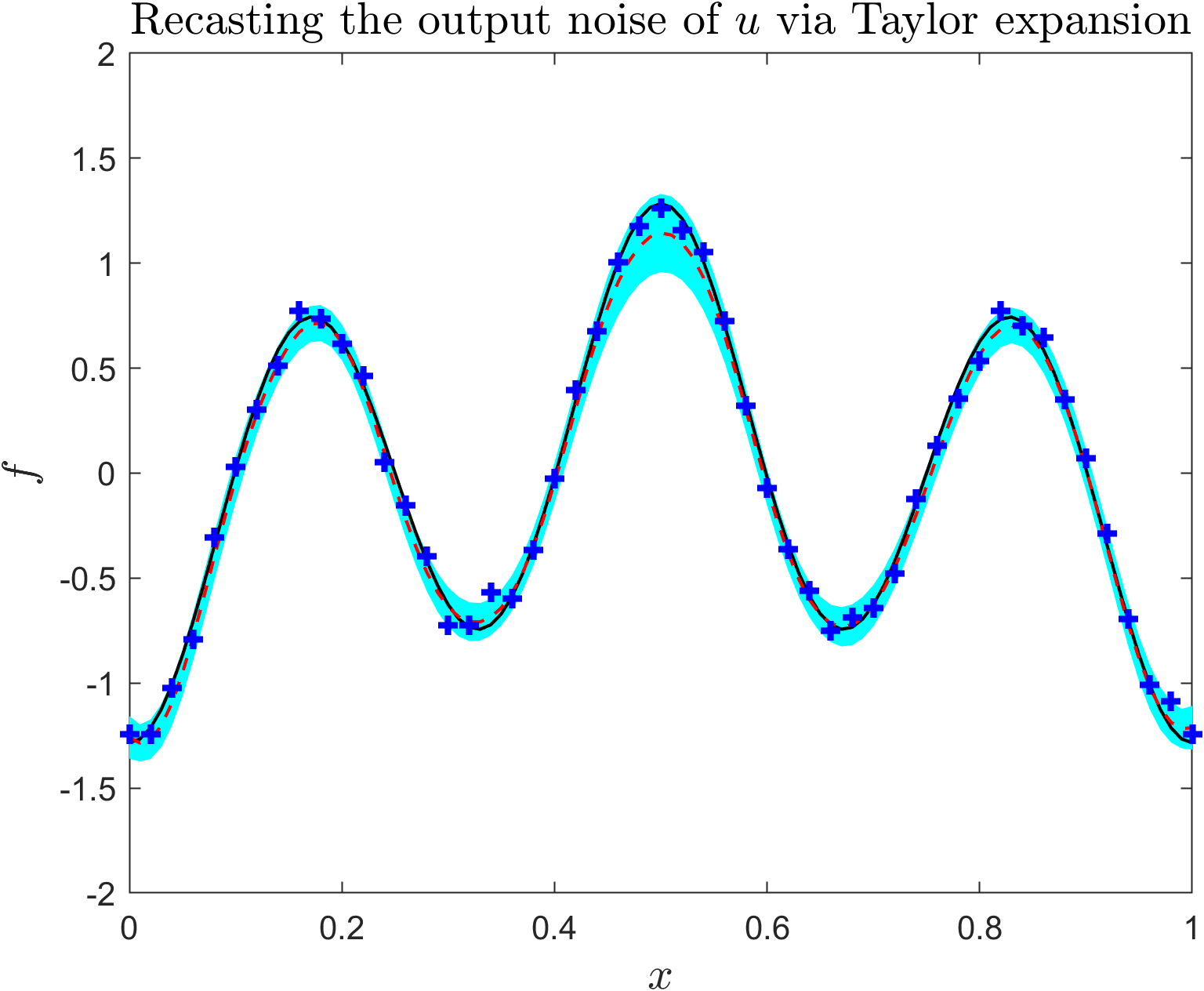}
    }
    \subfigure[Inference/fitting of $u$]{
    \includegraphics[scale=.4]{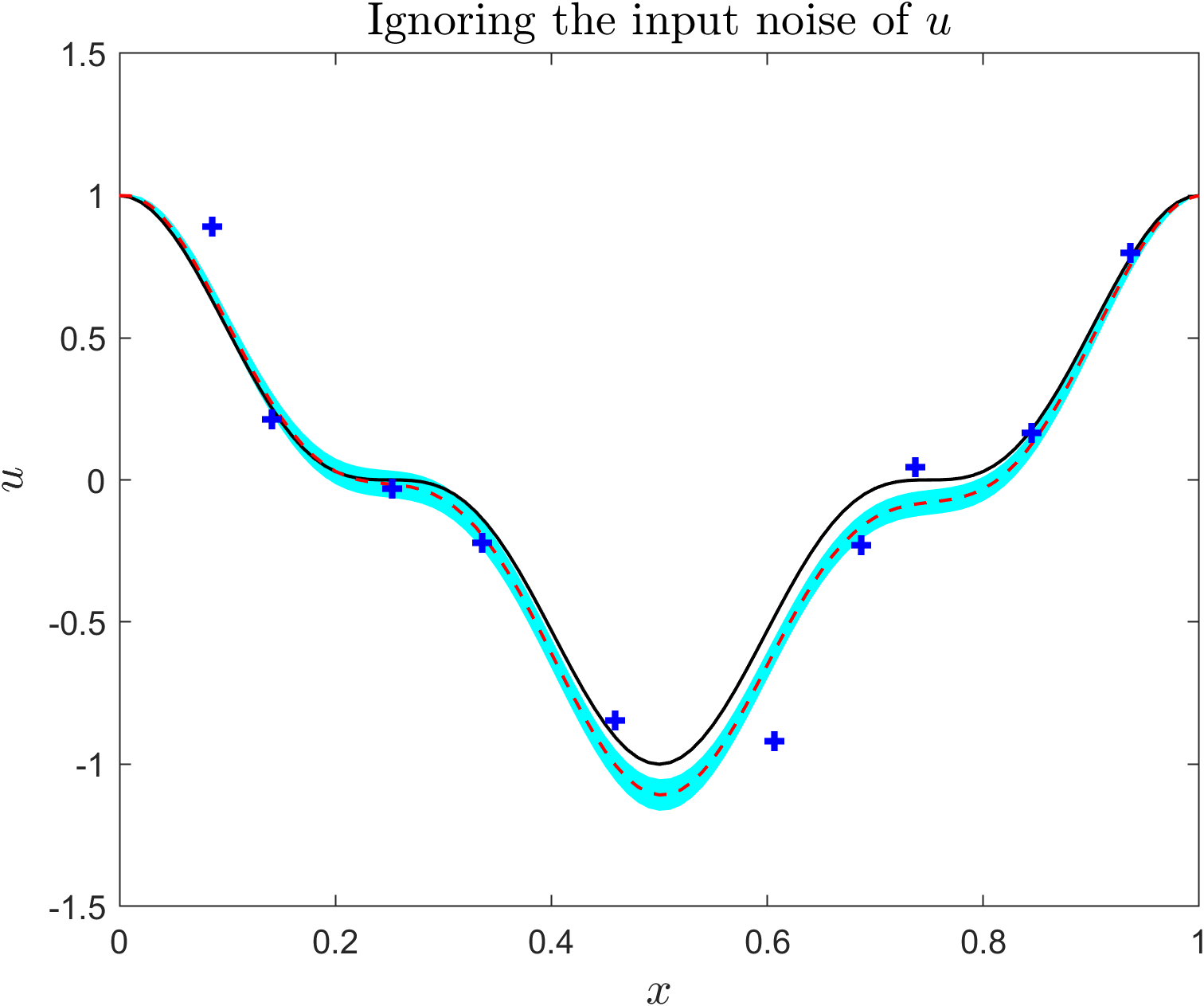}
    \includegraphics[scale=.4]{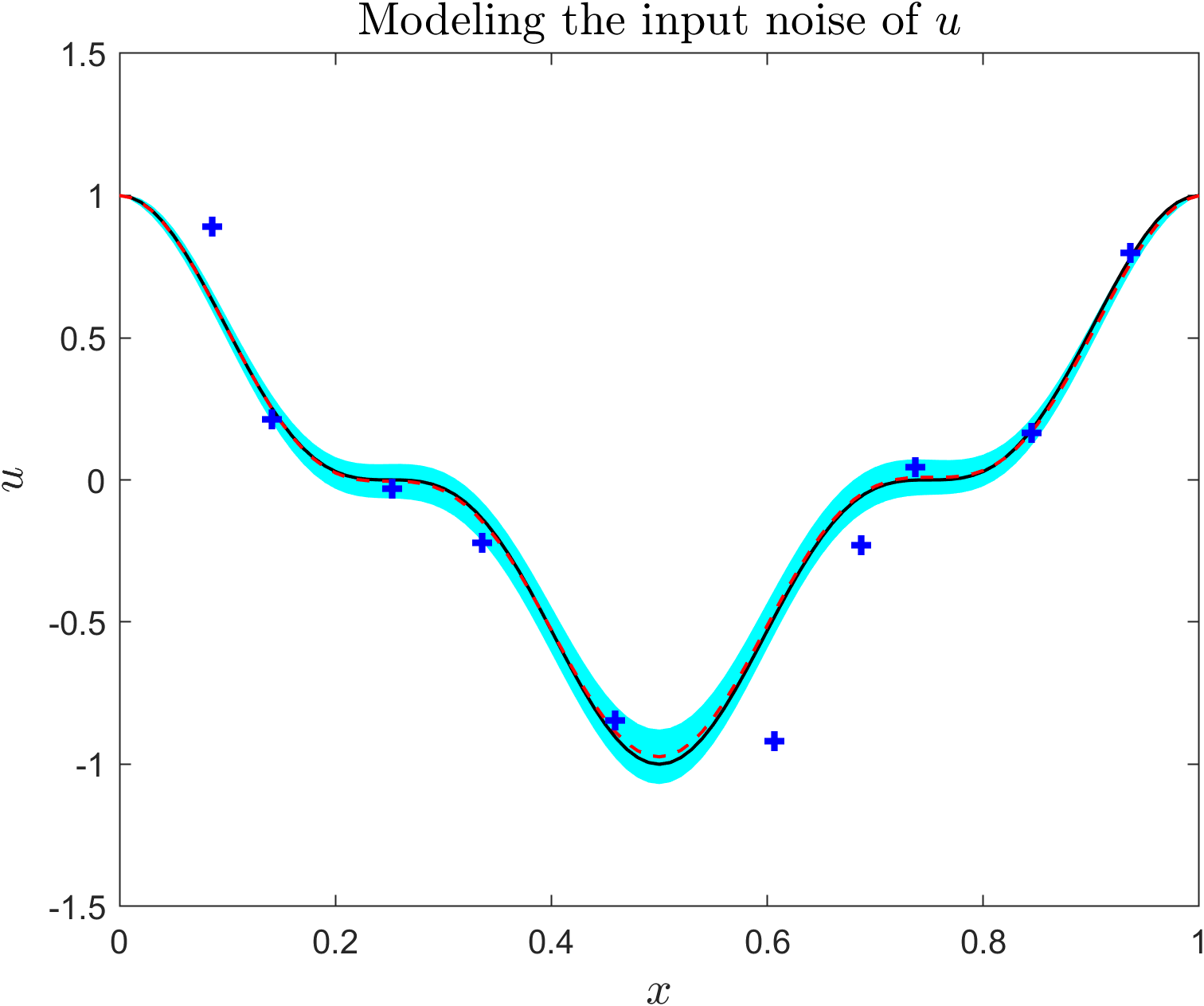}
    \includegraphics[scale=.4]{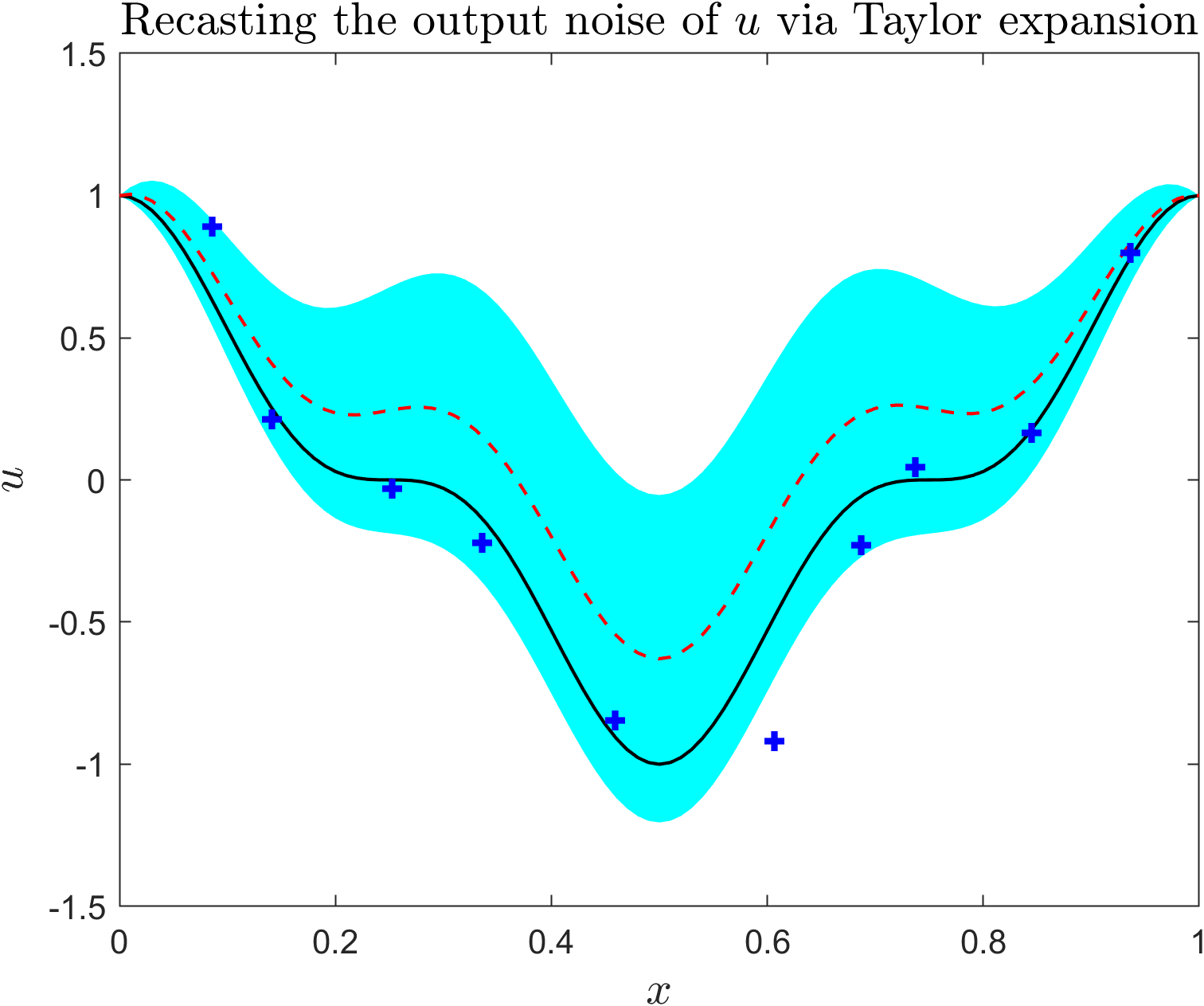}
    }
    \caption{Results from solving \eqref{eq:poisson} with $\lambda$ unknown based on noisy inputs-outputs data of $u$ and $f$. The boundary condition is clean and hard-encoded in the modeling. We note that in this case data of $u$ are noisy in both the input and the output while data of $f$ are noisy in the output but clean in the input, demonstrating the flexibility of our approach. From left to right are results from the original B-PINN method, our approach, and the recasting method.}
    \label{fig:example_1_3}
\end{figure}

\begin{table}[ht]
    \footnotesize
    \centering
    \begin{tabular}{c|c|c|c}
    \hline\hline
       & Ignoring the input noise & Modeling the input noise & Recasting \\
       \hline
       Error of $f$  & $4.85\%$ & $4.44\%$ & $7.80\%$\\
       \hline
       Error of $u$  & $12.54\%$ & $2.20\%$ & $43.90\%$ \\
       \hline
       Inference of $\lambda$ & $0.0591\pm0.0204$ & $0.1542\pm0.0668$ & $0.1810\pm 0.0644$\\
    \hline\hline
    \end{tabular}
    \caption{Inferences of $\lambda$ (mean $\pm$ standard deviation) and relative $L_2$ errors of inferences of $u$ and $f$ using different methods. The reference solution for $\lambda$ is $1.5$.}
    \label{tab:example_1_3}
\end{table}

We then move to an \textit{inverse problem} where $\kappa=0.01$ is known but $\lambda$ is not. In this case, we would like to identify $\lambda$ as well as to reconstruct $u$ from noisy data of $u$ and $f$. Specifically, 51 data points of $f$ and 10 data points of $u$, equidistantly sampled on $[0, 1]$, are available. 
To demonstrate the flexibility and compatibility of the proposed approach in handling a hybrid of clean and noisy inputs, in this case we corrupt the data of $u$ in both their inputs and outputs but corrupt data of $f$ only in their inputs. 
The scales of the additive Gaussian noise are $0.02$, $0.05$, and $0.05$ for the input of $u$, the output of $u$, and the output of $f$, respectively.
Results are shown in Fig.~\ref{fig:example_1_3} and Table~\ref{tab:example_1_3}. As shown in Fig.~\ref{fig:example_1_3}(a), the original B-PINN method is able to provide accurate inference as well as reasonable uncertainty of $f$ because data of $f$ are only corrupted in the output. However, as presented in Fig.~\ref{fig:example_1_3}(b) and Table~\ref{tab:example_1_3}, it fails to yield trustworthy inferences of $u$ and $\lambda$ (errors are large while predicted uncertainties are small) because it does not take into consideration the noise in the input of $u$. 
As a comparison, results of our approach, as displayed in Table~\ref{tab:example_1_3}, show significantly smaller error in inferring $\lambda$ and reconstructing $u$, demonstrating the effectiveness of the proposed approach in solving the inverse problem with noisy inputs-outputs data.
Moreover, Fig.~\ref{fig:example_1_3}(b) exhibits that our approach is capable of producing reliable predictions, in which the predicted uncertainty bounds the error between the predicted mean and the exact of $u$.
Furthermore, we show in Fig.~\ref{fig:example_1_3} that recasting the output noise via Taylor expansion yields under-confident predictions of both $u$ and $f$ and hence does not work as well as our approach in this case.
\subsection{1D Burgers equation with FNOs}\label{sec:example_2}

\begin{figure}[ht!]
    \centering
    \subfigure[Reconstruction of $u|_{t=0}$.]{
        \includegraphics[scale=.4]{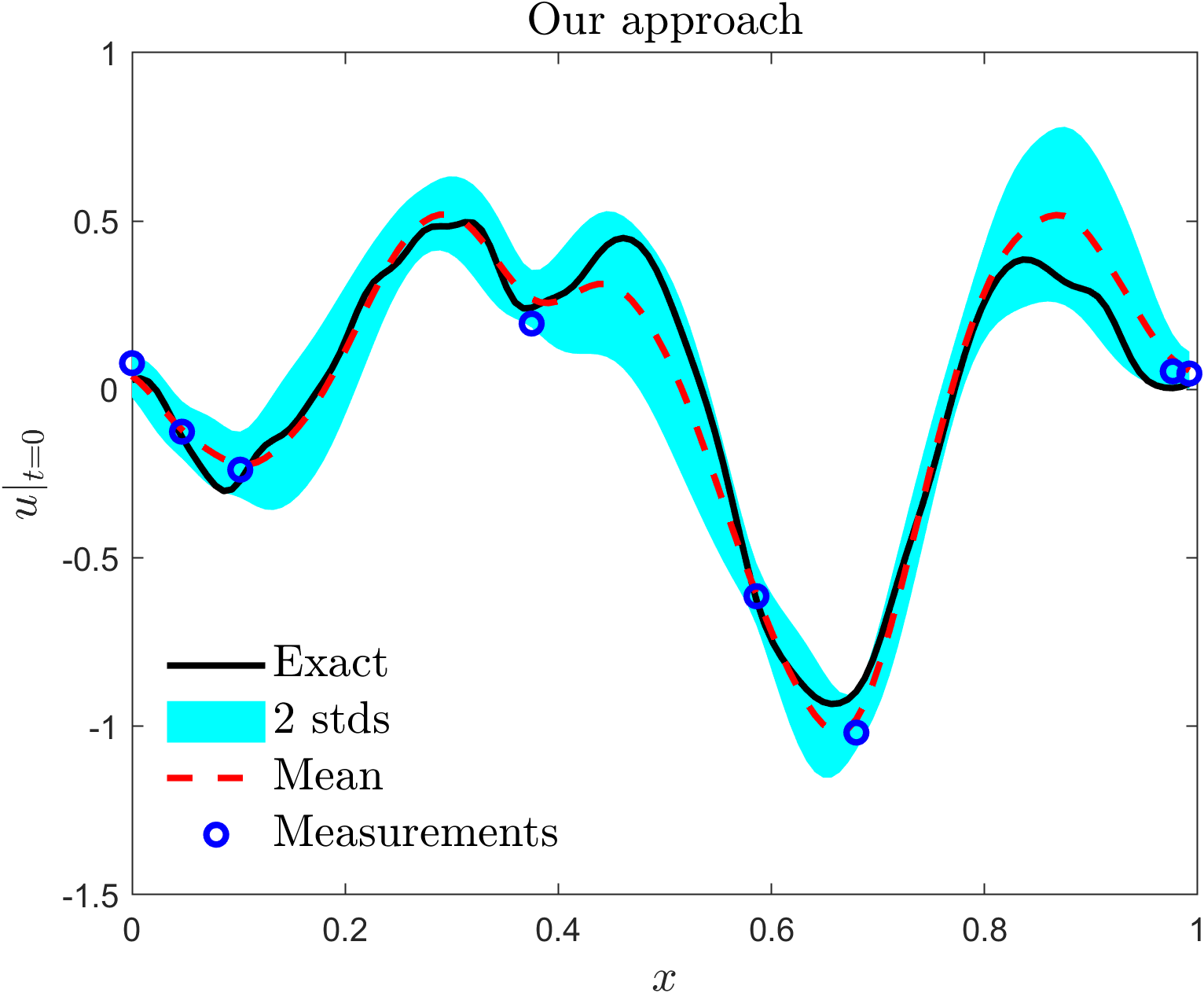}
        \includegraphics[scale=.4]{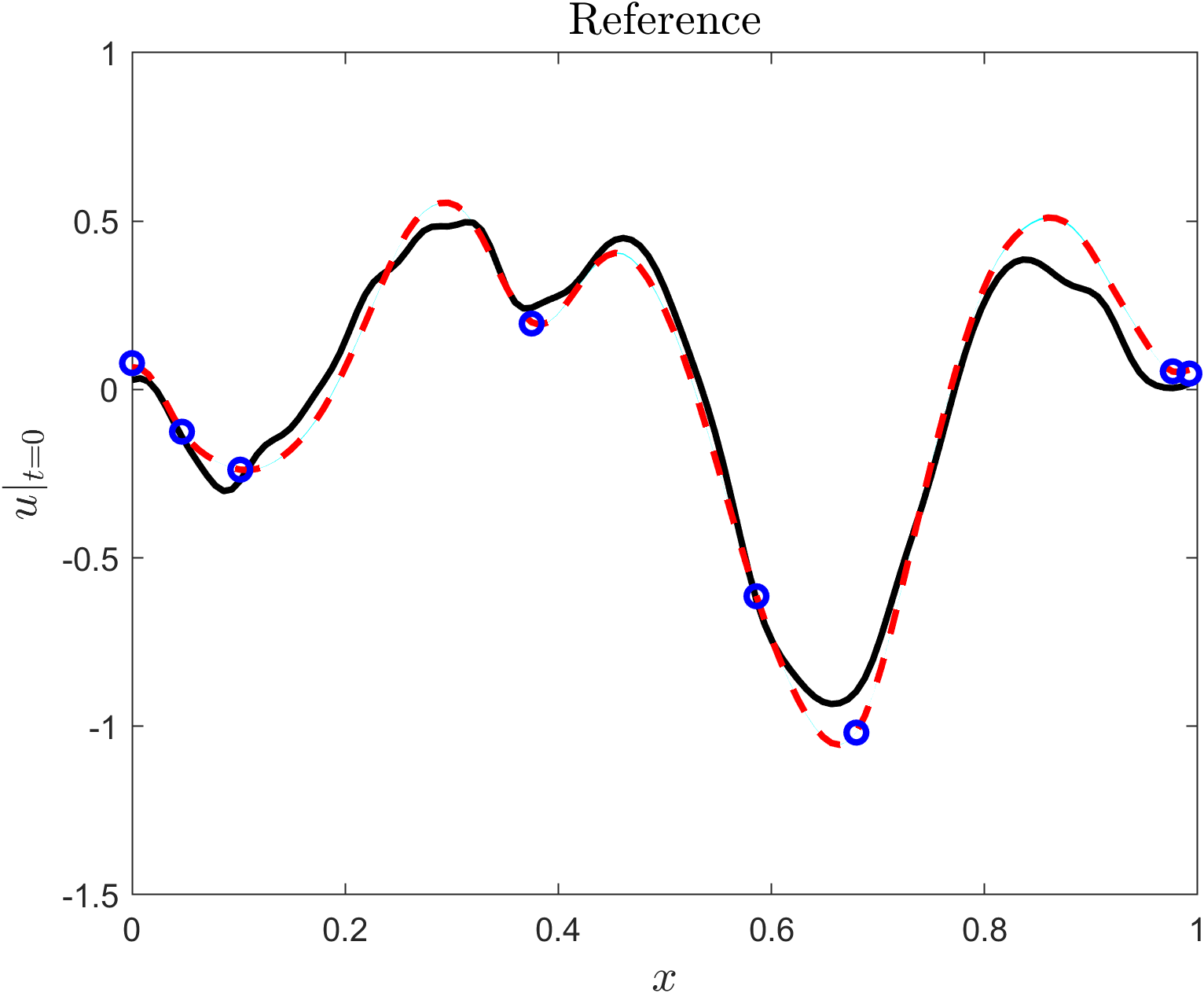}
        \includegraphics[scale=.4]{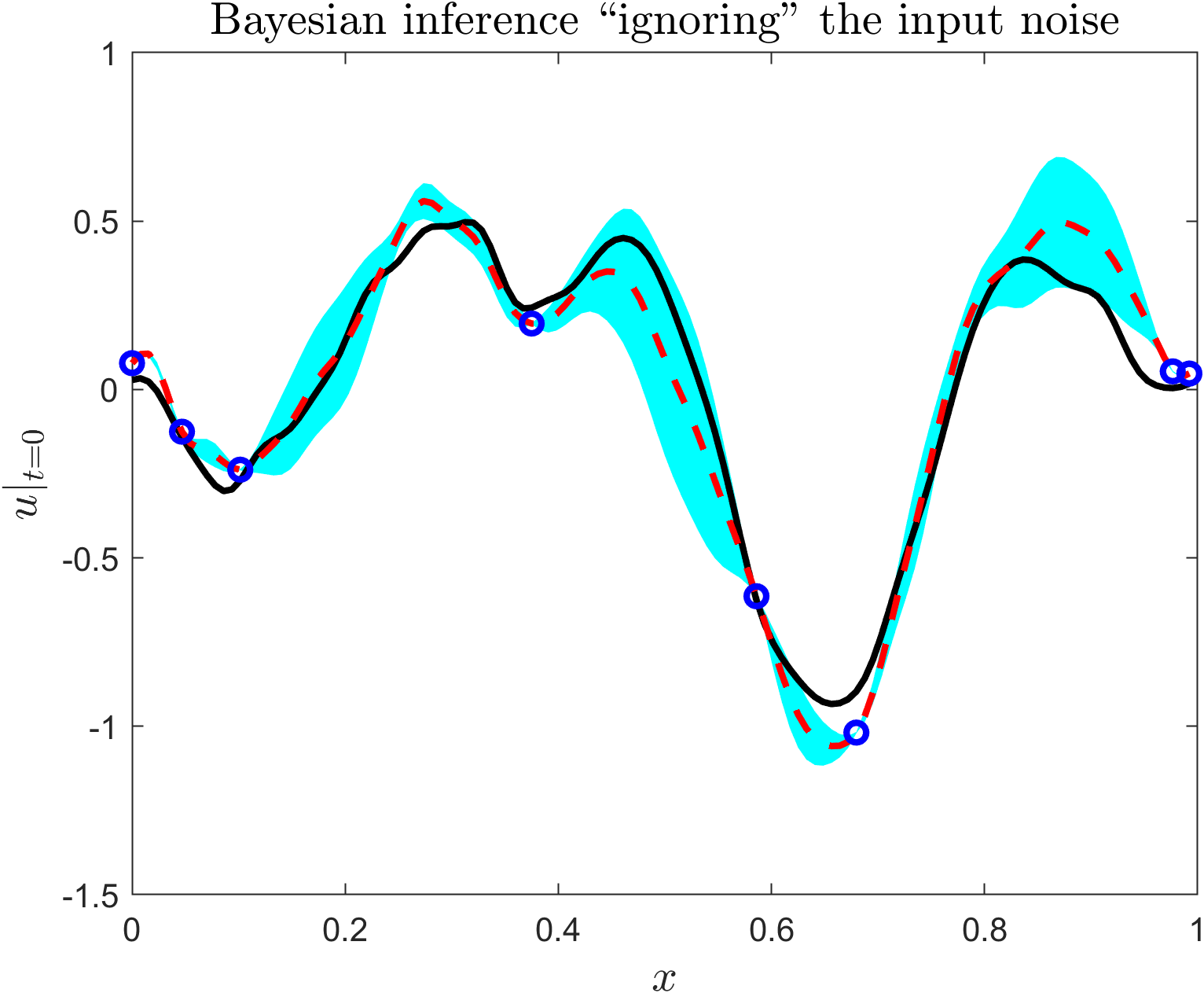}
    }
    \subfigure[Reconstruction of $u|_{t=1}$.]{
        \includegraphics[scale=.4]{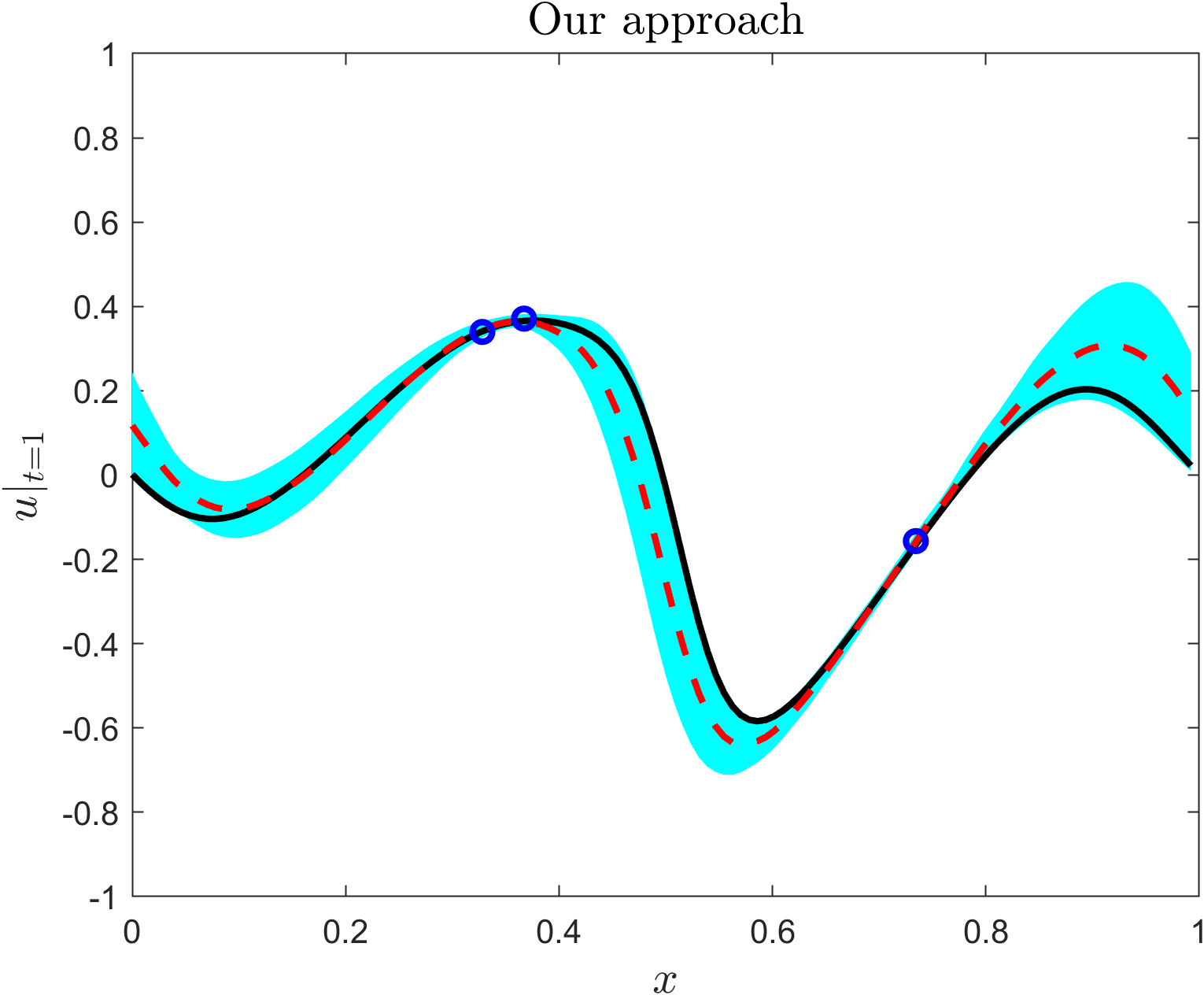}
        \includegraphics[scale=.4]{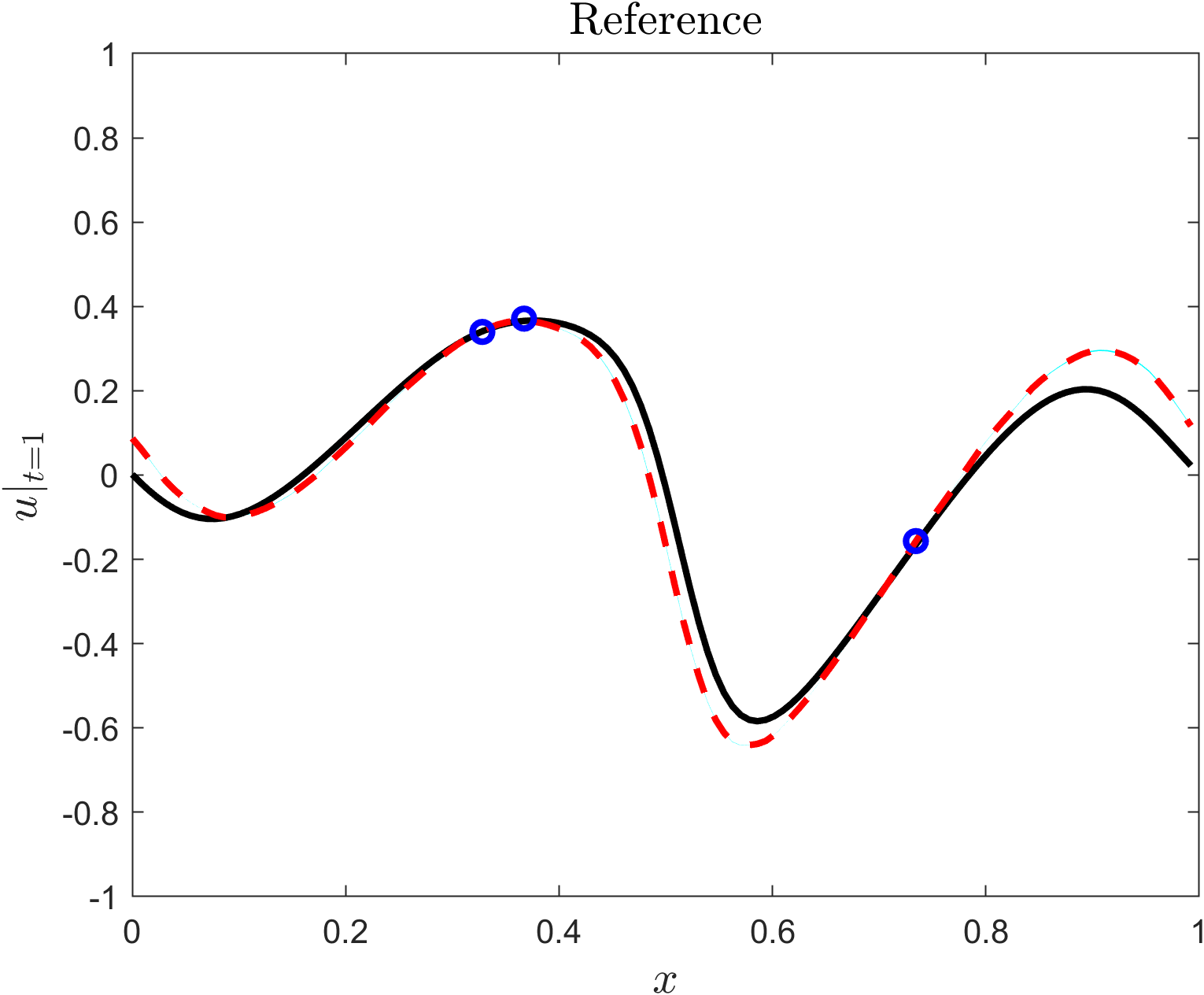}
        \includegraphics[scale=.4]{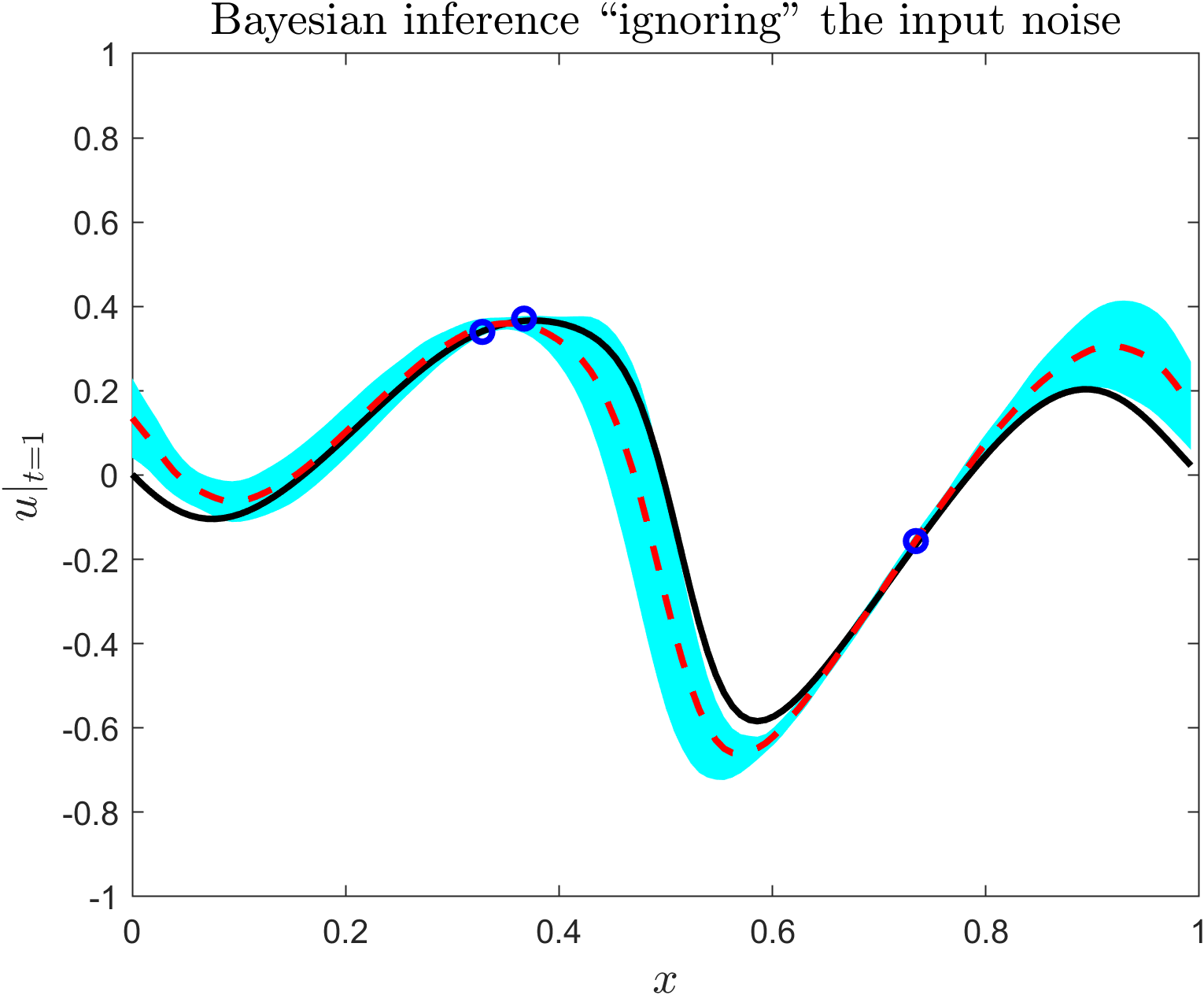}
    }
    \caption{Burgers equation: Reconstructing $u|_{t=0}$ and $u|_{t=1}$ from their noisy measurements and a pretrained FNO, which represents the physics and serves as an equation-free surrogate. From left to right are results from our approach with correctly specified input noise, the reference method, and our approach but misspecifying the input noise to a small value ($0.001$) in constructing the likelihood.
    Here, the reference method refers to the deterministic inference in which a maximum a posterior (MAP) estimate is performed without UQ.}
    \label{fig:example_2}
\end{figure}

\begin{figure}[ht!]
    \centering
    \subfigure[Reconstruction of $u|_{t=0}$.]{
        \includegraphics[scale=.3]{figs/example_2/f_hmc2.png}
        \includegraphics[scale=.3]{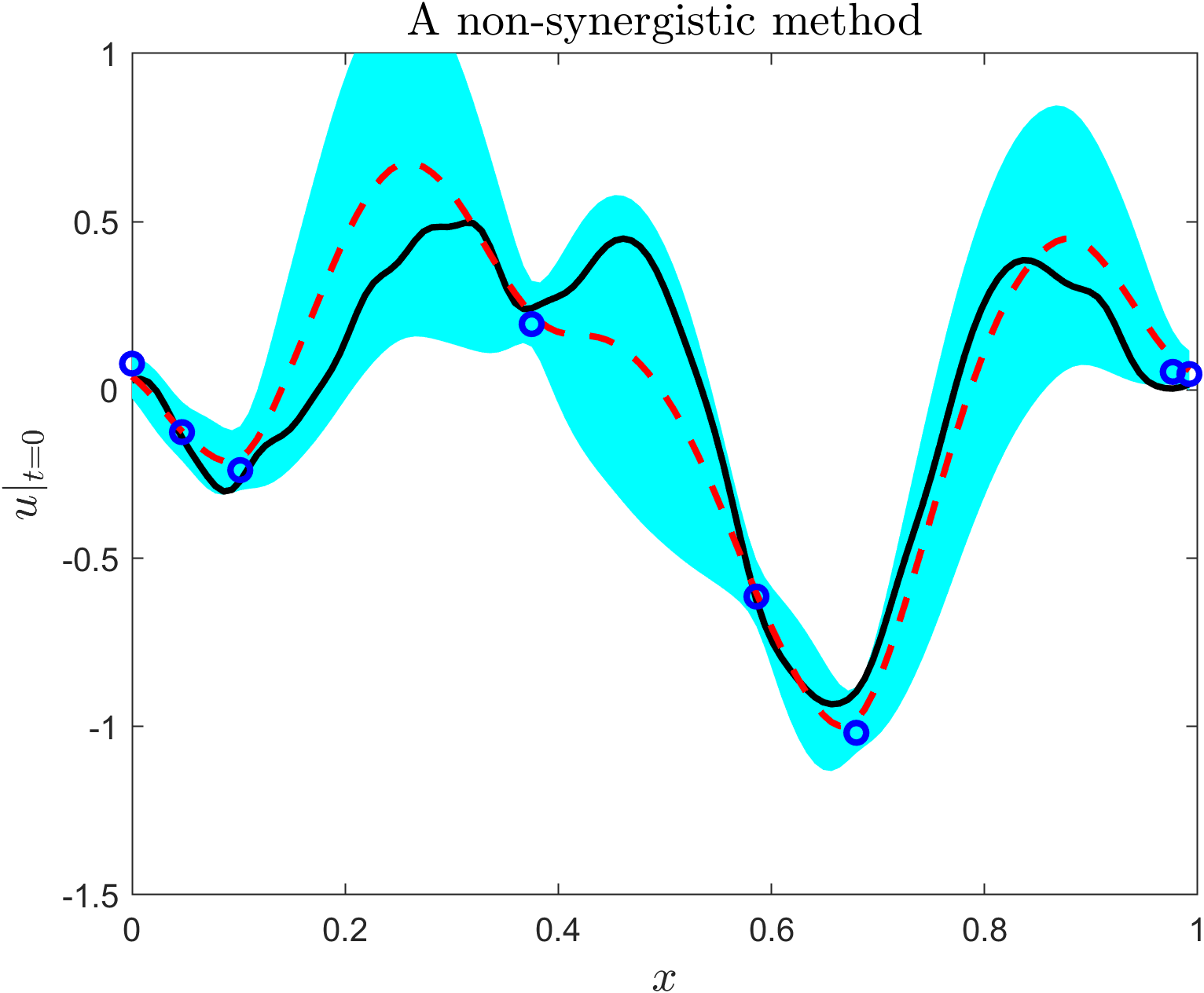}
    }
    \subfigure[Reconstruction of $u|_{t=1}$.]{
        \includegraphics[scale=.3]{figs/example_2/u_hmc2.png}
        \includegraphics[scale=.3]{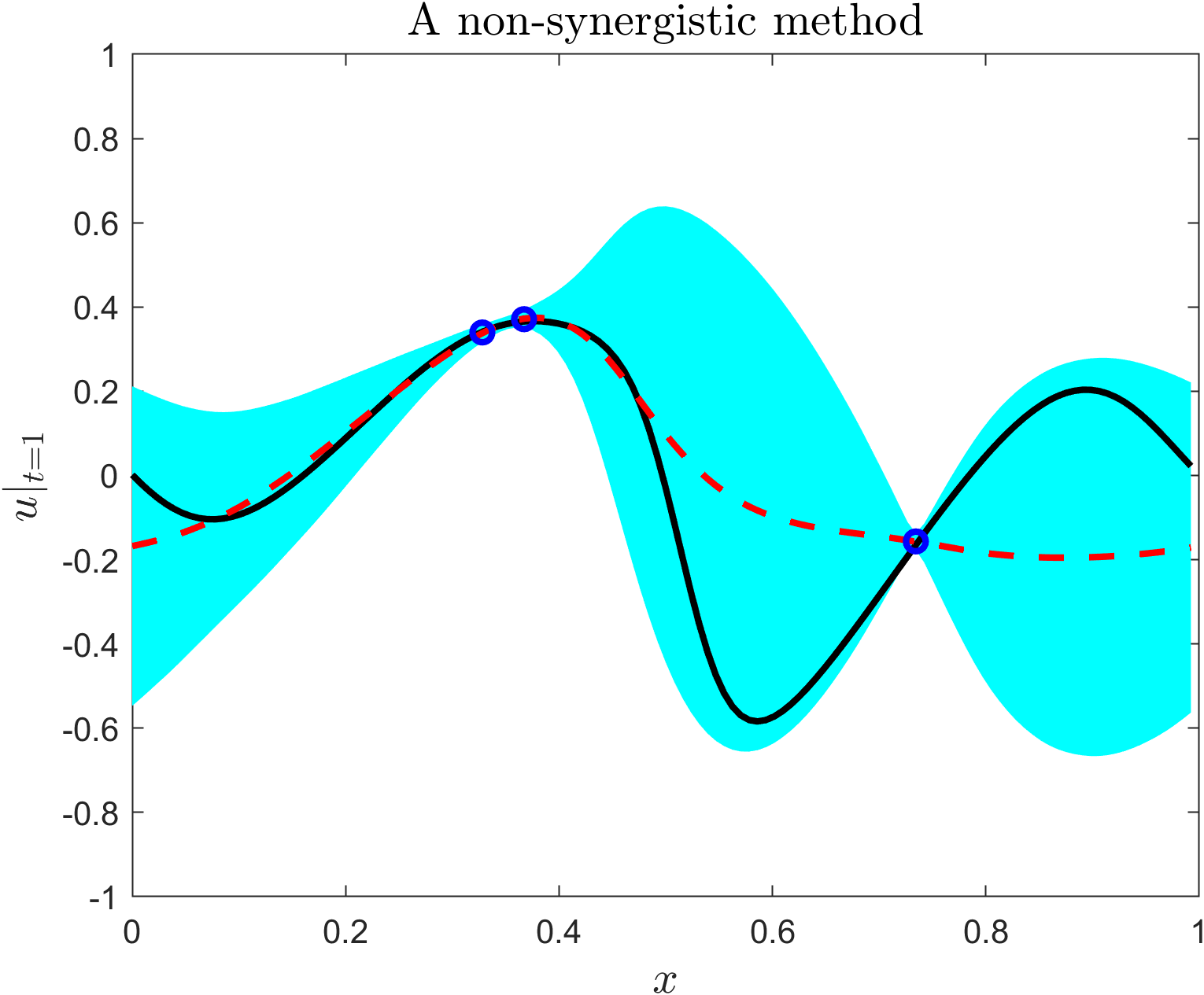}
    }
    \caption{Burgers equation: Comparison between our approach which reconstructs $u|_{t=0}$ and $u|_{t=1}$ synergistically and a non-synergistic method, in which these two functions are reconstructed independently from their respective sparse and noisy measurements without access to data of the other function.}
    \label{fig:example_2_2}
\end{figure}

We now employ the present method to quantify the uncertainty in the predictions of NOs with noisy inputs and outputs. Specifically, we consider the following 1-D Burgers equation with periodic boundary condition \cite{li2020fourier}:
\begin{equation}\label{eq:burgers}
\begin{aligned}
    & \frac{\partial u}{\partial t} + u \frac{\partial u}{\partial x}  = \nu \frac{\partial^2 u}{\partial x^2} , ~x\in [0, 1], ~t\in (0, 1],\\
    & u(x, t=0) = u_0, x \in [0, 1],
\end{aligned}
\end{equation}
where $\nu=0.1$ is the viscosity and $u_0$ is the initial condition. In this example, we first train an FNO to learn the solution operator from the initial condition $u|_{t=0}$ to $u|_{t=1}$ using clean and sufficient data. With the pretrained FNO, we assume that we have a downstream task in which there are available eight noisy measurements for $u|_{t=0}$ ($0.05$ noise scale) and three for $u|_{t=1}$ ($0.01$ noise scale), and we would like to reconstruct $u|_{t=0}$ and $u|_{t=1}$ from their sparse and noisy measurements with uncertainties.
We note that the measurements of $u|_{t=0}$ and $u|_{t=1}$ here are the noisy inputs and outputs, respectively.

Results of our approach are presented in the first column of Fig.~\ref{fig:example_2}, from which we observe that the reconstructions are generally accurate even with limited and noisy data and the errors between the references and the predicted means are bounded by the predicted uncertainties. 
In the second column of Fig.~\ref{fig:example_2}, we present results from reconstructing the input and output functions deterministically, which is achieved by performing a maximum a posteriori (MAP) estimate of the posterior in \eqref{eq:operator_posterior} with gradient descent. The comparison indicates the importance of UQ in such hybrid problems for trustworthy and reliable predictions. 
Another comparison is also made between ignoring and respecting the noise in the measurements of the input function. As shown in the third column of Fig.~\ref{fig:example_2}, misspecifying the scale of the input noise to a very small value ($0.001$) not only produces worse inference of the input function $u|_{t=0}$ but also impairs the inference of the output function $u|_{t=1}$, demonstrating the consequence of ignoring/misspecifying the noise in the input when solving the hybrid problem with pretrained NOs.

In this example, we further compare our approach with a non-synergistic learning method for reconstructing $u|_{t=0}$ and $u|_{t=1}$, in which these two functions are reconstructed independently from their respective sparse and noisy measurements without access to information of the other function.
Specifically, the posterior for the discretiziation of the input function becomes $p(\mathbf{v}_\psi|\mathcal{D}_v) \propto p(\mathbf{v}_\psi) p (\mathcal{D}_v|\mathbf{v}_\psi)$ and no pretrained FNO is employed, indicating that the physics is not utilized; see \eqref{eq:operator_likelihood_1}.
Meanwhile the posterior for inferring the output function reads $p(\mathbf{v}_\psi|\mathcal{D}_u) \propto p(\mathbf{v}_\psi) p (\mathcal{D}_u|\mathbf{v}_\psi)$ where the pretrained FNO is employed in the likelihood $p (\mathcal{D}_u|\mathbf{v}_\psi)$; see \eqref{eq:operator_likelihood_2}.
Results of this comparison are presented in Fig.~\ref{fig:example_2_2}, from which we observe significant increase in both the errors and the predicted uncertainties in reconstructing the input and output functions of the FNO. 
Although the non-synergistic method also provides reliable predictions in which the predicted uncertainties bound the errors between the predicted means and the references, its decoupling of the learning of $u|_{t=0}$ and $u|_{t=1}$ hinders the potential for synergistic learning. Conversely, in our approach a pretrained FNO is employed to represent the physics, bridging these two functions with the underlying physics. As a result, the reconstructions benefit from the synergy of information from the physics and data of both functions, leading to more accurate and confident predictions. 
The differences in predicted uncertainties between these two approaches showcase synergistic learning realized by our approach.
\subsection{1D time-dependent reaction-diffusion equation with multi-input DeepONets}\label{sec:example_3}

\begin{figure}[ht]
    \centering
    \subfigure[Reconstruction of $u$.]{
        \includegraphics[scale=.325]{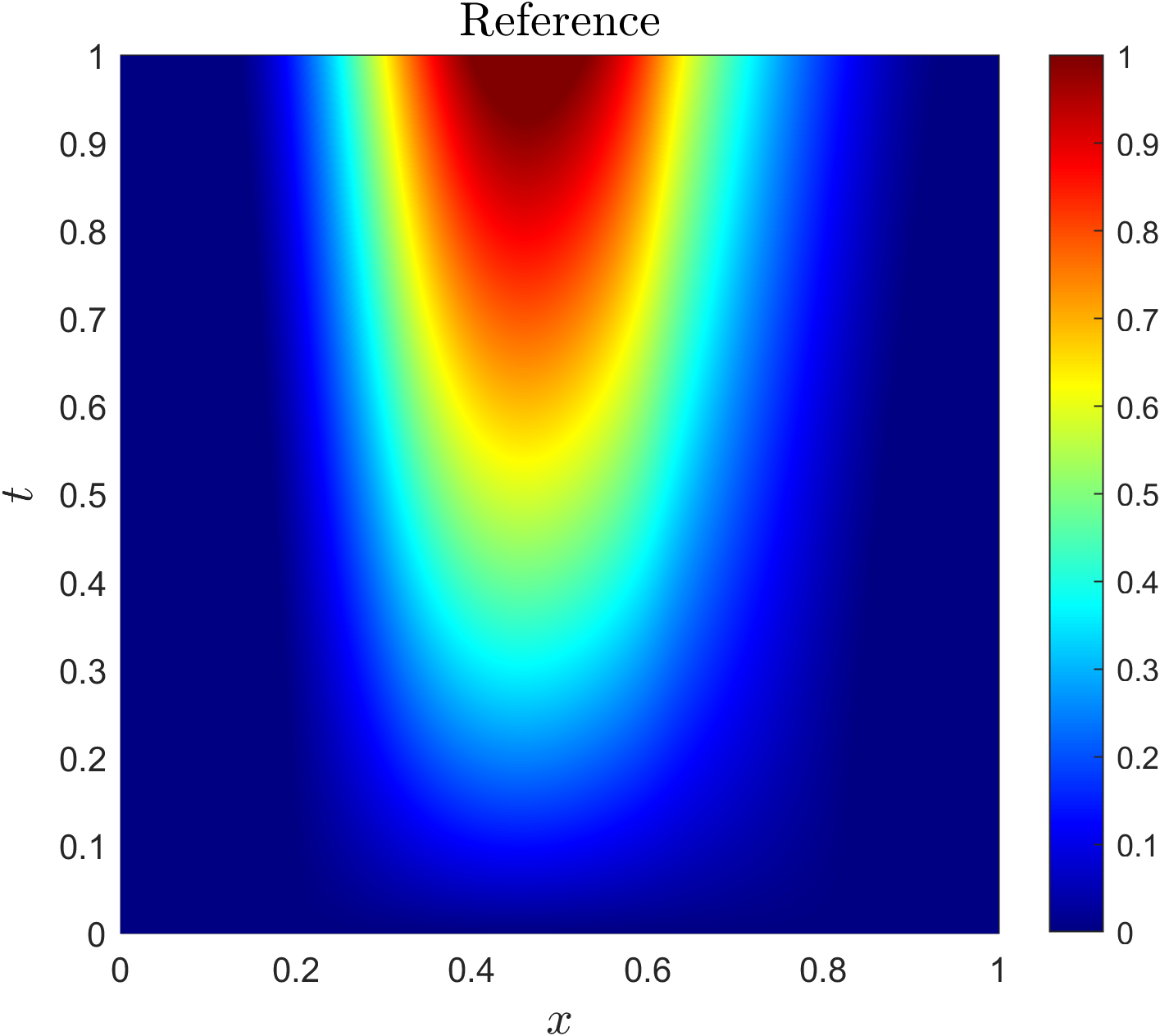}
        \includegraphics[scale=.325]{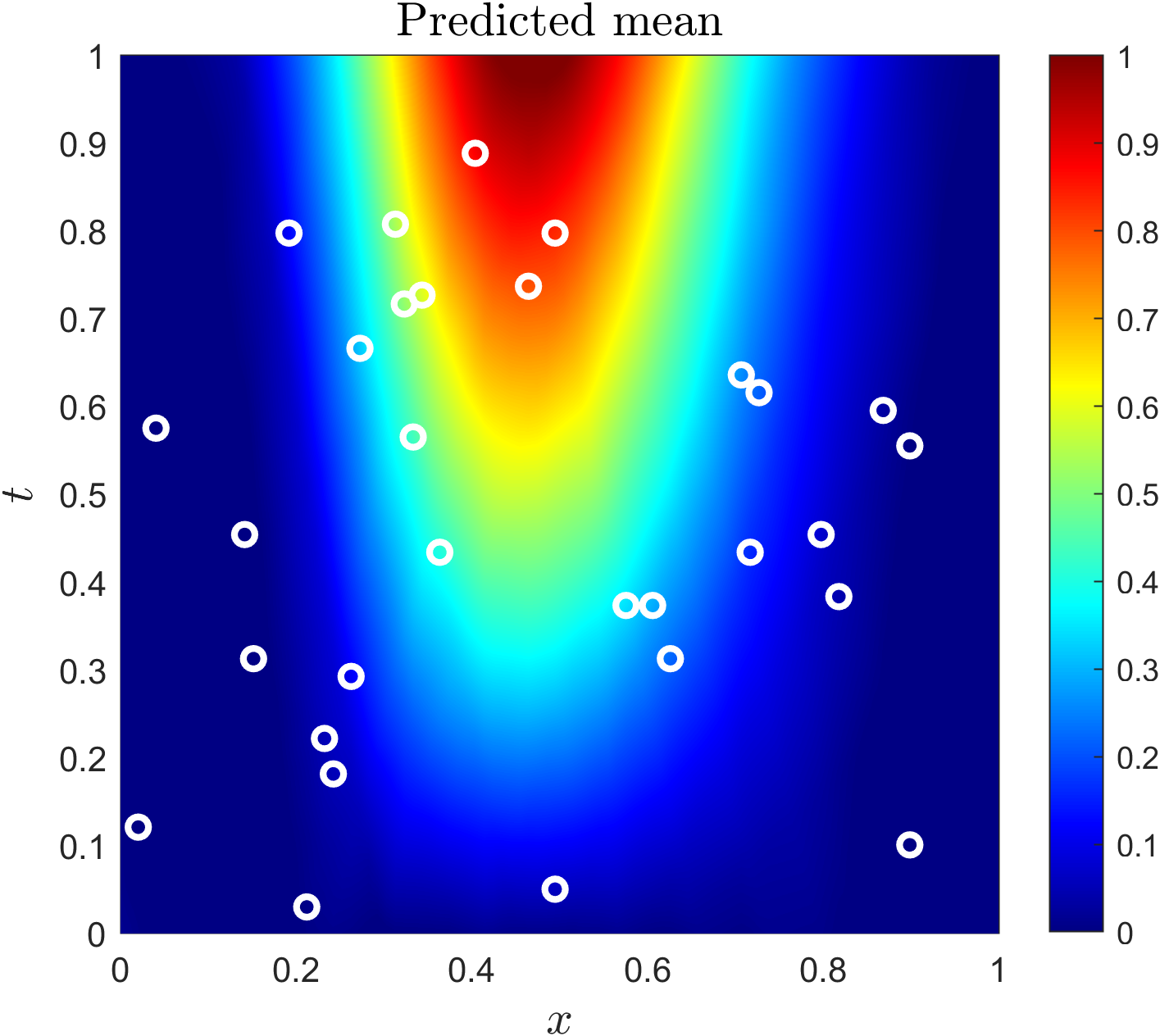}
        \includegraphics[scale=.325]{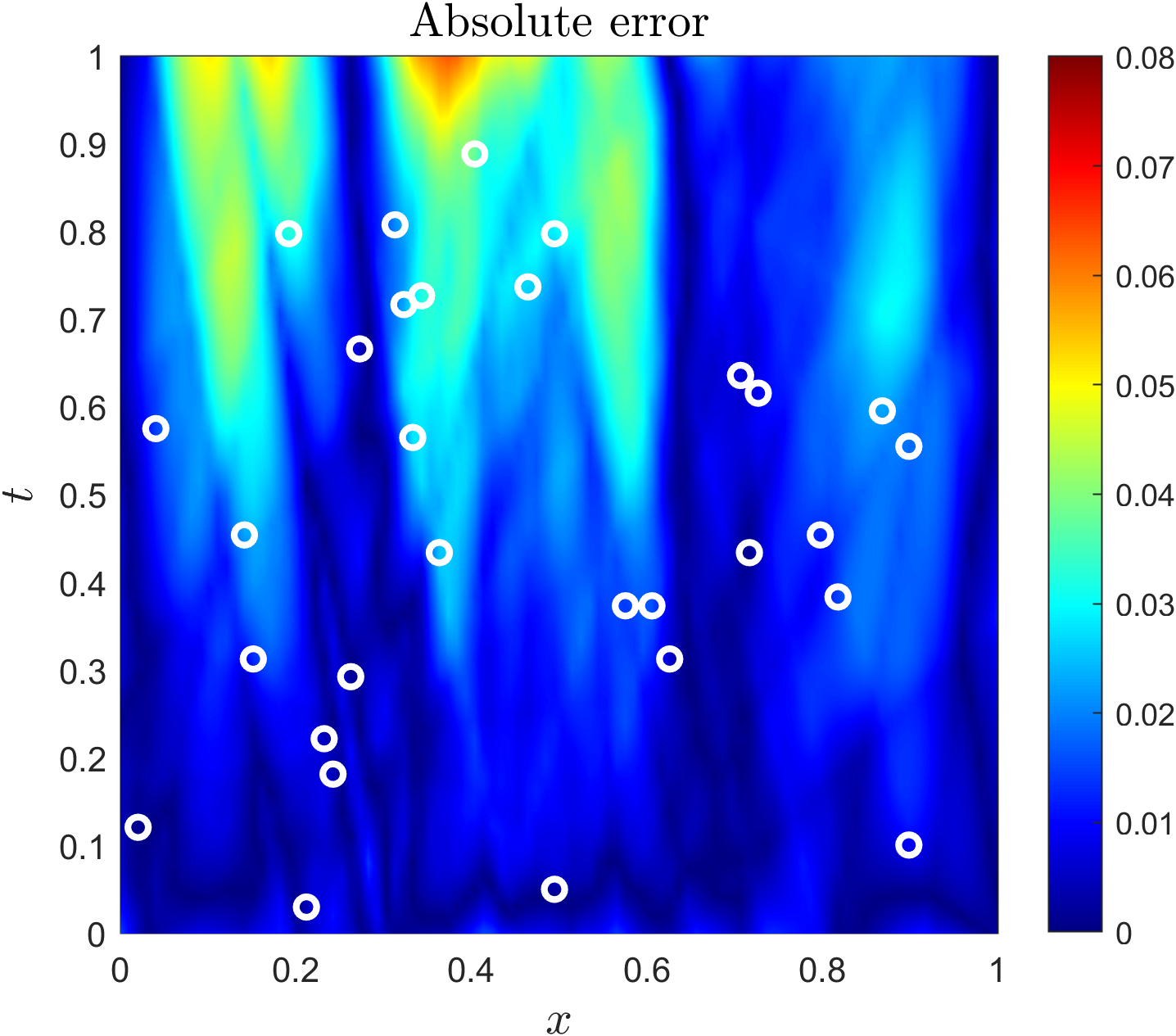}
        \includegraphics[scale=.325]{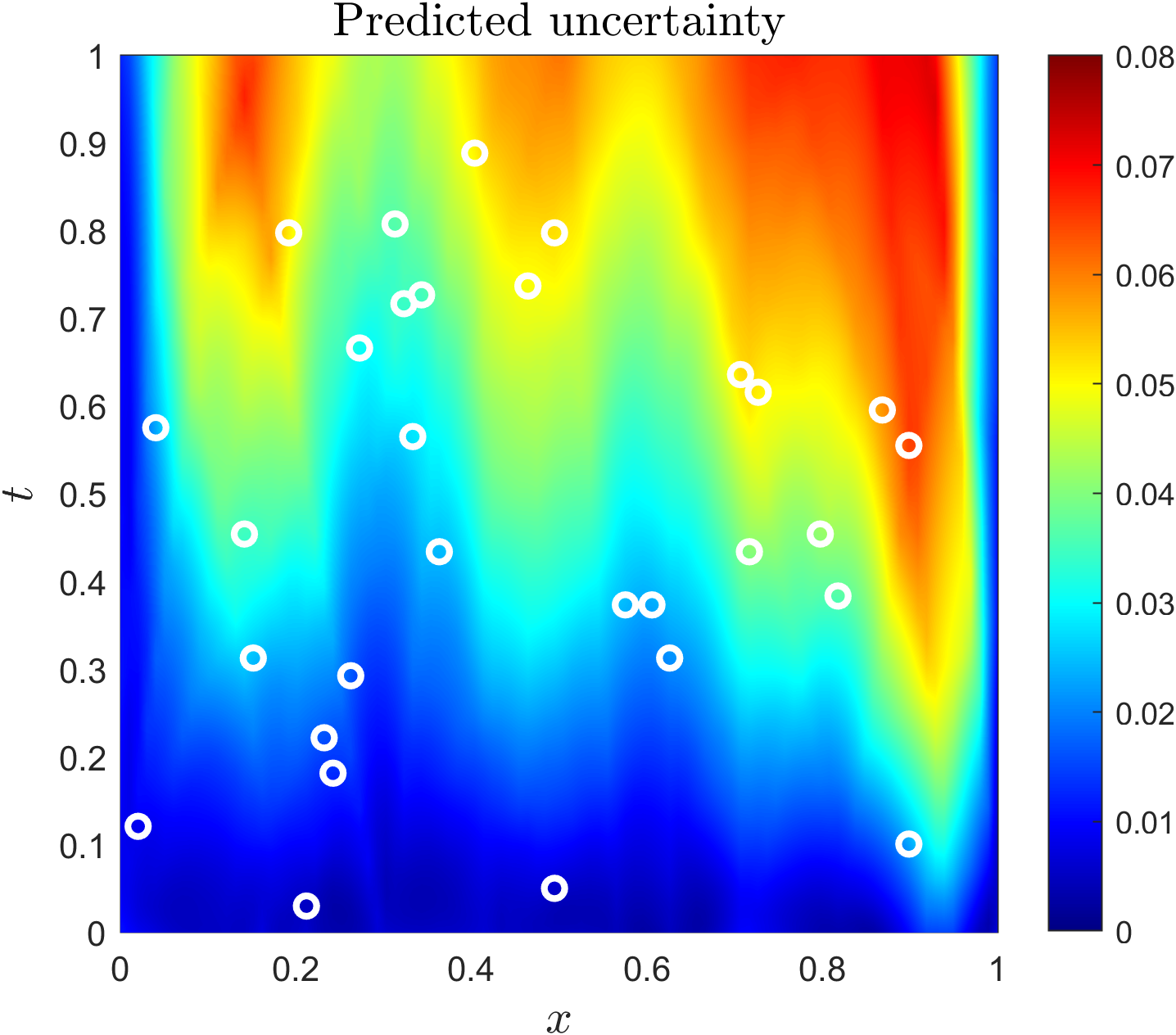}
    }
    \subfigure[Reconstruction of $k, f$.]{
        \includegraphics[scale=.3]{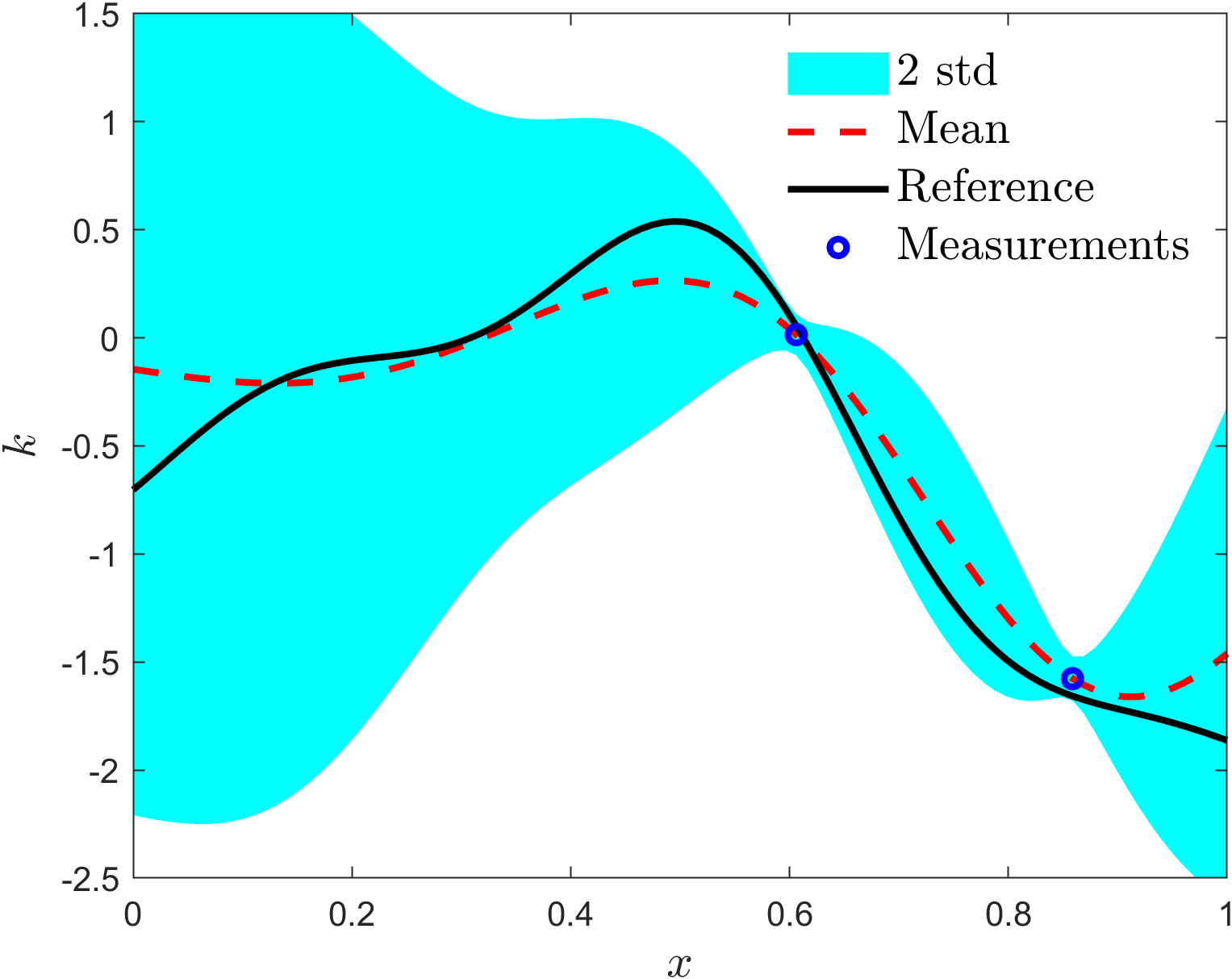}
        \includegraphics[scale=.3]{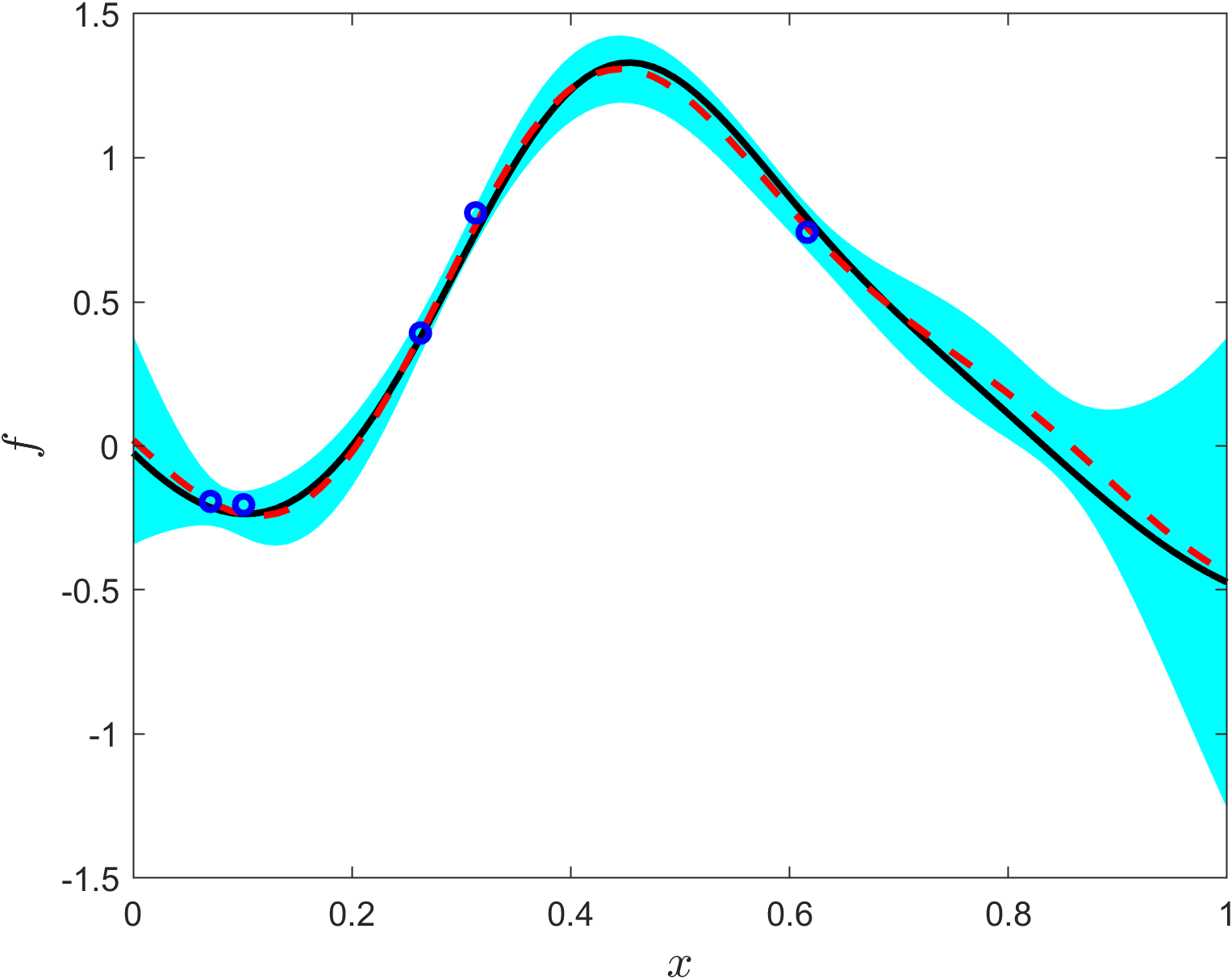}
    }
    \subfigure[A non-synergistic learning method.]{
        \includegraphics[scale=.3]{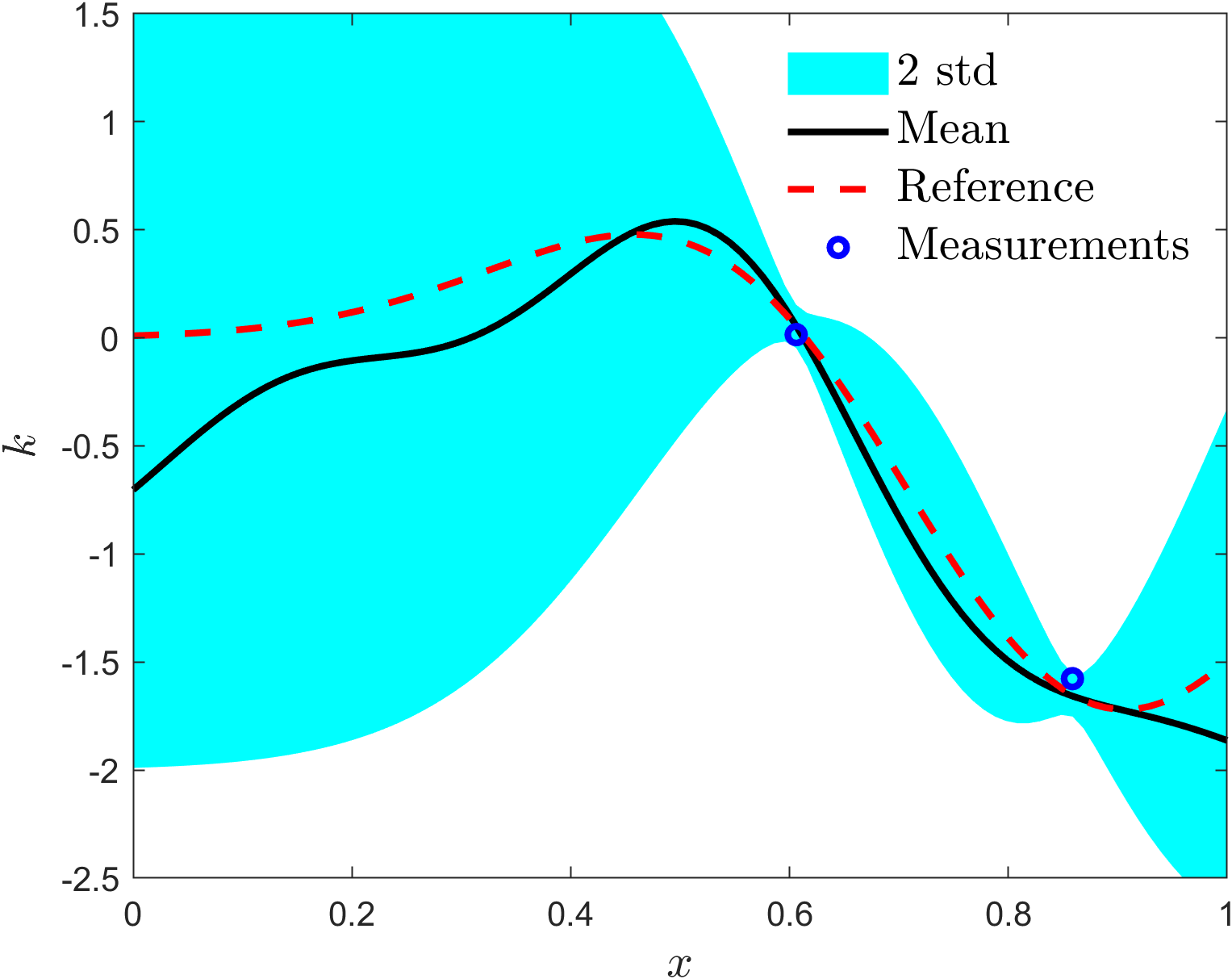}
        \includegraphics[scale=.3]{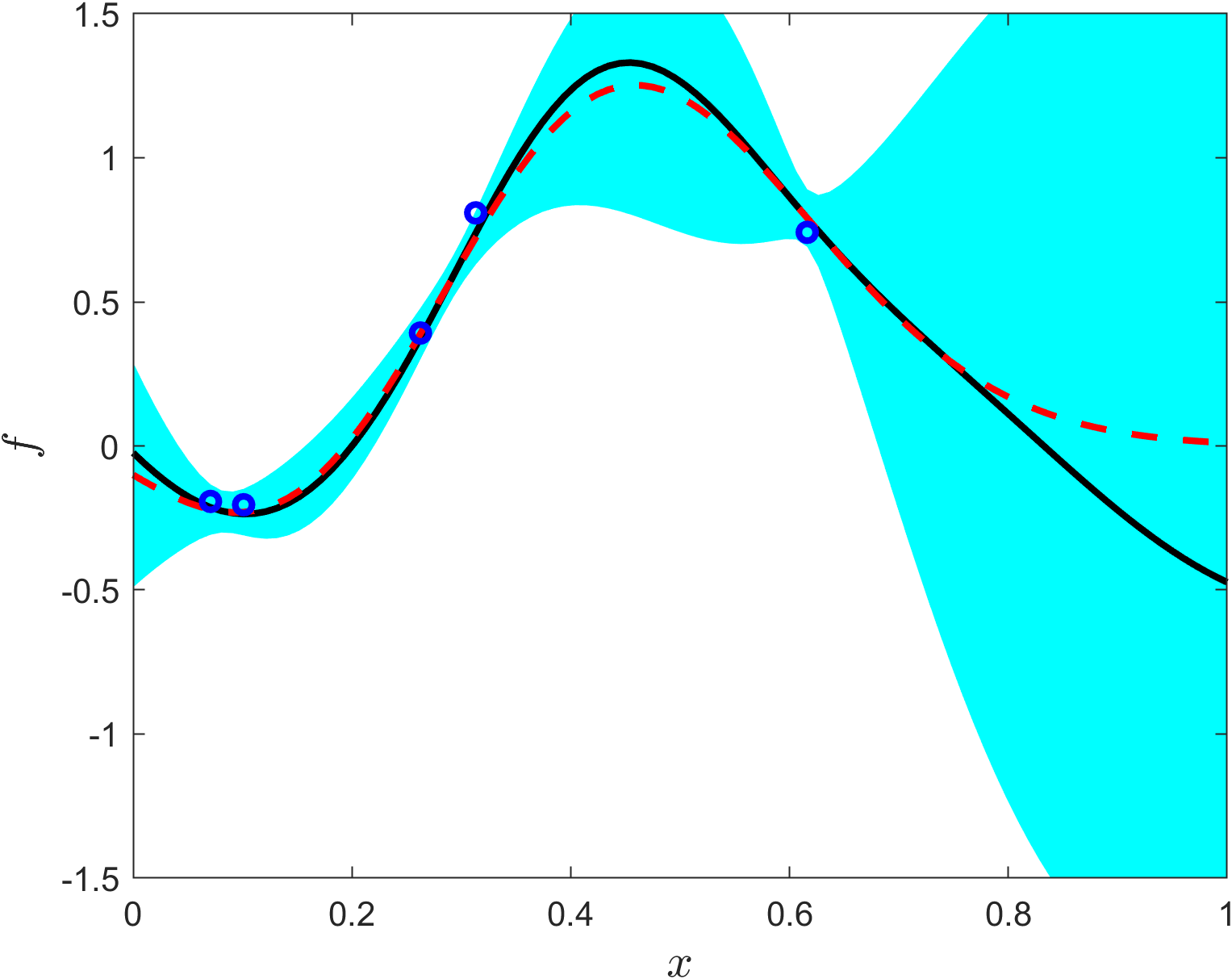}
    }
    \caption{1D time-dependent reaction-diffusion equation with heteroscedastic diffusion term: Reconstructions with uncertainties of $u, k, f$, as displayed in (a) and (b), from their sparse and noisy measurements and a pretrained multi-input DeepONet, which represents the physics in \eqref{eq:reaction_diffusion}. White circles represent locations of measurements of $u$.  In (c), a non-synergistic method is employed to reconstruct $k$ and $f$ independently from their respective measurements without access to data of the other function or data of $u$.}
    \label{fig:example_3_2}
\end{figure}

In this example, we consider the following reaction-diffusion PDE (adapted from \cite{jin2022mionet}):
\begin{equation}\label{eq:reaction_diffusion}
    \frac{\partial u}{\partial t} = \frac{\partial }{\partial x} (0.01 (|k(x)| + 1) \frac{\partial u}{\partial x}) + \kappa u^2 + f(x), x\in [0, 1], t\in[0, 1],
\end{equation}
with zero initial and boundary conditions, where $\kappa=0.01$, and $k(x)$ and $f(x)$ are space-dependent terms, representing the diffusion and the source term, respectively. 
Similar as in Sec.~\ref{sec:example_2}, we train an NO with clean and sufficient data to learn the solution operator from $k(x)$ and $f(x)$ to $u(x, t)$. 
We then would like to reconstruct $k, f, u$ from their sparse and noisy measurements. In particular, a multi-input DeepONet \cite{jin2022mionet} (see Fig.~\ref{fig:2} for the architecture of the NO) is employed to learn the solution operator, represent the physics and therefore bridge $k, f$ and $u$.
We then assume we have two random measurements of $k$, five random measurements of $f$, and $30$ random measurements of $u$, and measurements are corrupted by the additive Gaussian noise with scale $0.05$. 
By the independence in observing data and the independence between priors of $k$ and $f$, the posterior in this case reads:
\begin{equation}
    p(\mathbf{k}_\psi, \mathbf{f}_\psi | \mathcal{D}_f, \mathcal{D}_k, \mathcal{D}_u) \propto  p(\mathcal{D}_f| \mathbf{f}_\psi) p(\mathcal{D}_k| \mathbf{k}_\psi) p(\mathcal{D}_u| \mathbf{k}_\psi, \mathbf{f}_\psi) p(\mathbf{k}_\psi)p(\mathbf{f}_\psi),
\end{equation}
where $\mathbf{k}_\psi$ and $\mathbf{f}_\psi$ denote the discretization of $k$ and $f$, respectively.
A simpler case can be found in \ref{sec:appendix_additional}, where the diffusion term is a known constant and the task becomes reconstructing $f$ and $u|_{t=1}$ from their sparse and noisy measurements.

Results are displayed in Fig.~\ref{fig:example_3_2}(a-b), from which we can see all three functions are reconstructed accurately with reasonable predicted uncertainties that bound the errors between the predicted means and the references. 
Both the errors and the predicted uncertainties grow as one moves away from the region where data are available.
In Fig.~\ref{fig:example_3_2}(c) we present results from a synergistic method in inferring $k$ and $f$. Similar as in Sec.~\ref{sec:example_2}, the non-synergistic method reconstructs these two functions independently from their respective measurements without access to data of the other function, data of $u$, or the pretrained multi-input DeepONet.
Specifically, the posterior for the discretiziation of $k$ becomes $p(\mathbf{k}_\psi|\mathcal{D}_k) \propto p(\mathbf{k}_\psi) p (\mathcal{D}_k|\mathbf{k}_\psi)$ and the posterior for the discretization of $f$ is $p(\mathbf{f}_\psi|\mathcal{D}_f) \propto p(\mathbf{f}_\psi) p (\mathcal{D}_f|\mathbf{f}_\psi)$.
By comparison, we observe significant decrease in the errors and predicted uncertainties when our approach is employed, demonstrating the effectiveness of the proposed approach and the pretrained multi-input DeepONet in representing the underlying physics. The errors and predicted uncertainties from our approach do not develop as fast as the non-synergistic method as one moves away from the region where data are available, because information of $k$, $f$, and $u$ are coupled by the pretrained NO.

\subsection{120-dimensional Darcy problem with DeepONets}\label{sec:example_4}

\begin{figure}[ht]
    \centering
    \subfigure[Reconstruction of $\log(\lambda)$.]{
        \includegraphics[scale=.32]{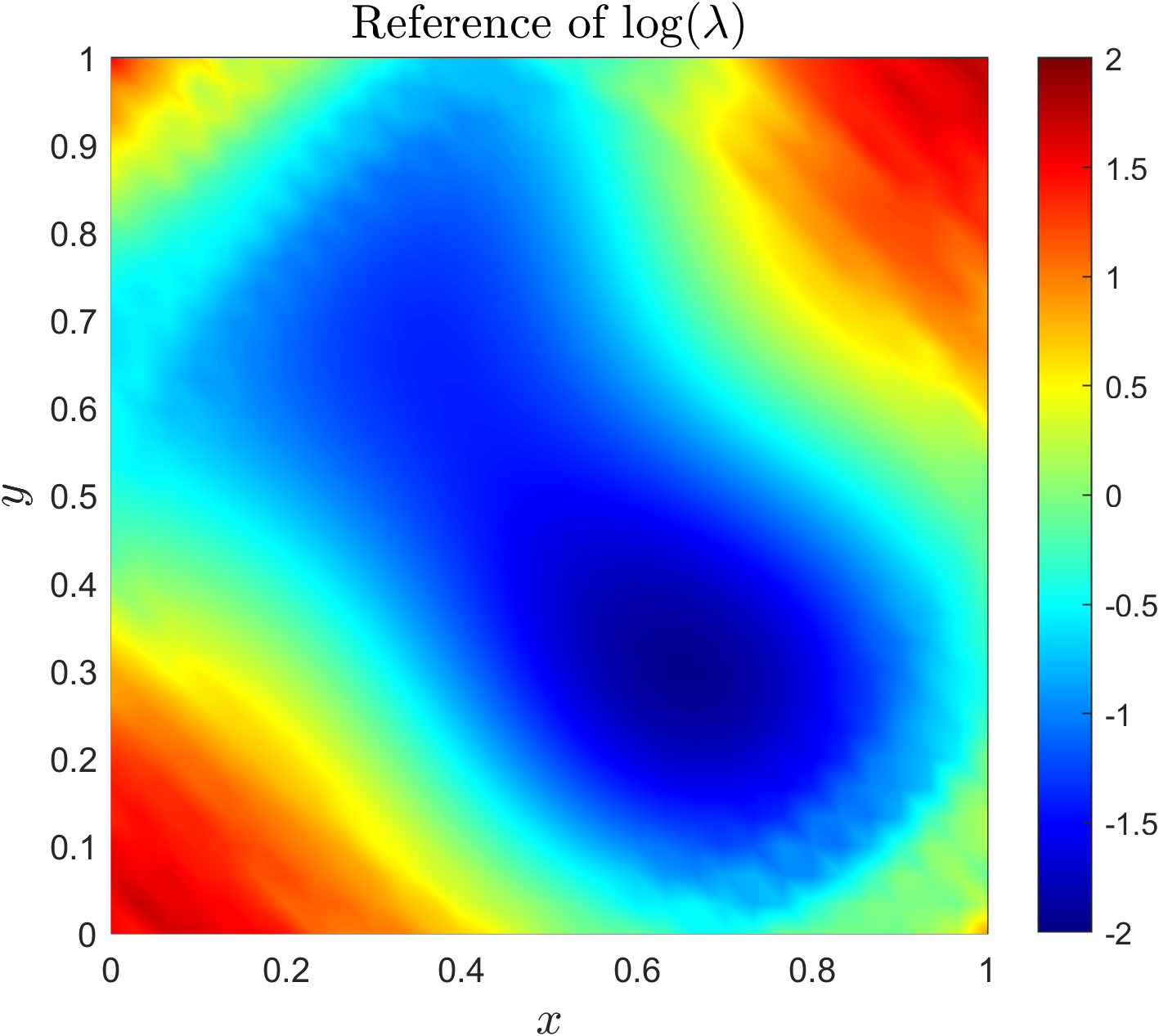}
        \includegraphics[scale=.32]{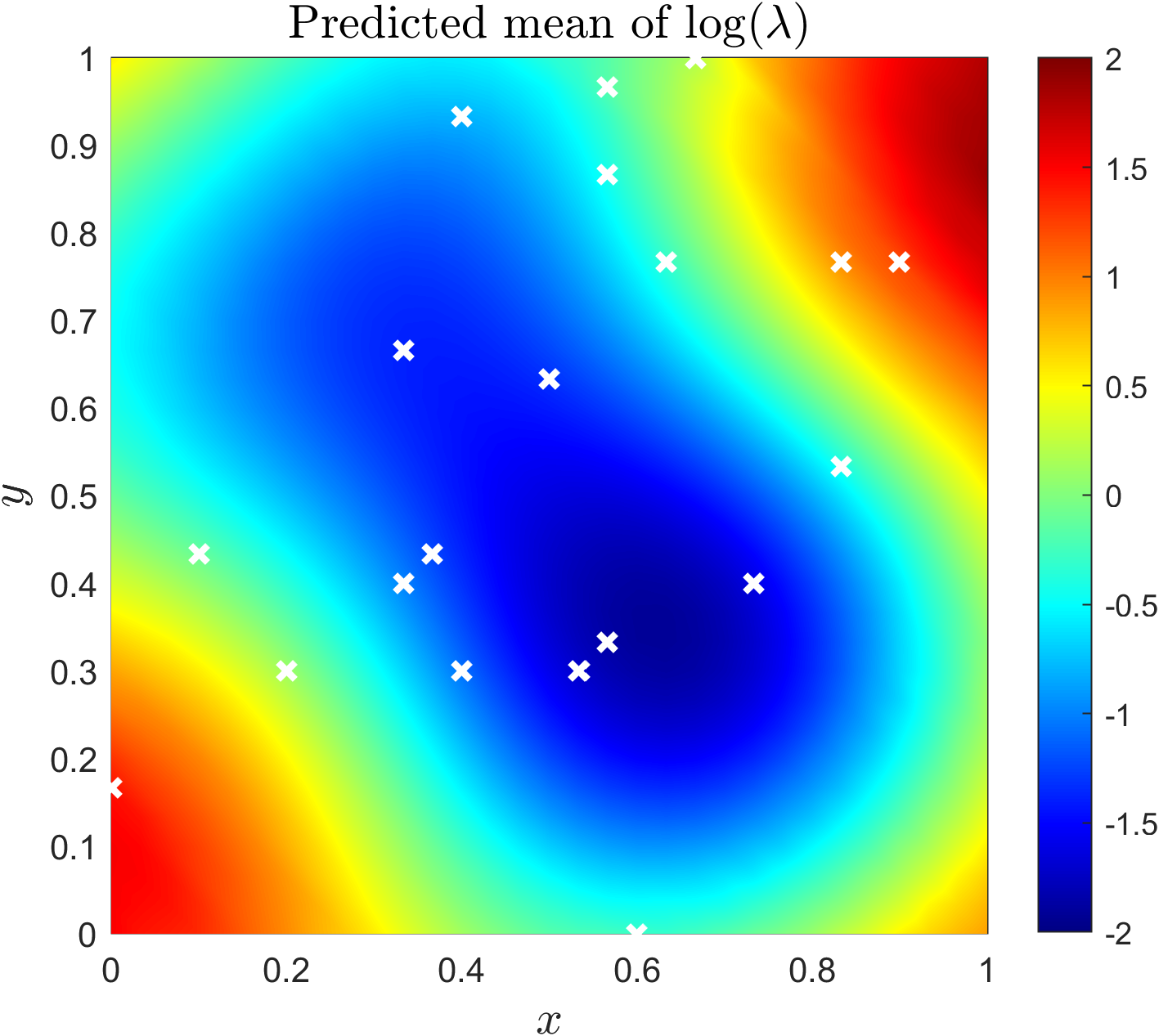}
        \includegraphics[scale=.32]{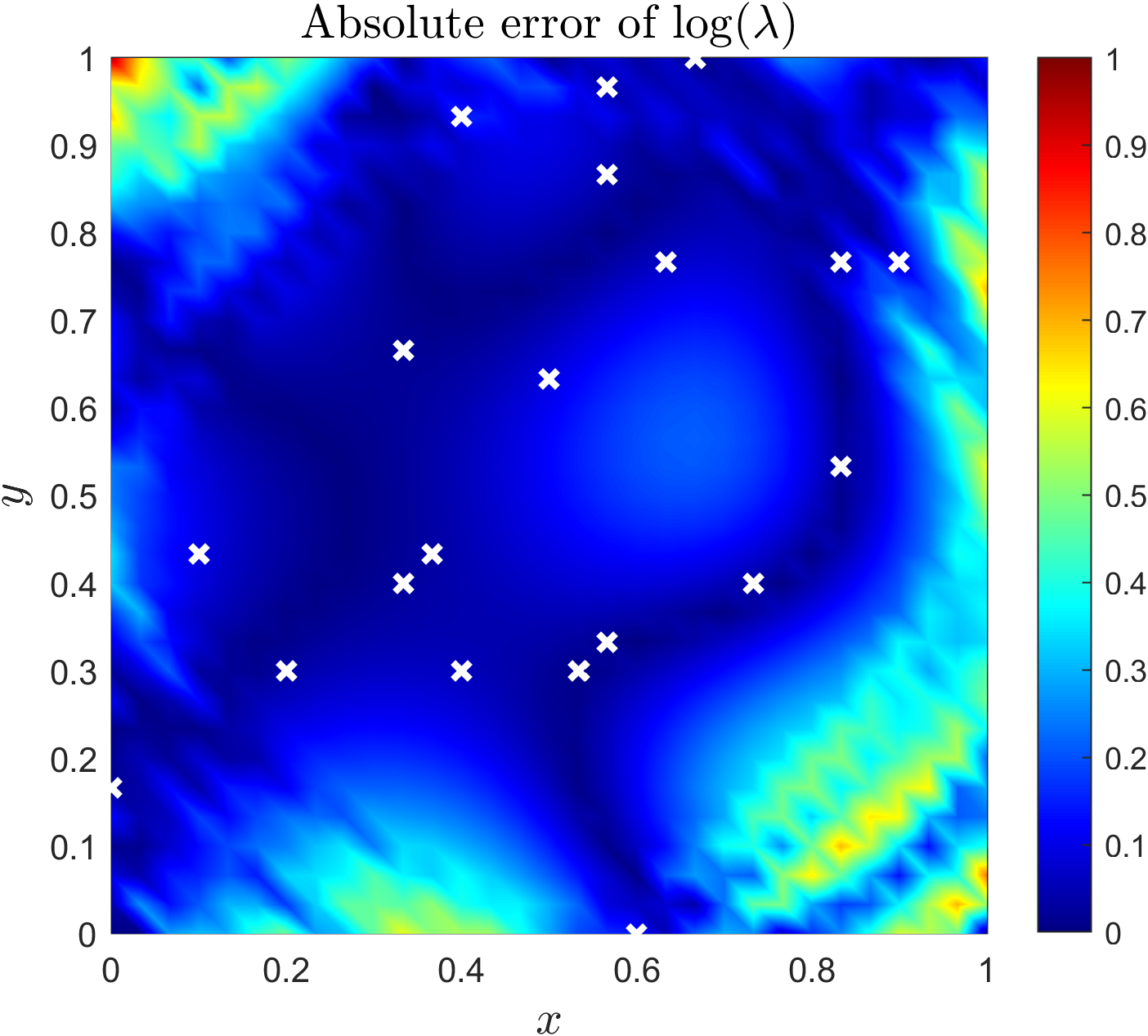}
        \includegraphics[scale=.32]{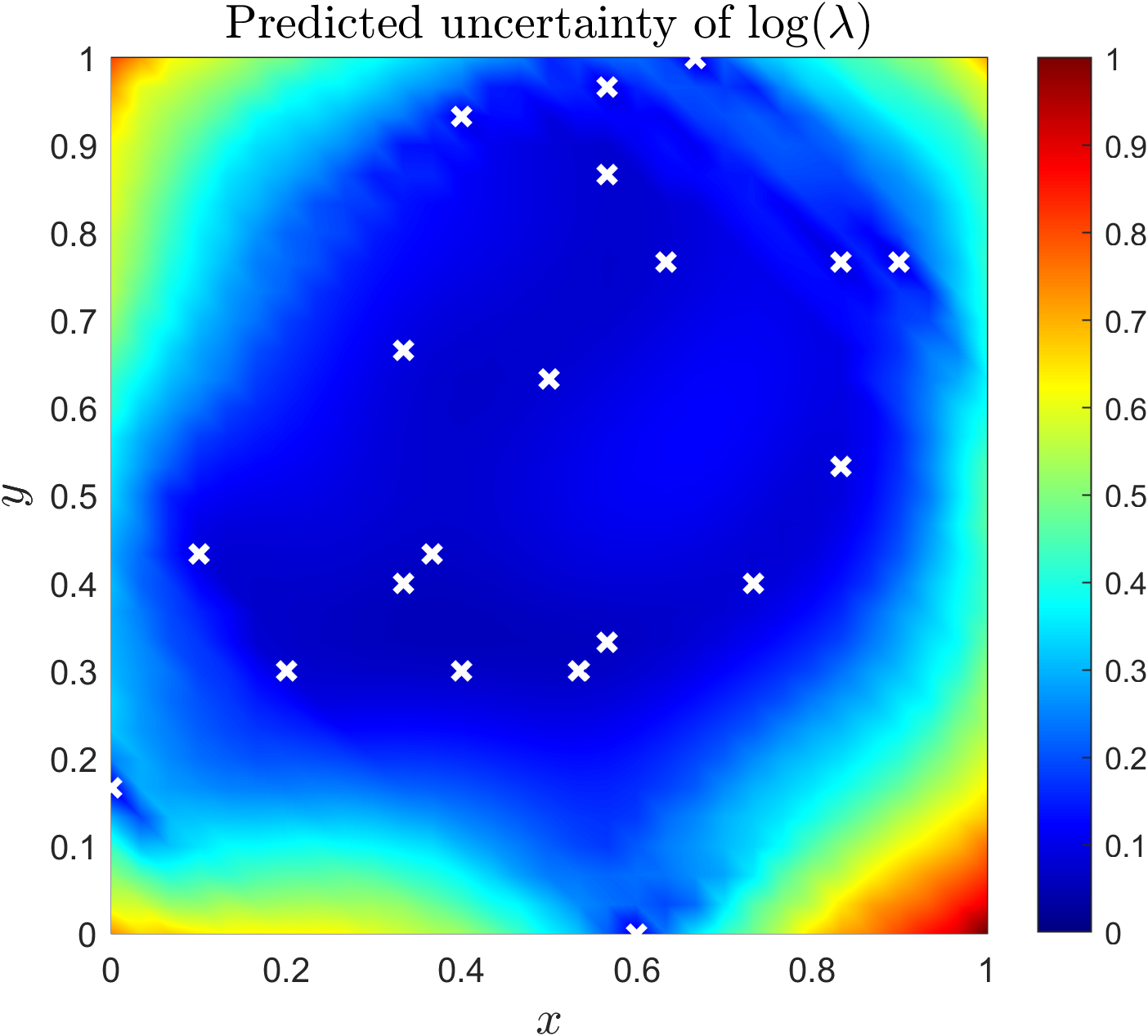}
    }
    \centering
    \subfigure[Reconstruction of $u$.]{
        \includegraphics[scale=.32]{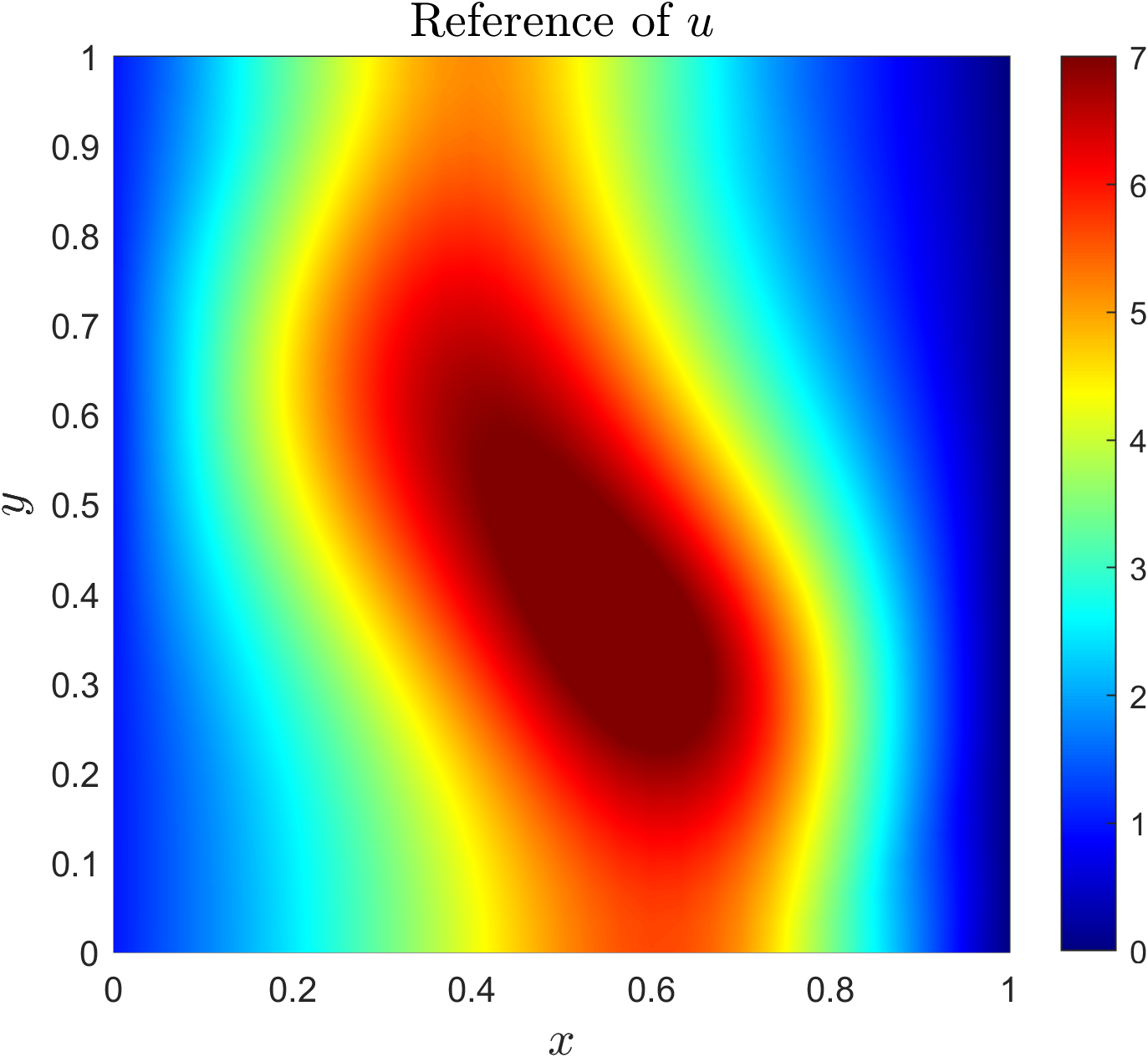}
        \includegraphics[scale=.32]{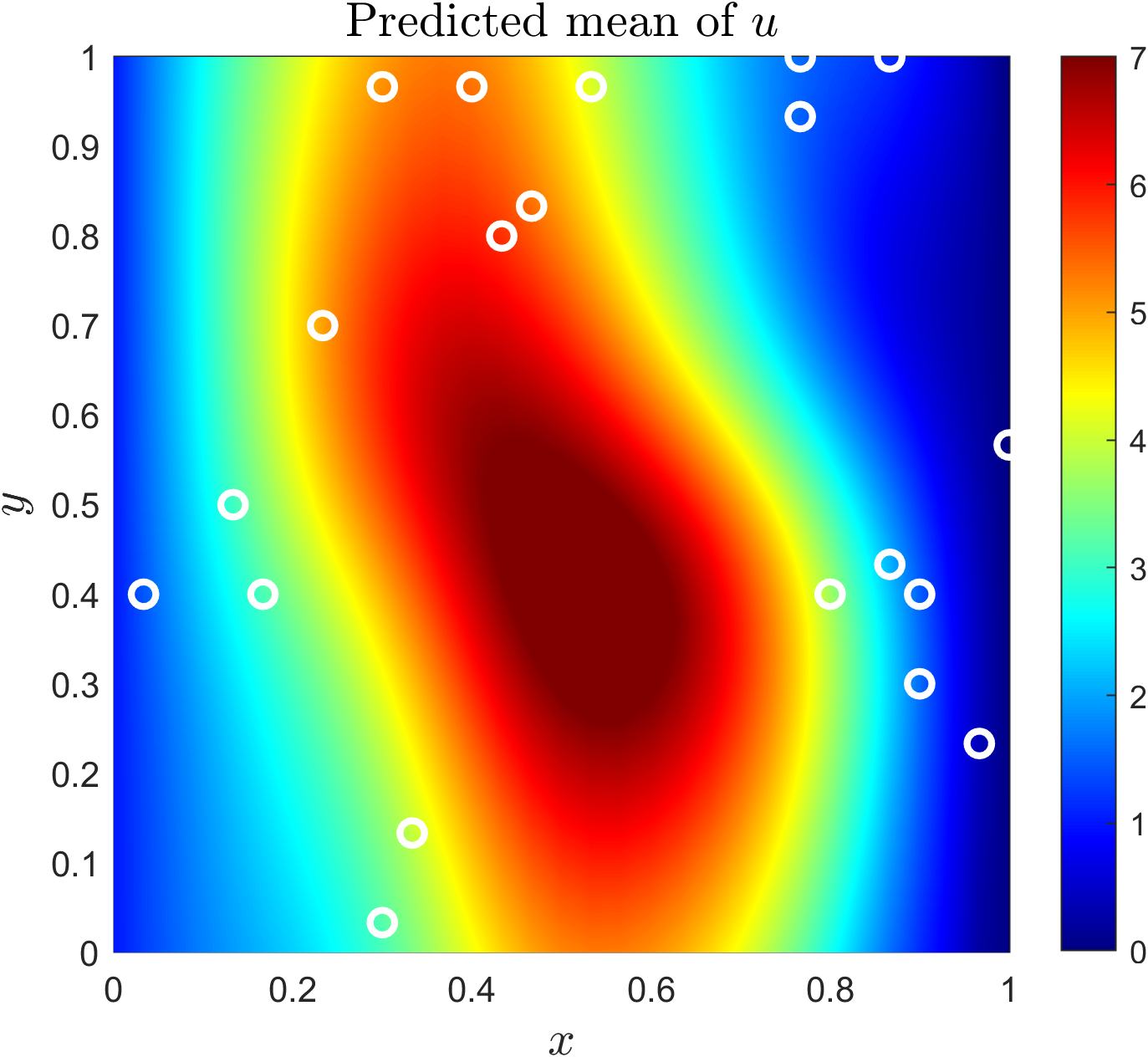}
        \includegraphics[scale=.32]{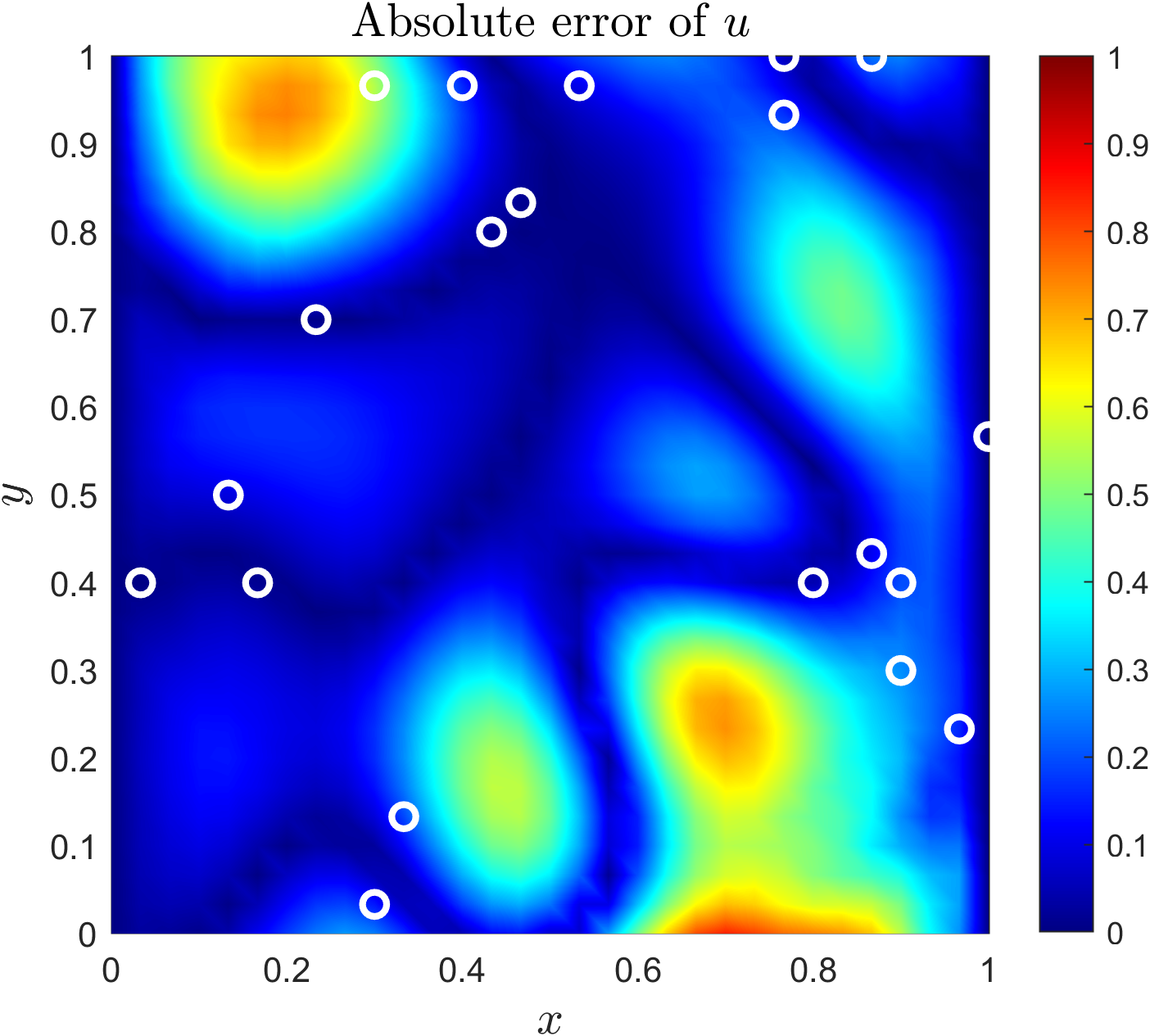}
        \includegraphics[scale=.32]{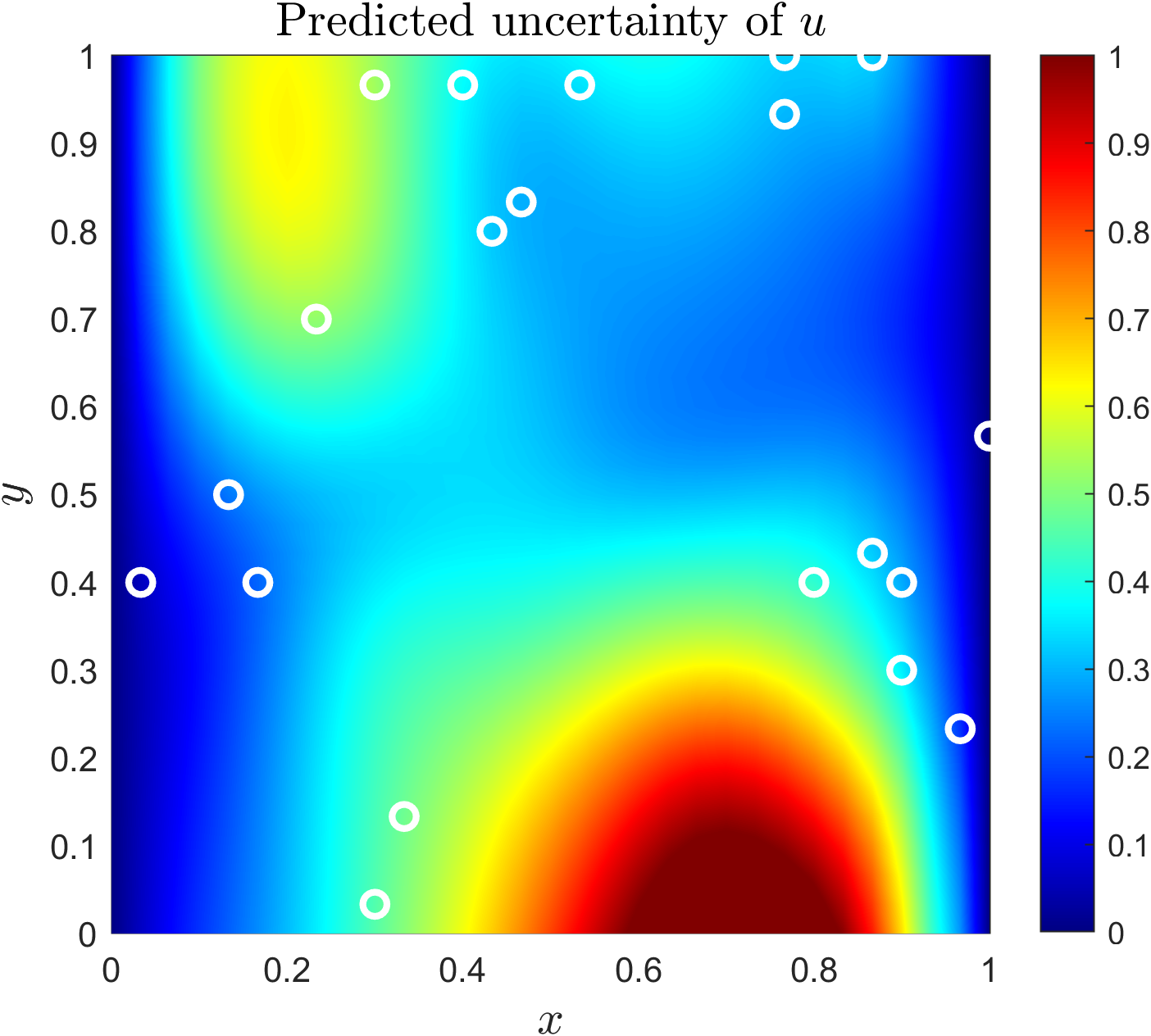}
    }
    \caption{120-dimensional Darcy problem: Reconstructions with uncertainties of $\log(\lambda)$ (shown in (a)) and $u$ (shown in (b)) from their sparse and noisy measurements as well as a pretrained DeepONet representing the physics in \eqref{eq:darcy}. The white cross stands for the location where $\lambda$ is observed while the white circle stands for the location where $u$ is observed.}
    \label{fig:example_4}
\end{figure}

We consider the Darcy's law describing a steady flow through porous media in two dimensions \cite{psaros2023uncertainty, zou2022neuraluq}:
\begin{equation}\label{eq:darcy}
    \nabla \cdot (\lambda(x, y) \nabla u(x, y)) = f, (x, y) \in (0, 1)^2,
\end{equation}
where $\lambda$ denotes the hydraulic conductivity field, $u$ denotes the hydraulic head, and $f = -40$ is a constant. The boundary conditions are:
\begin{subequations}\label{eq:darcy_bcs}
\begin{align}
    u(0, y) = 1, ~ u(1, y) = 0, ~y\in(0, 1)\\
    \partial_{\boldsymbol{n}} u(x, 0) = \partial_{\boldsymbol{n}} u(x, 1) = 0, ~x\in(0, 1),
\end{align}
\end{subequations}
where $\boldsymbol{n}$ denotes the unit normal vector of the boundary.

In this study, we use the following model to describe the hydraulic conductivity field, i.e. $\lambda(x, y) = \exp(k(x, y))$, where $k(x, y)$ is a sample of a truncated Karhunen-Lo\`eve expansion of a Gaussian process with zero mean and the following exponential squared kernel:
\begin{equation}
    k(x_1, y_1, x_2, y_2) = \exp(-\frac{(x_1-x_2)^2}{2l^2}-\frac{(y_1-y_2)^2}{2l^2}), ~x_1,x_2,y_1,y_2 \in [0, 1],
\end{equation}
where $l=0.25$ is the correlation length. We keep the first $120$ leading terms of the expansion, which leads to a  $120$-dimensional parametric problem. 

Similar as previous examples, a DeepONet is pretrained with clean data to represent the physical relation between $\lambda$ and $u$: it approximates the solution operator which maps $\log(\lambda)$ to $u$.  We assume that we have noisy and gappy data of $u$ and $log(\lambda)$ and we reconstruct these two functions synergistically based on the pretrained NO. In our numerical experiments, 20 measurements of $u$ and 20 measurements of $\log(\lambda)$ are randomly sampled and corrupted by additive Gaussian noises with scales $0.5$ and $0.05$, respectively. Results are presented in Fig.~\ref{fig:example_4}. As shown, we observe that the predicted means are accurate compared to the references and the errors between the predicted means and the reference are mostly bounded by the predicted uncertainties.

\section{Summary}\label{sec:5}

In this paper, we proposed a Bayesian approach to quantify uncertainties arsing from the noisy inputs-outputs in two prevalent scientific machine learning (SciML) models, the physics-informed neural networks (PINNs), which encode the physics via automatic differentation \cite{raissi2019physics}, and neural operators (NOs) \cite{kovachki2021neural}, which learn the hidden physics from data and serve as equation-free surrogates \cite{li2020fourier, meng2022learning, psaros2023uncertainty, zou2022neuraluq}. In particular, we targeted solving hybrid problems using PINNs and NOs. In contrast to most existing work on uncertainty quantification (UQ) in SciML in which only noisy outputs are considered \cite{psaros2023uncertainty}, our approach encompasses the scenarios, where the inputs of SciML models (e.g., the spatial-temporal coordinate in PINNs and the input functions in NOs) are corrupted by noises as well.
The consideration of the noise in the inputs provides two immediate benefits: (1) the deployment of SciML models in tackling problems where uncertainties are associated with spatial-temporal coordinates, e.g. particle image velocimetry (PIV) \cite{cai2019dense, cai2019particle} and hydrological modeling \cite{kavetski2006bayesian1, kavetski2006bayesian2}, is facilitated; (2) accounting for input noise allows for the compatibility of pretrained Neural Operators (NOs) with noisy input function measurements. This versatility makes the NOs  more robust and adaptable in practical applications.

The proposed approach can seamlessly be combined with the PINNs, vanilla Fourier neural operators (FNOs) \cite{li2020fourier}, vanilla DeepONets \cite{lu2021learning}, and multi-input DeepONets \cite{jin2022mionet} in addressing the \textit{hybrid} problem with noisy inputs and outputs.  We considered four numerical examples to demonstrate the effectiveness of the present method. Specifically, we considered: (1) a 1D Poisson equation with PINNs, where both the spatial coordinate and the corresponding responses were measured with noises; (2) a 1D Burgers equation, where the solution was reconstructed from the noisy and sparse measurements on the input/output functions with pretrained FNOs; (3) a 1D time-dependent reaction-diffusion equation with heterogeneous diffusion term in which the solution, the source term and the diffusion term were reconstructed from noisy measurements with pretrained multi-input DeepONet; and (4) a 120-dimensional Darcy problem, where the hydraulic head and they hydraulic conductivity were regressed synergistically from sparse and noisy data. The results show that the proposed method is able to provide accurate inferences and quantify uncertainties based on noisy inputs-outputs data. We expect our approach to be effectively and reliably applied to various practical tasks arising from computational science and engineering.

\section*{Acknowledgement}

GEK acknowledge the support from the AIM for Composites, an Energy Frontier Research Center funded by the U.S. Department of Energy (DOE), Office of Science, Basic Energy Sciences (BES) under Award DE-SC0023389 and the MURI/AFOSR project (FA9550-20-1-0358). XM would like to acknowledge the support of the National Natural Science Foundation of China (No. 12201229) and the Xiaomi Young Talents Program.

\bibliographystyle{unsrt}
\bibliography{references}

\begin{thebibliography}{10}

\bibitem{karniadakis2021physics}
George~Em Karniadakis, Ioannis~G Kevrekidis, Lu~Lu, Paris Perdikaris, Sifan Wang, and Liu Yang.
\newblock Physics-informed machine learning.
\newblock {\em Nature Reviews Physics}, 3(6):422--440, 2021.

\bibitem{psaros2023uncertainty}
Apostolos~F Psaros, Xuhui Meng, Zongren Zou, Ling Guo, and George~Em Karniadakis.
\newblock Uncertainty quantification in scientific machine learning: Methods, metrics, and comparisons.
\newblock {\em Journal of Computational Physics}, page 111902, 2023.

\bibitem{zou2022neuraluq}
Zongren Zou, Xuhui Meng, Apostolos~F Psaros, and George~Em Karniadakis.
\newblock {NeuralUQ}: A comprehensive library for uncertainty quantification in neural differential equations and operators.
\newblock {\em arXiv preprint arXiv:2208.11866}, 2022.

\bibitem{yang2022scalable}
Yibo Yang, Georgios Kissas, and Paris Perdikaris.
\newblock Scalable uncertainty quantification for deep operator networks using randomized priors.
\newblock {\em Computer Methods in Applied Mechanics and Engineering}, 399:115399, 2022.

\bibitem{lin2021accelerated}
Guang Lin, Christian Moya, and Zecheng Zhang.
\newblock Accelerated replica exchange stochastic gradient {Langevin} diffusion enhanced {Bayesian DeepONet} for solving noisy parametric {PDEs}.
\newblock {\em arXiv preprint arXiv:2111.02484}, 2021.

\bibitem{moya2023deeponet}
Christian Moya, Shiqi Zhang, Guang Lin, and Meng Yue.
\newblock {DeepONet-Grid-UQ}: A trustworthy deep operator framework for predicting the power grid’s post-fault trajectories.
\newblock {\em Neurocomputing}, 535:166--182, 2023.

\bibitem{yang2021b}
Liu Yang, Xuhui Meng, and George~Em Karniadakis.
\newblock {B-PINNs}: Bayesian physics-informed neural networks for forward and inverse {PDE} problems with noisy data.
\newblock {\em Journal of Computational Physics}, 425:109913, 2021.

\bibitem{linka2022bayesian}
Kevin Linka, Amelie Sch{\"a}fer, Xuhui Meng, Zongren Zou, George~Em Karniadakis, and Ellen Kuhl.
\newblock Bayesian physics informed neural networks for real-world nonlinear dynamical systems.
\newblock {\em Computer Methods in Applied Mechanics and Engineering}, 402:115346, 2022.

\bibitem{zou2023correcting}
Zongren Zou, Xuhui Meng, and George~Em Karniadakis.
\newblock Correcting model misspecification in physics-informed neural networks ({PINNs}).
\newblock {\em arXiv preprint arXiv:2310.10776}, 2023.

\bibitem{yang2022multi}
Mingyuan Yang and John~T Foster.
\newblock Multi-output physics-informed neural networks for forward and inverse {PDE} problems with uncertainties.
\newblock {\em Computer Methods in Applied Mechanics and Engineering}, 402:115041, 2022.

\bibitem{zhang2023discovering}
Zhen Zhang, Zongren Zou, Ellen Kuhl, and George~Em Karniadakis.
\newblock Discovering a reaction-diffusion model for {Alzheimer's} disease by combining {PINNs} with symbolic regression.
\newblock {\em arXiv preprint arXiv:2307.08107}, 2023.

\bibitem{zhang2019quantifying}
Dongkun Zhang, Lu~Lu, Ling Guo, and George~Em Karniadakis.
\newblock Quantifying total uncertainty in physics-informed neural networks for solving forward and inverse stochastic problems.
\newblock {\em Journal of Computational Physics}, 397:108850, 2019.

\bibitem{yang2020physics}
Liu Yang, Dongkun Zhang, and George~Em Karniadakis.
\newblock Physics-informed generative adversarial networks for stochastic differential equations.
\newblock {\em SIAM Journal on Scientific Computing}, 42(1):A292--A317, 2020.

\bibitem{meng2022learning}
Xuhui Meng, Liu Yang, Zhiping Mao, Jos{\'e} del {\'A}guila~Ferrandis, and George~Em Karniadakis.
\newblock Learning functional priors and posteriors from data and physics.
\newblock {\em Journal of Computational Physics}, 457:111073, 2022.

\bibitem{zou2023hydra}
Zongren Zou and George~Em Karniadakis.
\newblock {L-HYDRA}: Multi-head physics-informed neural networks.
\newblock {\em arXiv preprint arXiv:2301.02152}, 2023.

\bibitem{YIN2023105424}
Minglang Yin, Zongren Zou, Enrui Zhang, Cristina Cavinato, Jay~D. Humphrey, and George~Em Karniadakis.
\newblock A generative modeling framework for inferring families of biomechanical constitutive laws in data-sparse regimes.
\newblock {\em Journal of the Mechanics and Physics of Solids}, 181:105424, 2023.

\bibitem{winovich2019convpde}
Nick Winovich, Karthik Ramani, and Guang Lin.
\newblock {ConvPDE-UQ}: Convolutional neural networks with quantified uncertainty for heterogeneous elliptic partial differential equations on varied domains.
\newblock {\em Journal of Computational Physics}, 394:263--279, 2019.

\bibitem{raissi2019physics}
Maziar Raissi, Paris Perdikaris, and George~E Karniadakis.
\newblock Physics-informed neural networks: A deep learning framework for solving forward and inverse problems involving nonlinear partial differential equations.
\newblock {\em Journal of Computational physics}, 378:686--707, 2019.

\bibitem{lu2021learning}
Lu~Lu, Pengzhan Jin, Guofei Pang, Zhongqiang Zhang, and George~Em Karniadakis.
\newblock Learning nonlinear operators via {DeepONet} based on the universal approximation theorem of operators.
\newblock {\em Nature machine intelligence}, 3(3):218--229, 2021.

\bibitem{li2020fourier}
Zongyi Li, Nikola Kovachki, Kamyar Azizzadenesheli, Burigede Liu, Kaushik Bhattacharya, Andrew Stuart, and Anima Anandkumar.
\newblock Fourier neural operator for parametric partial differential equations.
\newblock {\em arXiv preprint arXiv:2010.08895}, 2020.

\bibitem{cai2019particle}
Shengze Cai, Jiaming Liang, Qi~Gao, Chao Xu, and Runjie Wei.
\newblock Particle image velocimetry based on a deep learning motion estimator.
\newblock {\em IEEE Transactions on Instrumentation and Measurement}, 69(6):3538--3554, 2019.

\bibitem{cai2019dense}
Shengze Cai, Shichao Zhou, Chao Xu, and Qi~Gao.
\newblock Dense motion estimation of particle images via a convolutional neural network.
\newblock {\em Experiments in Fluids}, 60:1--16, 2019.

\bibitem{kavetski2006bayesian1}
Dmitri Kavetski, George Kuczera, and Stewart~W Franks.
\newblock Bayesian analysis of input uncertainty in hydrological modeling: 1. theory.
\newblock {\em Water resources research}, 42(3), 2006.

\bibitem{kavetski2006bayesian2}
Dmitri Kavetski, George Kuczera, and Stewart~W Franks.
\newblock Bayesian analysis of input uncertainty in hydrological modeling: 2. application.
\newblock {\em Water resources research}, 42(3), 2006.

\bibitem{dellaportas1995bayesian}
Petros Dellaportas and David~A Stephens.
\newblock Bayesian analysis of errors-in-variables regression models.
\newblock {\em Biometrics}, pages 1085--1095, 1995.

\bibitem{fan1993nonparametric}
Jianqing Fan and Young~K Truong.
\newblock Nonparametric regression with errors in variables.
\newblock {\em The Annals of Statistics}, pages 1900--1925, 1993.

\bibitem{gleser1981estimation}
Leon~Jay Gleser.
\newblock Estimation in a multivariate ``errors in variables'' regression model: large sample results.
\newblock {\em The Annals of Statistics}, pages 24--44, 1981.

\bibitem{wright1999bayesian}
WA~Wright.
\newblock Bayesian approach to neural-network modeling with input uncertainty.
\newblock {\em IEEE Transactions on Neural Networks}, 10(6):1261--1270, 1999.

\bibitem{tresp1993training}
Volker Tresp, Subutai Ahmad, and Ralph Neuneier.
\newblock Training neural networks with deficient data.
\newblock {\em Advances in Neural Information Processing Systems}, 6, 1993.

\bibitem{van2000learning}
J{\"u}rgen Van~Gorp, Johan Schoukens, and Rik Pintelon.
\newblock Learning neural networks with noisy inputs using the errors-in-variables approach.
\newblock {\em IEEE Transactions on Neural Networks}, 11(2):402--414, 2000.

\bibitem{williams2006gaussian}
Christopher~KI Williams and Carl~Edward Rasmussen.
\newblock {\em Gaussian processes for machine learning}, volume~2.
\newblock MIT press Cambridge, MA, 2006.

\bibitem{atkinson1998statistical}
Greg Atkinson and Alan~M Nevill.
\newblock Statistical methods for assessing measurement error (reliability) in variables relevant to sports medicine.
\newblock {\em Sports medicine}, 26(4):217--238, 1998.

\bibitem{schennach2016recent}
Susanne~M Schennach.
\newblock Recent advances in the measurement error literature.
\newblock {\em Annual Review of Economics}, 8:341--377, 2016.

\bibitem{girard2002gaussian}
Agathe Girard, Carl Rasmussen, Joaquin~Q Candela, and Roderick Murray-Smith.
\newblock Gaussian process priors with uncertain inputs application to multiple-step ahead time series forecasting.
\newblock {\em Advances in Neural Information Processing systems}, 15, 2002.

\bibitem{mchutchon2011gaussian}
Andrew McHutchon and Carl Rasmussen.
\newblock Gaussian process training with input noise.
\newblock {\em Advances in Neural Information Processing Systems}, 24, 2011.

\bibitem{de2023convergence}
Maarten~V de~Hoop, Nikola~B Kovachki, Nicholas~H Nelsen, and Andrew~M Stuart.
\newblock Convergence rates for learning linear operators from noisy data.
\newblock {\em SIAM/ASA Journal on Uncertainty Quantification}, 11(2):480--513, 2023.

\bibitem{patel2022error}
Ravi Patel, Indu Manickam, Myoungkyu Lee, and Mamikon Gulian.
\newblock {Error-in-variables modelling for operator learning}.
\newblock In {\em Mathematical and Scientific Machine Learning}, pages 142--157. PMLR, 2022.

\bibitem{garg2022variational}
Shailesh Garg and Souvik Chakraborty.
\newblock {Variational Bayes Deep Operator Network: A data-driven Bayesian solver for parametric differential equations}.
\newblock {\em arXiv preprint arXiv:2206.05655}, 2022.

\bibitem{li2021physics}
Zongyi Li, Hongkai Zheng, Nikola Kovachki, David Jin, Haoxuan Chen, Burigede Liu, Kamyar Azizzadenesheli, and Anima Anandkumar.
\newblock Physics-informed neural operator for learning partial differential equations.
\newblock {\em arXiv preprint arXiv:2111.03794}, 2021.

\bibitem{kovachki2021neural}
Nikola~B Kovachki, Zongyi Li, Burigede Liu, Kamyar Azizzadenesheli, Kaushik Bhattacharya, Andrew~M Stuart, and Anima Anandkumar.
\newblock Neural operator: Learning maps between function spaces with applications to {PDEs}.
\newblock {\em J. Mach. Learn. Res.}, 24(89):1--97, 2023.

\bibitem{tang2023adversarial}
Kejun Tang, Jiayu Zhai, Xiaoliang Wan, and Chao Yang.
\newblock Adversarial adaptive sampling: Unify {PINN} and optimal transport for the approximation of pdes.
\newblock {\em arXiv preprint arXiv:2305.18702}, 2023.

\bibitem{tang2023pinns}
Kejun Tang, Xiaoliang Wan, and Chao Yang.
\newblock {DAS-PINNs}: A deep adaptive sampling method for solving high-dimensional partial differential equations.
\newblock {\em Journal of Computational Physics}, 476:111868, 2023.

\bibitem{sirignano2018dgm}
Justin Sirignano and Konstantinos Spiliopoulos.
\newblock {DGM: A deep learning algorithm for solving partial differential equations}.
\newblock {\em Journal of computational physics}, 375:1339--1364, 2018.

\bibitem{chen2021physics}
Zhao Chen, Yang Liu, and Hao Sun.
\newblock Physics-informed learning of governing equations from scarce data.
\newblock {\em Nature communications}, 12(1):6136, 2021.

\bibitem{chen2023leveraging}
Paula Chen, Tingwei Meng, Zongren Zou, J{\'e}r{\^o}me Darbon, and George~Em Karniadakis.
\newblock Leveraging multi-time {Hamilton-Jacobi} {PDEs} for certain scientific machine learning problems.
\newblock {\em arXiv preprint arXiv:2303.12928}, 2023.

\bibitem{chen2023leveraging2}
Paula Chen, Tingwei Meng, Zongren Zou, Jérôme Darbon, and George~Em Karniadakis.
\newblock {Leveraging Hamilton-Jacobi PDEs with time-dependent Hamiltonians for continual scientific machine learning}, 2023.

\bibitem{lu2021deepxde}
Lu~Lu, Xuhui Meng, Zhiping Mao, and George~Em Karniadakis.
\newblock {DeepXDE}: A deep learning library for solving differential equations.
\newblock {\em SIAM review}, 63(1):208--228, 2021.

\bibitem{wang2021learning}
Sifan Wang, Hanwen Wang, and Paris Perdikaris.
\newblock Learning the solution operator of parametric partial differential equations with physics-informed {DeepONets}.
\newblock {\em Science advances}, 7(40):eabi8605, 2021.

\bibitem{jin2022mionet}
Pengzhan Jin, Shuai Meng, and Lu~Lu.
\newblock {MIONet}: Learning multiple-input operators via tensor product.
\newblock {\em SIAM Journal on Scientific Computing}, 44(6):A3490--A3514, 2022.

\bibitem{lu2022comprehensive}
Lu~Lu, Xuhui Meng, Shengze Cai, Zhiping Mao, Somdatta Goswami, Zhongqiang Zhang, and George~Em Karniadakis.
\newblock A comprehensive and fair comparison of two neural operators (with practical extensions) based on fair data.
\newblock {\em Computer Methods in Applied Mechanics and Engineering}, 393:114778, 2022.

\bibitem{neal2011mcmc}
Radford~M Neal et~al.
\newblock {MCMC using Hamiltonian dynamics}.
\newblock {\em Handbook of Markov Chain Monte Carlo}, 2(11):2, 2011.

\bibitem{andrieu2008tutorial}
Christophe Andrieu and Johannes Thoms.
\newblock A tutorial on adaptive {MCMC}.
\newblock {\em Statistics and computing}, 18:343--373, 2008.

\bibitem{kingma2014adam}
Diederik~P Kingma and Jimmy Ba.
\newblock Adam: A method for stochastic optimization.
\newblock {\em arXiv preprint arXiv:1412.6980}, 2014.

\end{thebibliography}

\appendix
\section{A pedagogical example: function approximation}\label{sec:appendix_a}

\begin{figure}[ht]
    \centering
    \subfigure[]{
    \includegraphics[scale=.4]{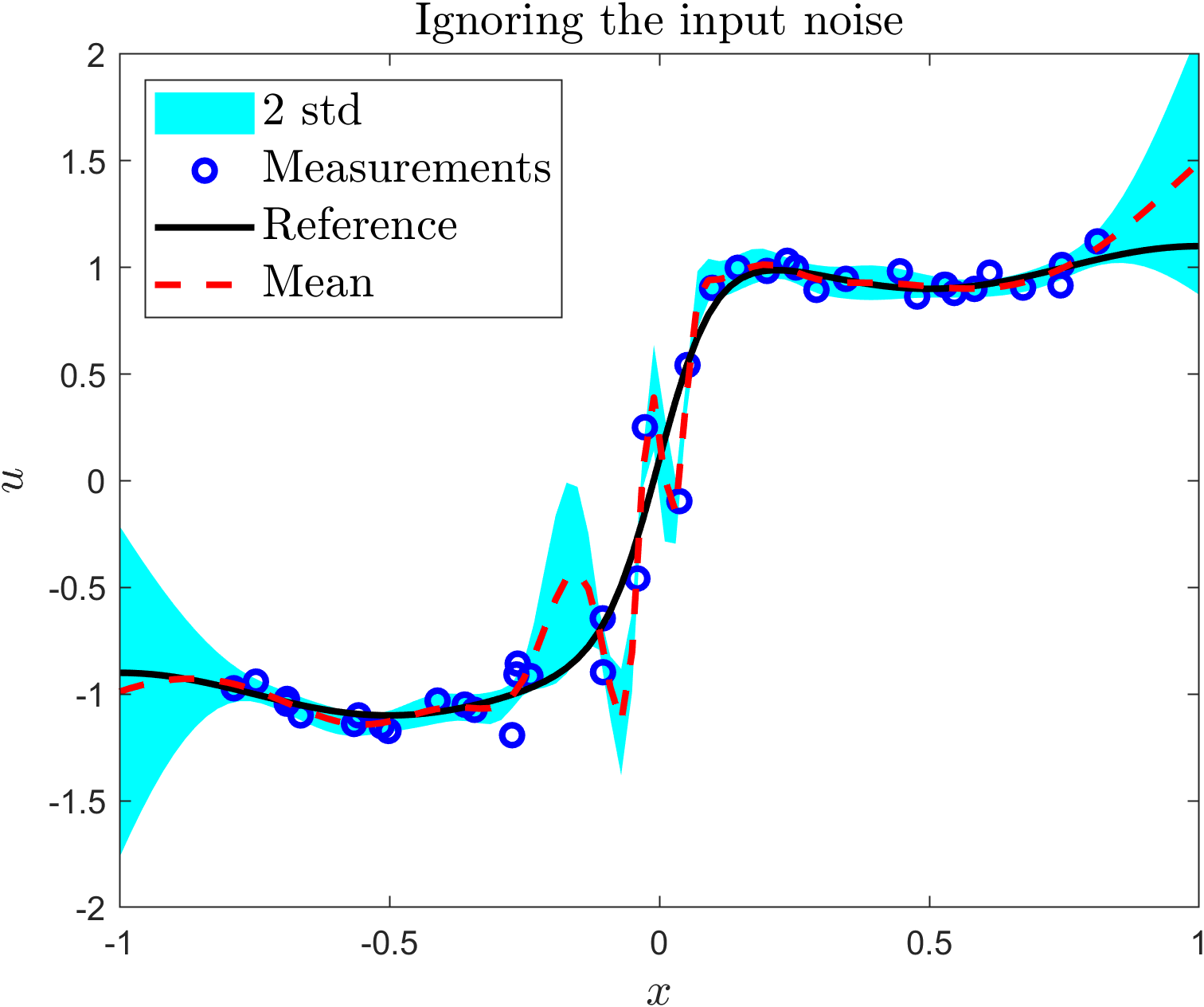}
    \includegraphics[scale=.4]{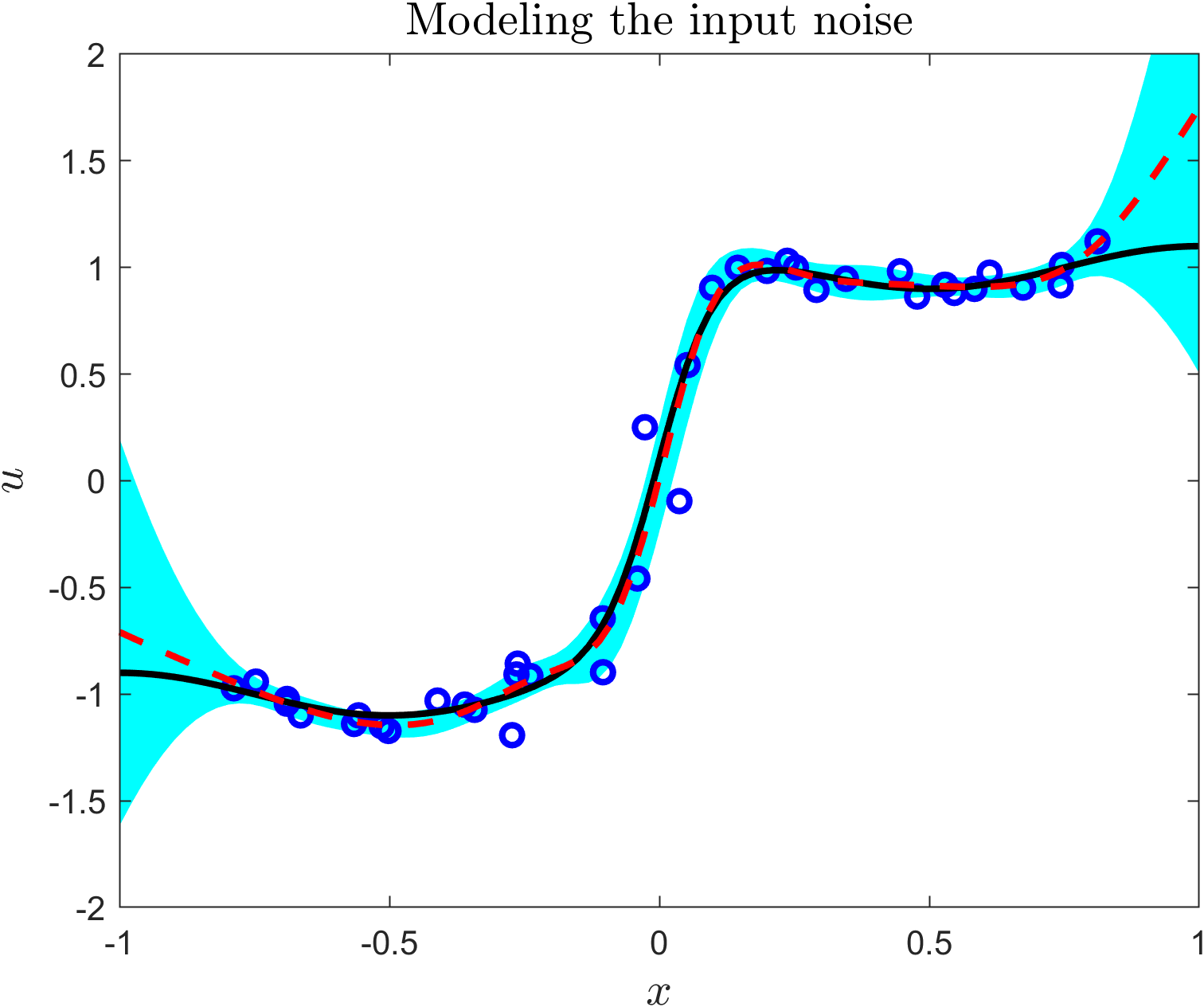}
    \includegraphics[scale=.4]{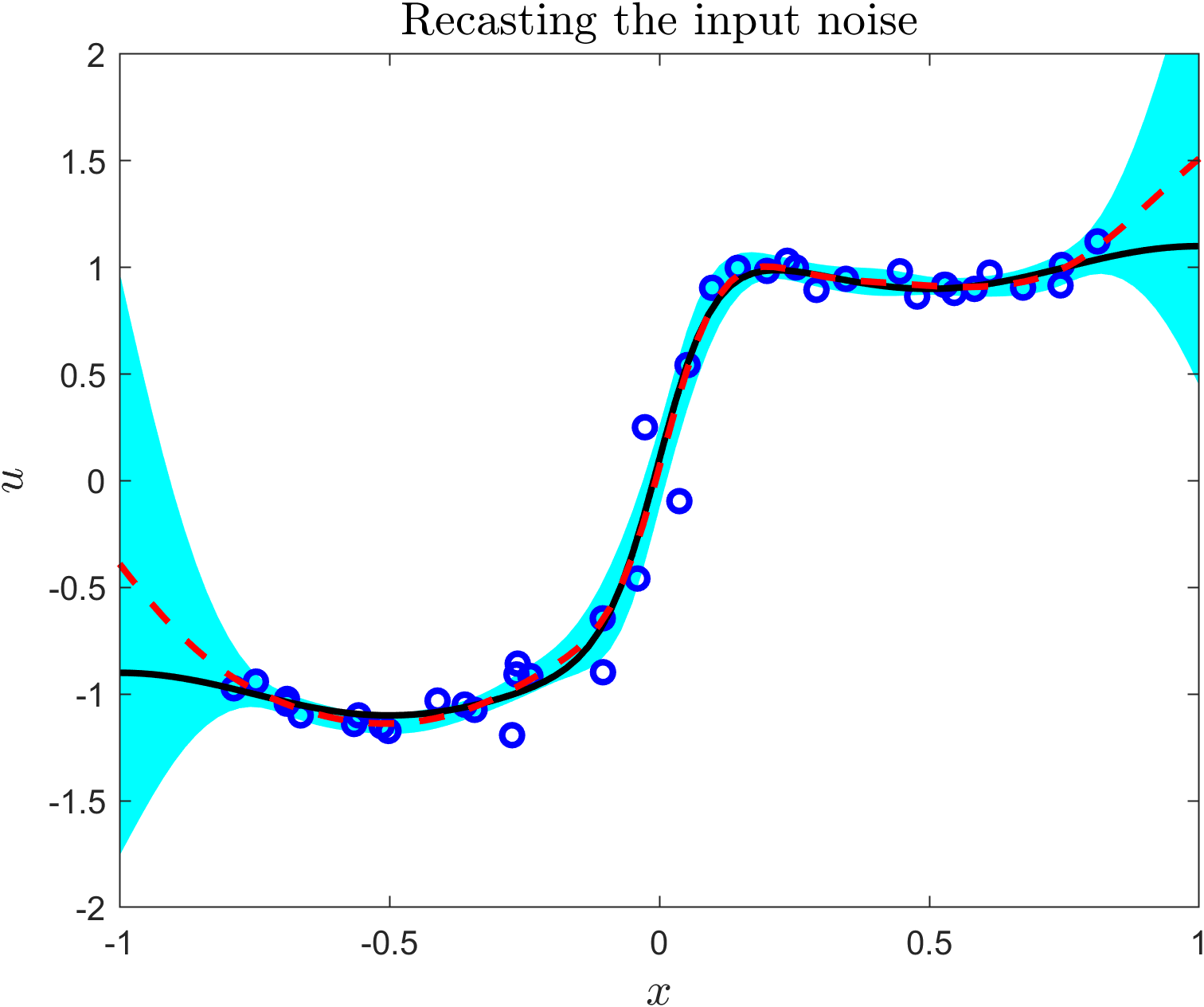}
    }
    \subfigure[]{
    \includegraphics[scale=.4]{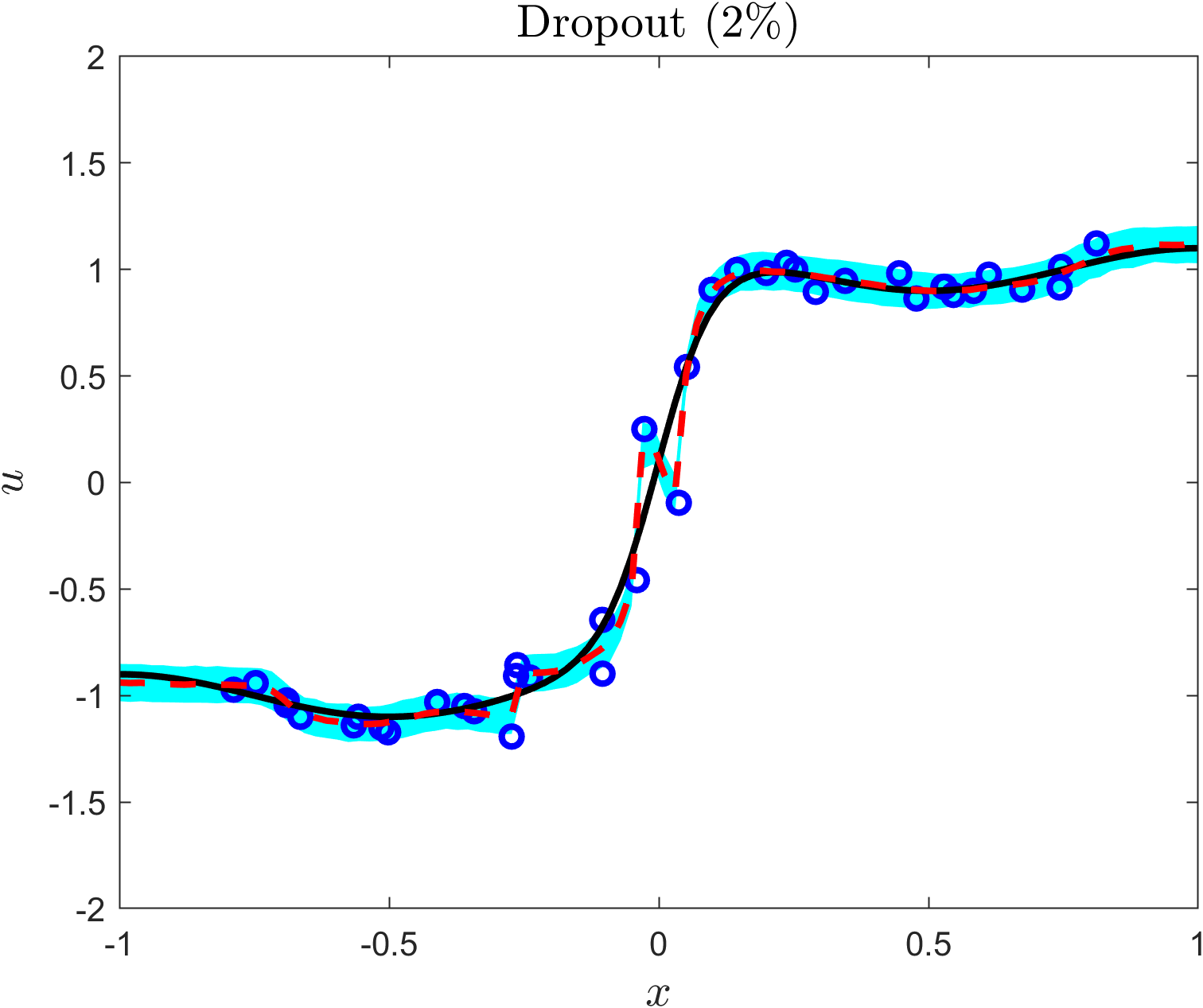}
    \includegraphics[scale=.4]{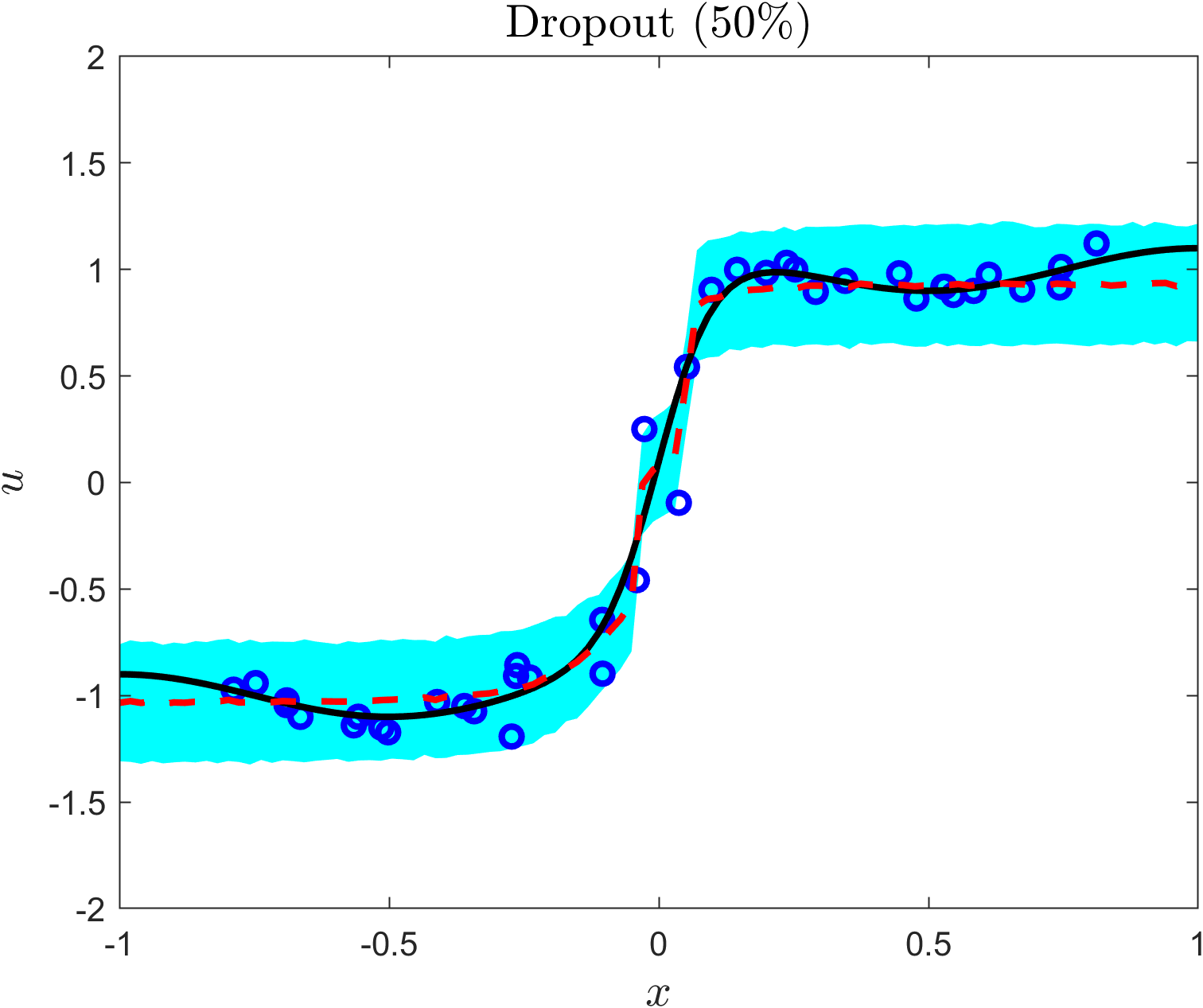}
    }
    \subfigure[]{
    \includegraphics[scale=.4]{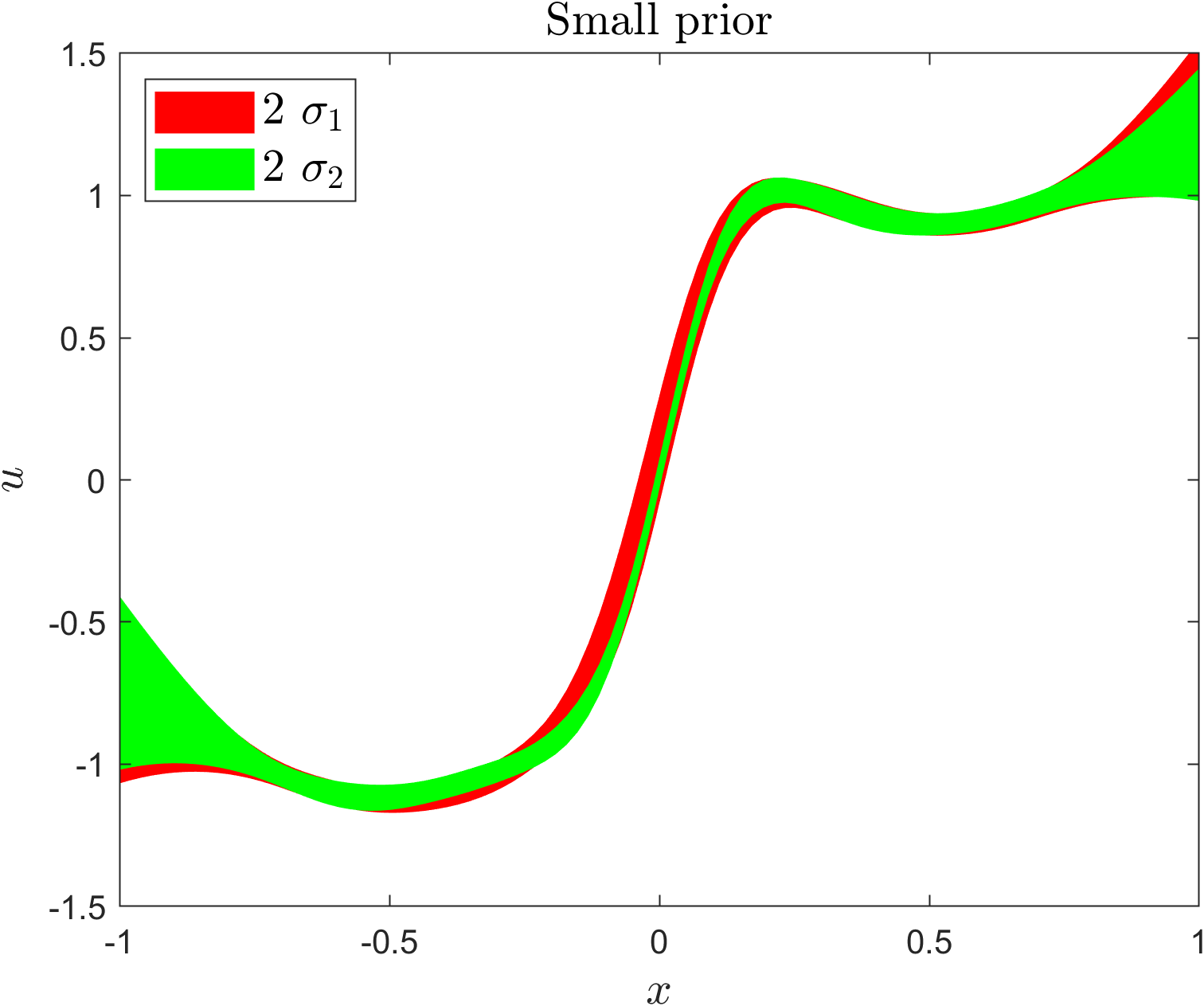}
    }
    \caption{Function approximation: Regressing \eqref{eq:function} from noisy input-output data. (a) From left to right are results from BNNs ignoring the input noise, modeling the input noise, and recasting the input noise as a heterscedastic output noise via first-order Taylor expansion. (b) shows results from the Dropout method with different dropout rates. (c) presents predicted uncertainties of BNNs ignoring and modeling the input noise, represented by $2 \sigma_2$ and $2 \sigma_1$, respectively, with smaller standard deviation ($0.5$, compared to $1$ in (a)) in the prior of the NN parameter. 
    } 
    \label{fig:example_0}
\end{figure}

In this section, we consider a pedagogical example in which we regress a function from its noisy inputs-outputs data using neural networks (NNs). We assume the target function is:
\begin{equation}\label{eq:function}
    u(x) = 0.1\cos(2\pi x) + \tanh(3\pi x), x\in[-1, 1].
\end{equation}
The training data is expressed as $\{(x_i, u_i)\}_{i=1}^N$ where $N$ denotes the number of data and $x_i, u_i, i=1,...,N$ are observed with additive Gaussian noises.
Intuitively, noisy input would result in larger error in the region where the target function is ``steep'' and smaller error in the region where the target function is ``flat''. The quality of the predicted uncertainty is evaluated in a similar manner. We note that, in this example, we focus on the regressed function and hence, in testing stage, clean input (clean $x$) is used for evaluation. In this example, the input noise refers to the noise from observing $x$ while the output noise refers to the noise from observing $u$. We assume scales of the input and output noises are $0.03$ and $0.05$, respectively, and choose an NN with two hidden layers, each of which has $50$ neurons and hyperbolic tangent activation function. The prior of the NN parameter is assumed as independent Gaussian with mean zero and standard deviation one.

In Fig.~\ref{fig:example_0}(a), we present results from BNNs ignoring the input, modeling the input noise, and recasting the input noise as a heteroscedastic output noise via first-order Taylor expansion. Both proposed approaches provide accurate inferences and errors are bounded by predicted uncertainties. We observe a severe overfitting phenomenon from the result of the original BNN method, demonstrating one of the major consequences of ignoring the input noise. This overfitting issue appears more significantly in the region where the target function is steep. In Fig.~\ref{fig:example_0}(b), we display results from the Dropout method for quantifying uncertainty \cite{psaros2023uncertainty, zou2022neuraluq}, in which the predicted uncertainty depends on the choice of the dropout rate. When the dropout rate is small ($2\%$), the predicted uncertainty is satisfactory in the region where the target function is flat while it suffers from the overfitting issue as well. When the dropout rate is increased ($50\%$) for the purpose of mitigating overfitting, the predicted uncertainty becomes unnecessarily large, resulting in under-confident predictions.

One may argue that the overfitting issue caused by ignoring the input noise in Fig.~\ref{fig:example_0}(a) could be resolved by proper regularization, such as smaller standard deviation for the prior of NN parameter. We further compare modeling and ignoring the input noise under the setting where a ``smaller'' prior is used (standard deviation $0.5$). As shown in Fig.~\ref{fig:example_0}(c), the overfitting issue is indeed resolved, at the cost of too small predicted uncertainty, presented by $2\sigma_2$, (and hence over-confident prediction) in the region where the input noise is supposed to have a significant impact. As a comparison, our approach provide more reasonable predicted uncertainty, represented by $2\sigma_1$ in Fig.~\ref{fig:example_0}(c), e.g., the ``horizontal'' band is much larger, demonstrating that the effect of the input noise is incorporated.

\section{Details of hyperparameters in the numerical experiments}\label{sec:appendix_b}

In this section, we provide details of all numerical examples presented in Sec.~\ref{sec:4}. We use Hamiltonian Monte Carlo (HMC) \cite{neal2011mcmc} with adaptive step size \cite{andrieu2008tutorial} to estimate all posterior distributions, in which the number of steps for the leap-frog scheme is set to $50$ and $1,000$ posterior samples are obtained. The number of burn-in samples and the initial step size are tuned such that the acceptance rate between $50\%$ and $70\%$ are achieved. A open-source Python library for UQ in SciML, termed NeuralUQ \cite{zou2022neuraluq}, is used for implementations. For examples involving NOs, the functions for hybrid problems are randomly sampled from the testing data rather than data used to train NOs beforehand.

In Sec.~\ref{sec:example_1}, we employ BNNs with two hidden layers, each of which is equipped hyperbolic tangent as the activation function and has $50$ neurons. The prior of NN parameter, i.e. $p(\theta),$ is set to independent Gaussian with mean zero and standard deviation one. The prior of the actual inputs, i.e. $p(\chi)$, is set to independent Gaussian with mean zero and standard deviation $100$, representing weak regularization of the actual inputs.
We observed that in practice, omitting $p(\chi)$ in the posterior, which indicates $\chi$ follows independent uniform distribution on a sufficiently large interval, was able to provide accurate inferences and reasonable uncertainties as well. 

In Section~\ref{sec:example_2}, a vanilla FNO is employed to learn the solution operator of \eqref{eq:burgers}, which maps the initial condition to the solution at $t=1$. All hyperparameters of this FNO and its training are the same as the ones in \cite{li2020fourier}. We use the same data from \cite{li2020fourier} ($1,000$ for training and $200$ for testing where the initial condition is generated from the distribution $\mathcal{N}(0, 625(-\Delta + 25 I)^{-2})$) and discretize the input and output functions on a uniform mesh with $128$ grids. The pretrained FNO used for the downstream hybrid problem achieved $0.50\%$ relative $L_2$ error in average on testing data. The prior of the discretized input function is set to multi-variate Gaussian with mean zero and covariance matrix estimated from the $1,000$ data used to train the FNO. 

In Section~\ref{sec:example_3}, a multi-input DeepONet \cite{jin2022mionet} is employed to learn the solution operator of \eqref{eq:reaction_diffusion}, which maps the function $k$ in the diffusion term and the source term $f$ to the solution. Both input functions are discretized on a uniform mesh with $100$ grids.
There are two branch nets in the multi-input DeepONet and they both have one hidden layer with $200$ neurons and ReLU activation function. The trunk net is of the same architecture. The output dimensions of the trunk net and the branch nets are $200$. The Adam optimizer \cite{kingma2014adam} with learning rate $0.001$ and default setting is employed to train the multi-input DeepONet for $200,000$ iterations. The data of $k$ and $f$ are generated by a Gaussian process with zero mean and the following exponential squared kernel:
\begin{equation}
    k(x_1, x_2) = \exp(-\frac{(x_1-x_2)^2}{2l^2}), ~x_1,x_2\in [0, 1],
\end{equation}
where $l=0.2$ is the correlation length. We used $5,000$ data to train the DeepONet and achieved $2.77\%$ relative $L_2$ error in average on $1,000$ testing data. 

In Section~\ref{sec:example_4}, a vanilla DeepONet is employed to learn the solution operator of \eqref{eq:darcy}, which maps the logarithm of the hydraulic conductivity $\log(\lambda)$ to the hydraulic head $u$. 
Both $\log(\lambda)$ and $u$ are discretized on a $31\times 31$ uniform grid.
The branch net of the DeepONet has three hidden layers while the trunk net has two hidden layers. Each layer has $128$ neurons and has hyperbolic tangent as the activation function. The output dimensions of the branch and trunk nets are $128$. We employed the same Adam optimizer to train the DeepONet for $50,000$ iterations on $10,000$ training data and achieved $2.08\%$ relative $L_2$ error on $1,000$ testing data.

\section{Additional results}\label{sec:appendix_additional}

\subsection{1D nonlinear Poisson equation with PINNs}

\begin{figure}[ht!]
    \centering
    \subfigure[Inference/fitting of $f$.]{
    \includegraphics[scale=.4]{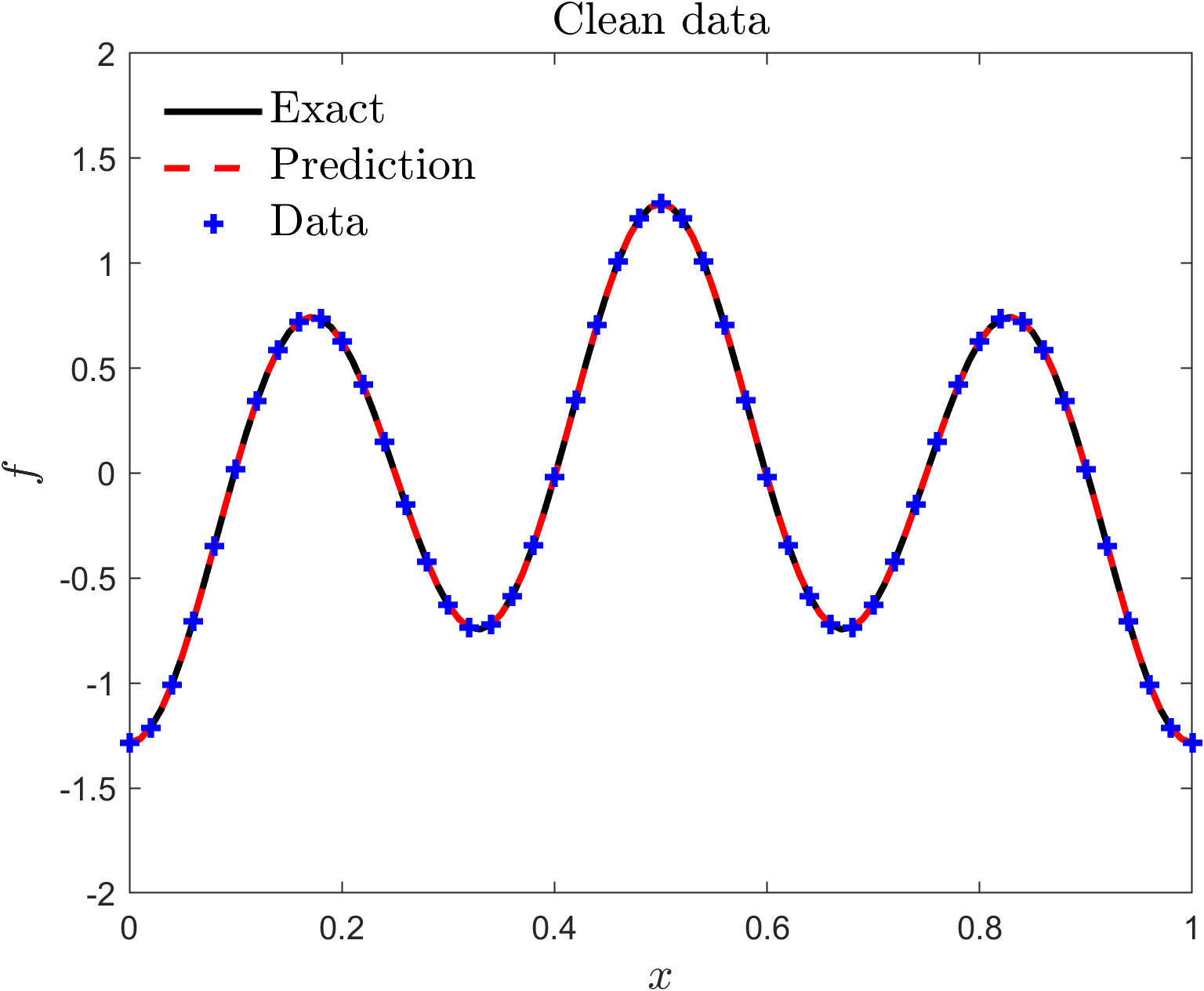}
    \includegraphics[scale=.4]{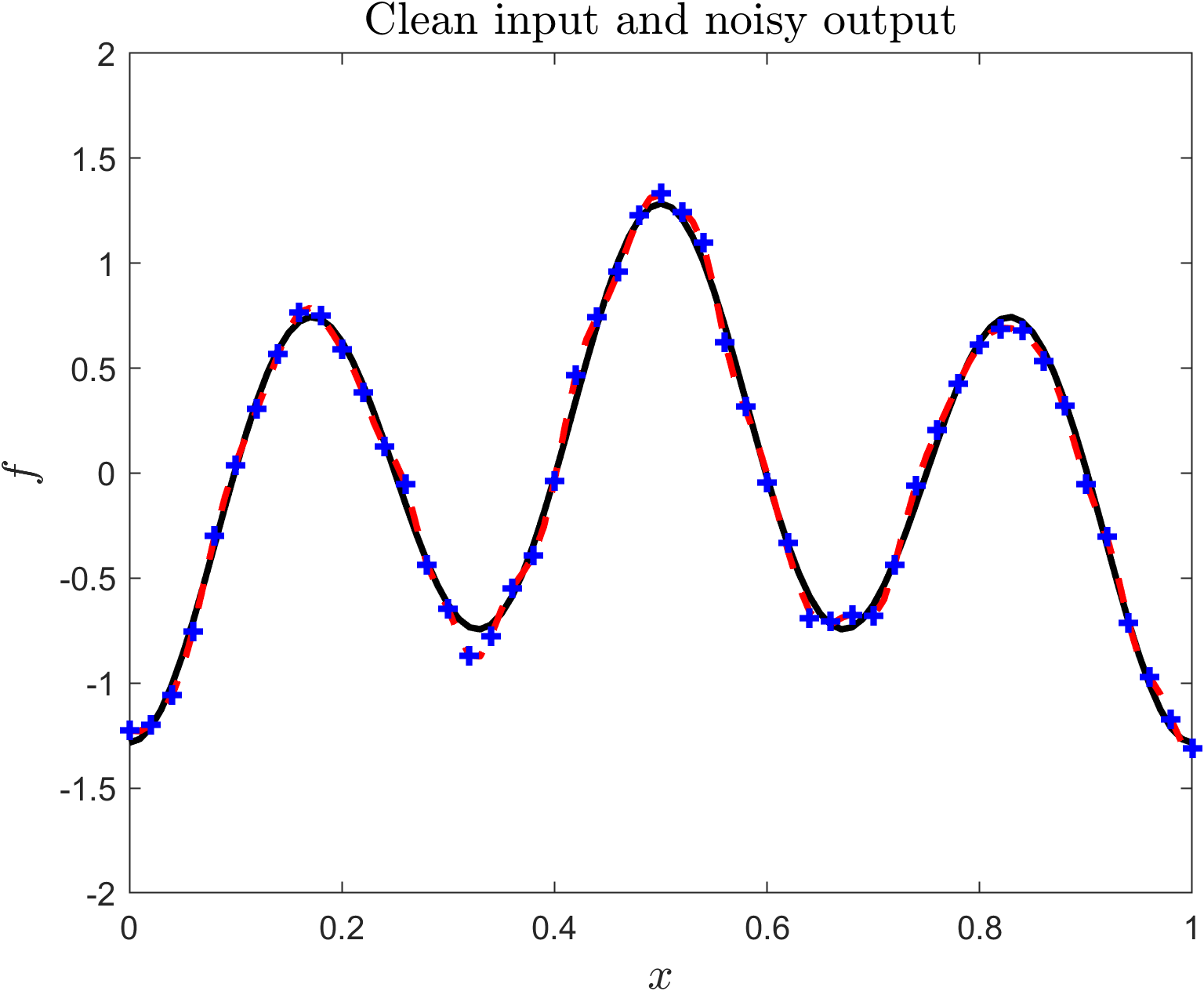}
    \includegraphics[scale=.4]{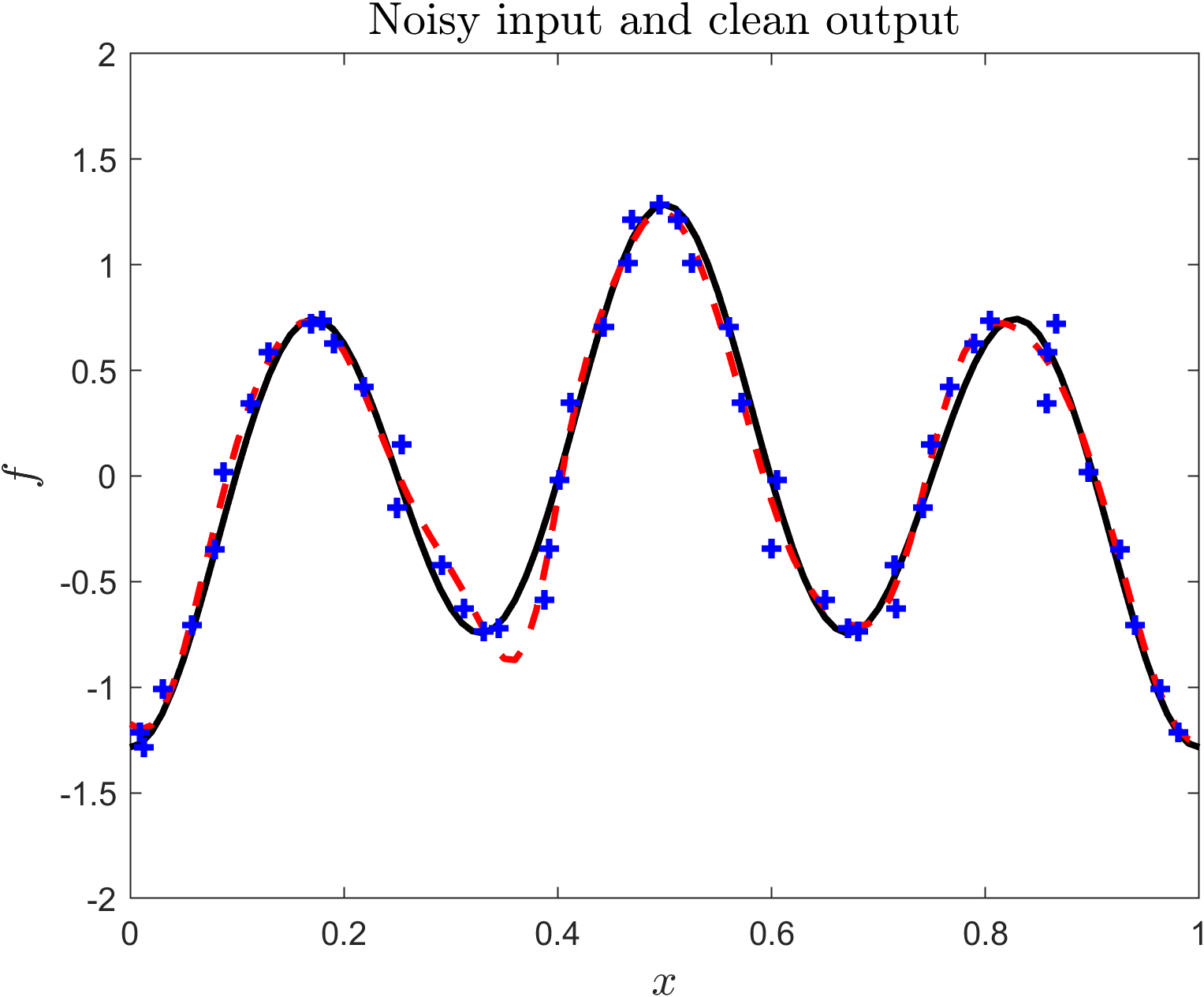}
    }
    \subfigure[Inference of $u$.]{
    \includegraphics[scale=.4]{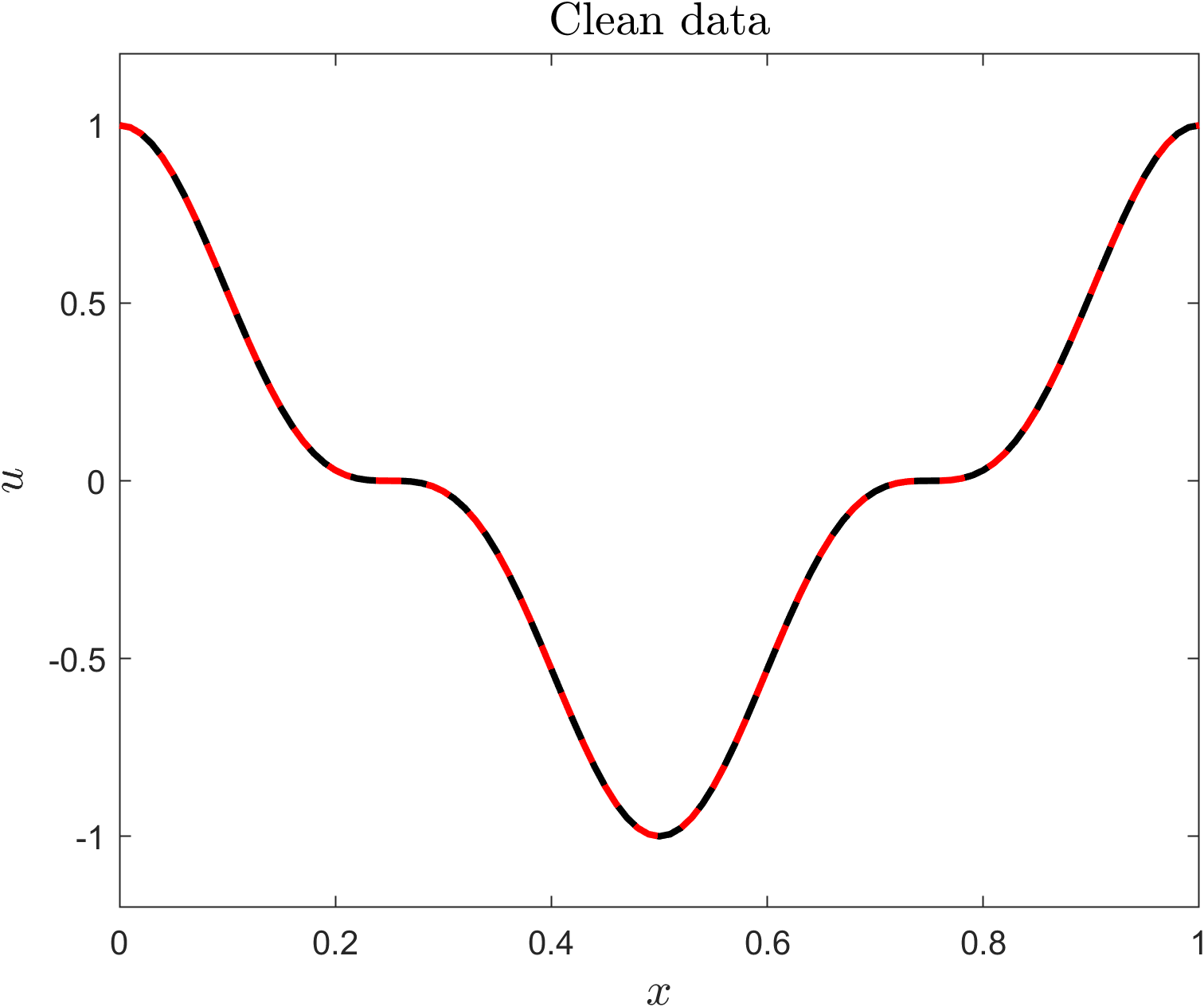}
    \includegraphics[scale=.4]{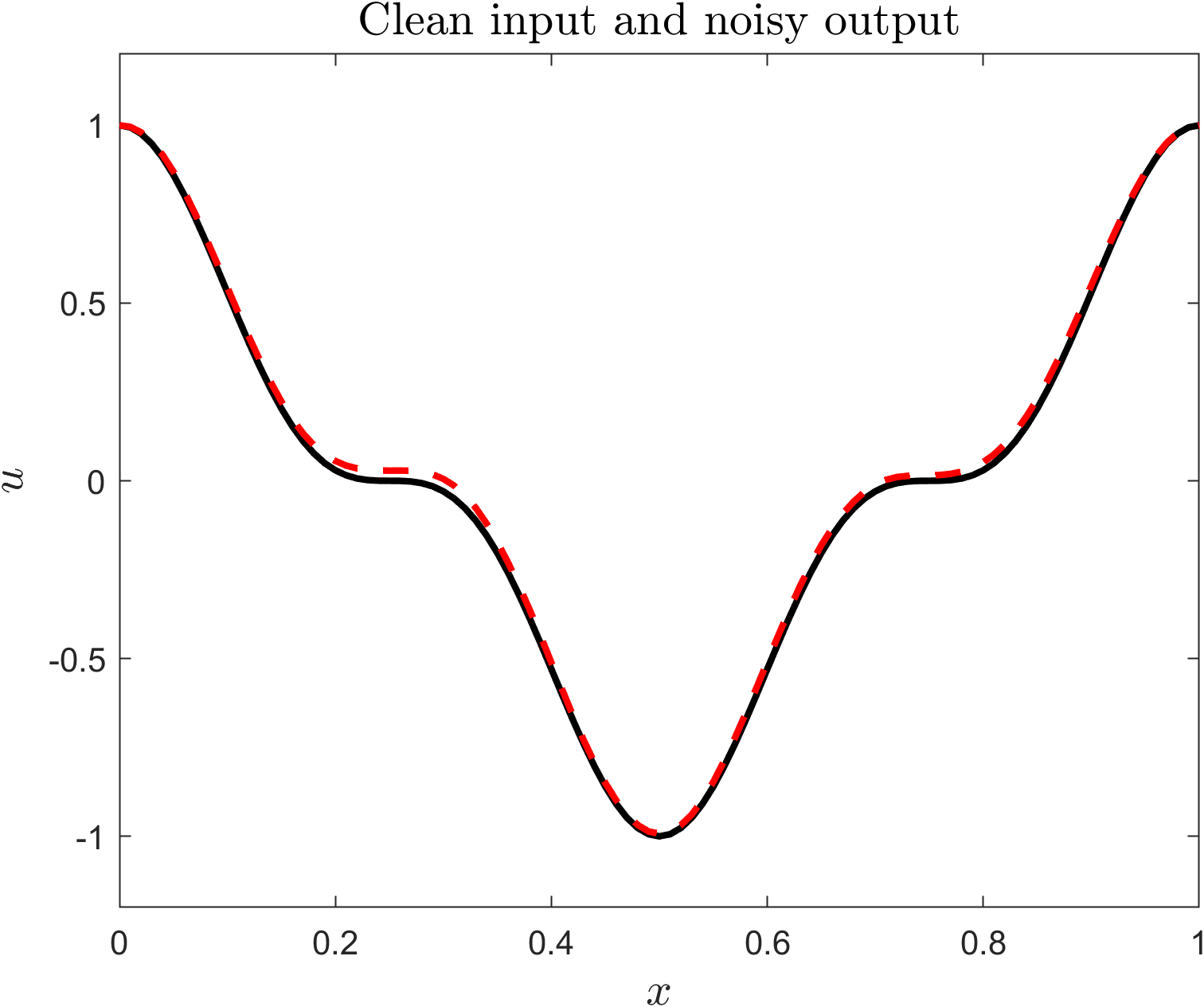}
    \includegraphics[scale=.4]{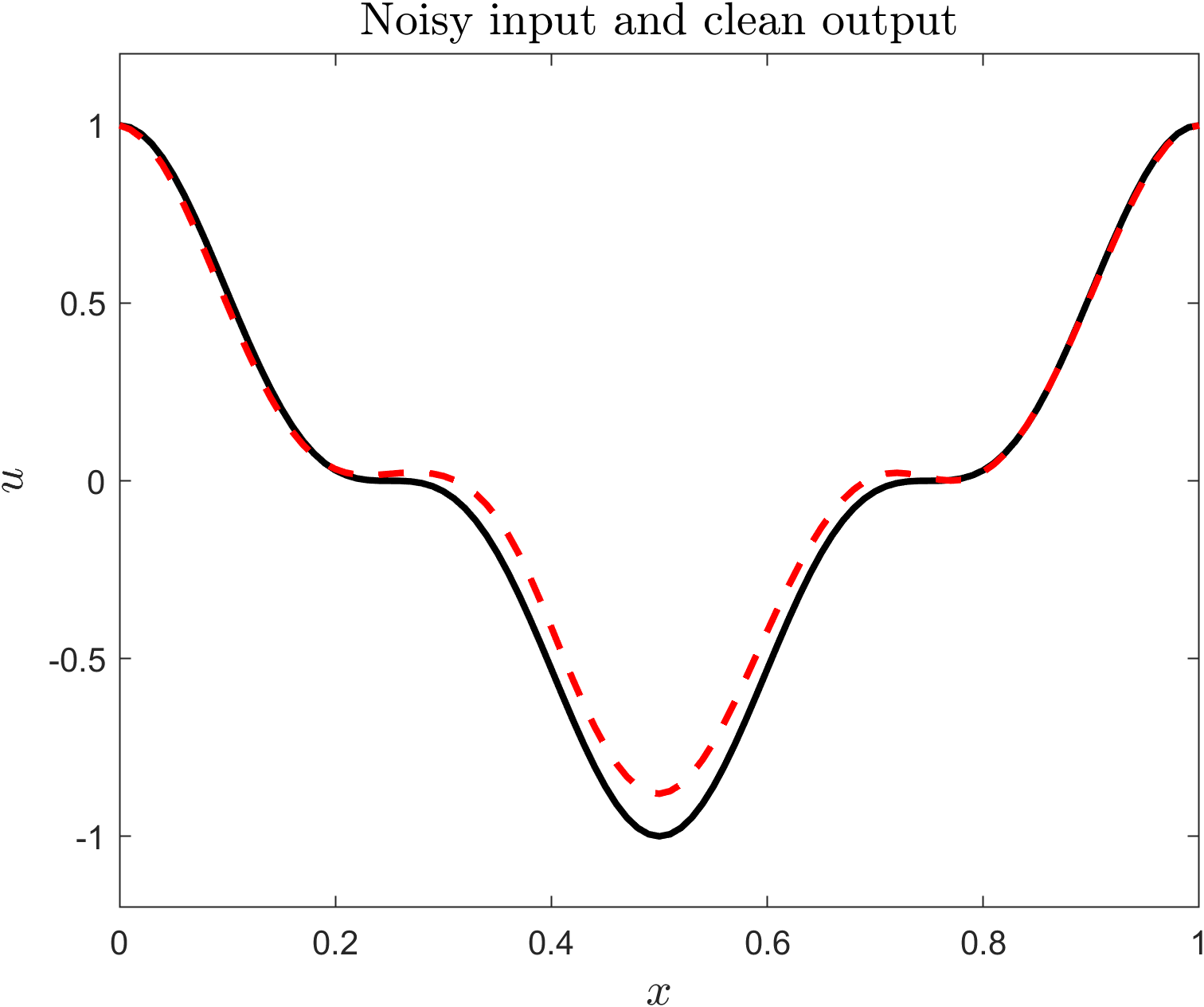}
    }
    \caption{The effect of noisy inputs-outputs in solving \eqref{eq:poisson} using PINNs. Three scenarios are considered: clean data, noisy output but clean input data, and clean output but noisy input data of $f$. The noises for input and output data are additive Gaussian noises with the same standard deviation $0.01$. Comparison shows that noise in the input data has more significant effect upon the performance of PINNs than noise in the output data. $L_2$ regularization is employed to prevent overfitting in all two noisy-data scenarios. Quantitative results can be found in \ref{tab:example_1_1}.}
    \label{fig:example_1_1}
\end{figure}

Here we present additional results in Fig.~\ref{fig:example_1_1} for the example in Sec.~\ref{sec:example_1} where we employed PINNs to solve \eqref{eq:poisson} with different types of data. We observe that, even with the same noise level, the noise in the input data has more significant effect upon the performance of PINNs than noise in the output data, indicating the necessity and importance of respecting the input noise in solving PDEs.

\subsection{1D time-dependent reaction-diffusion with DeepONets}

\begin{figure}[h!]
    \centering
    \subfigure[Reconstruction of $f$.]{
        \includegraphics[scale=.3]{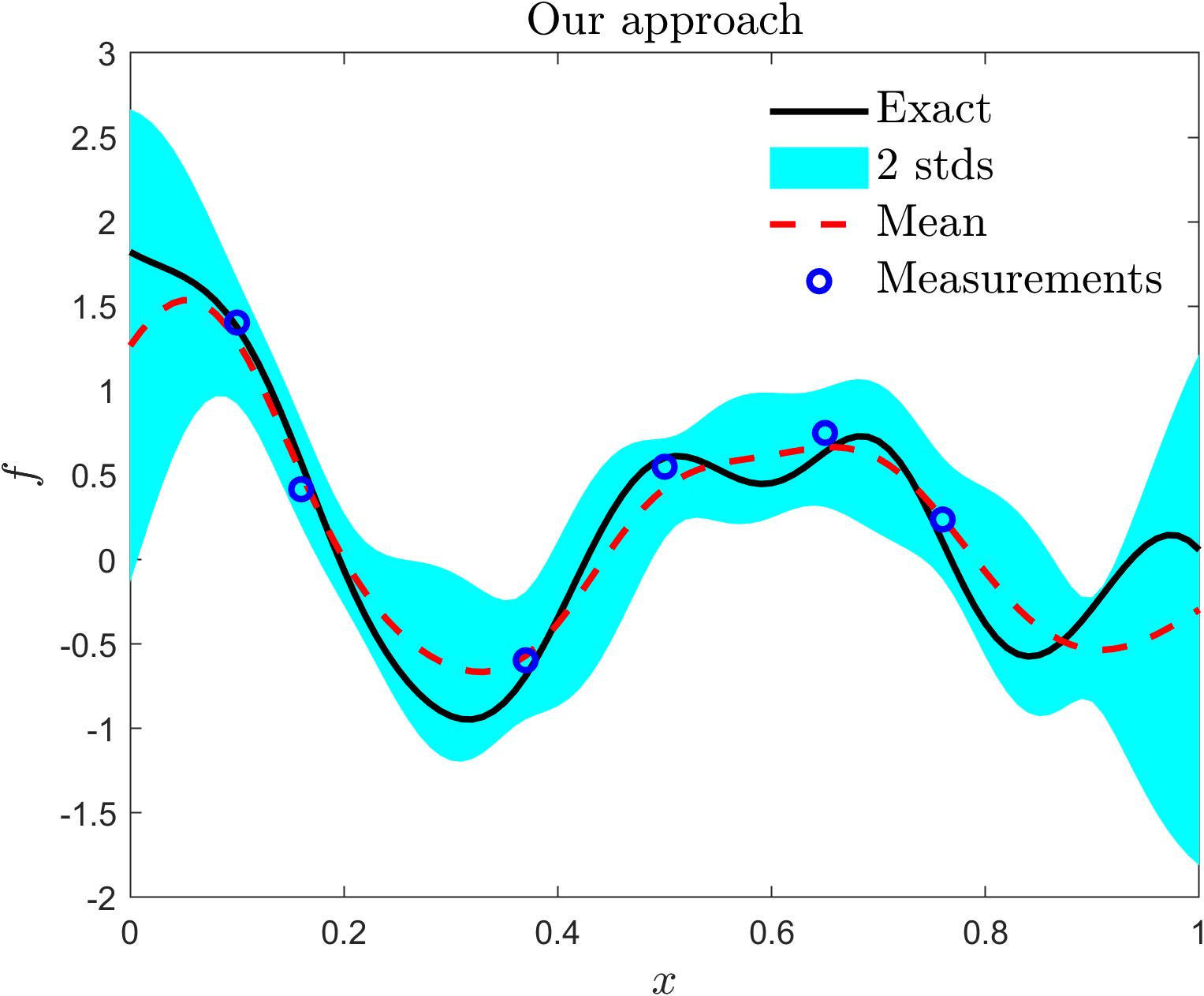}
        \includegraphics[scale=.3]{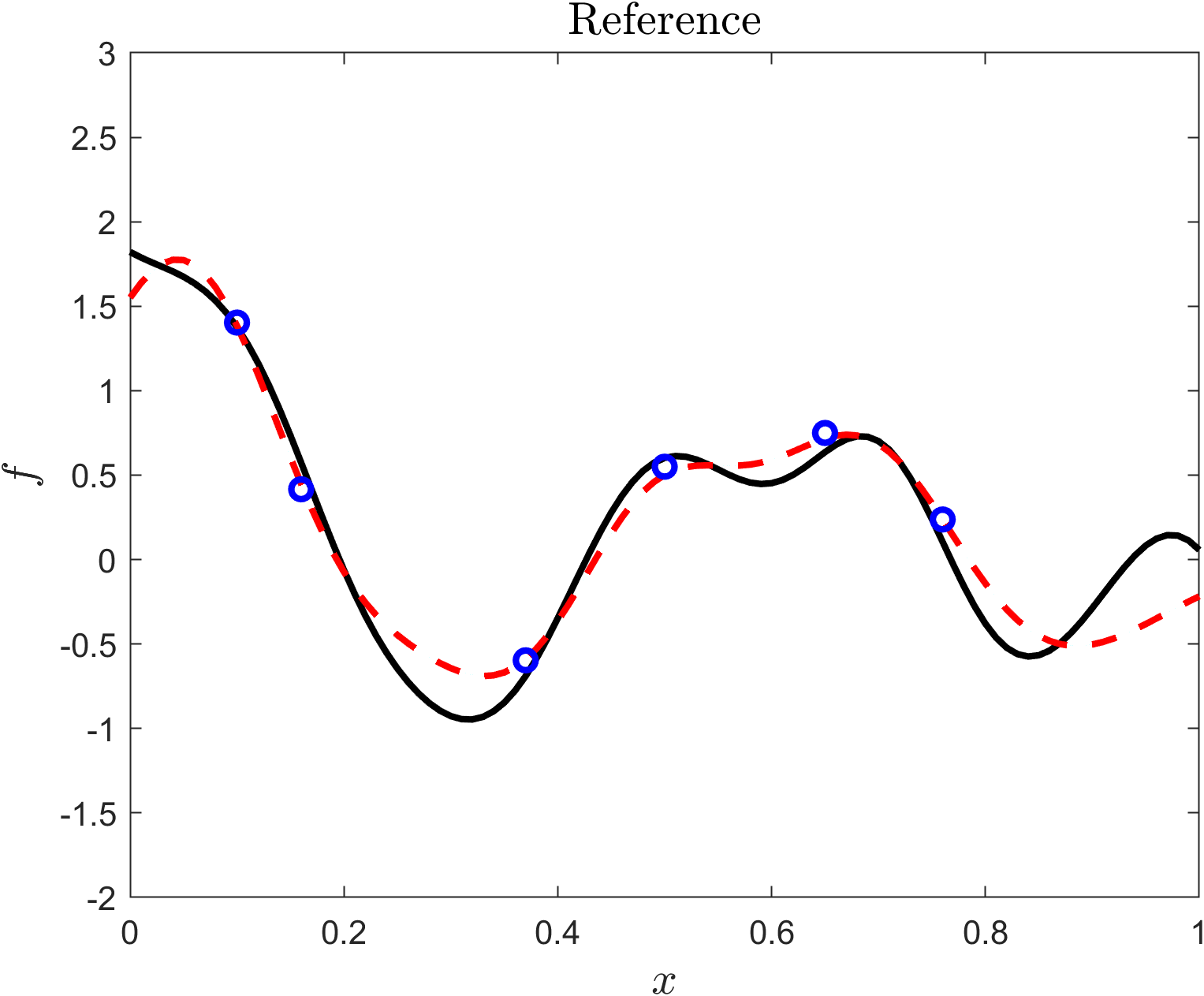}
        \includegraphics[scale=.3]{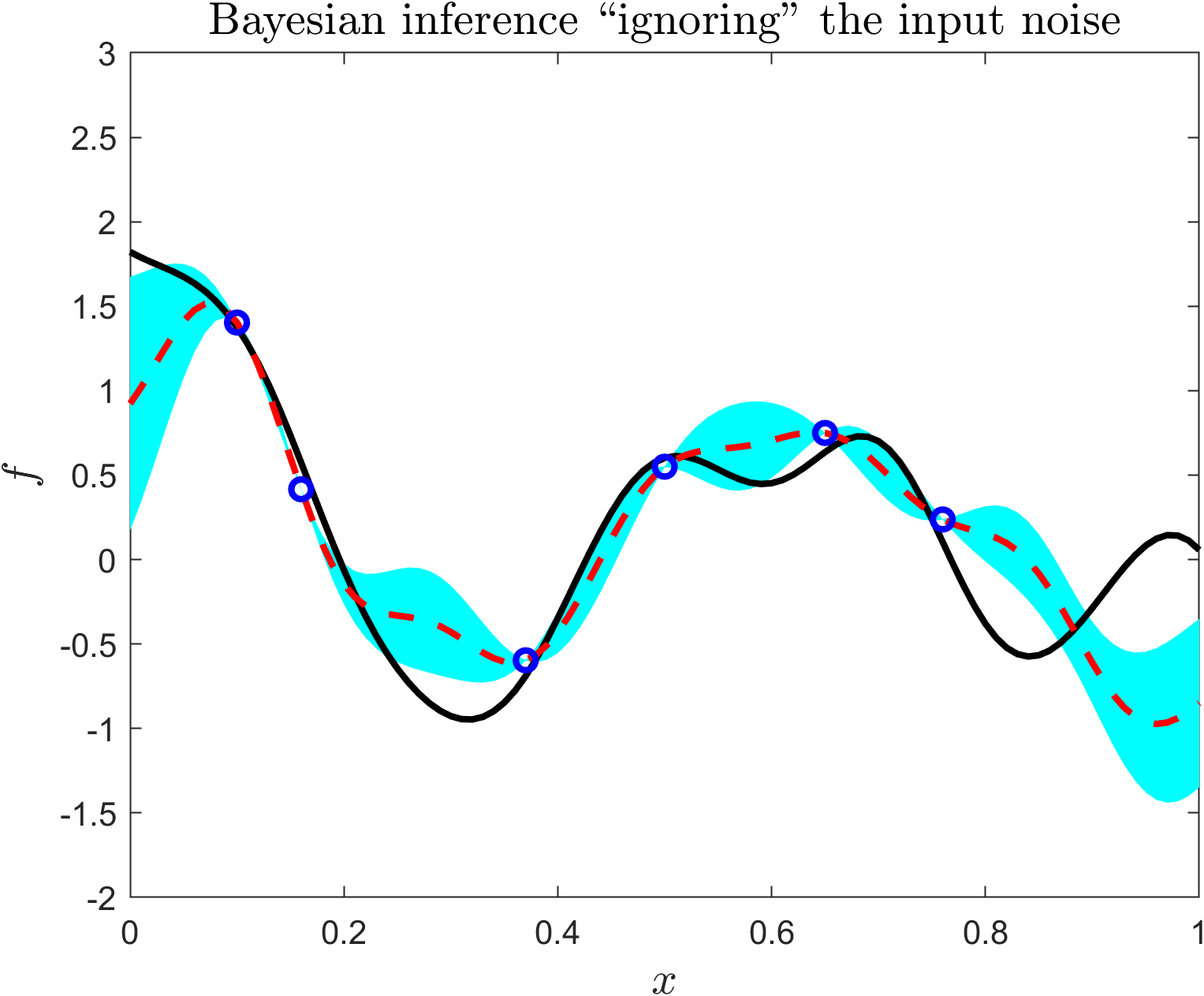}
        \includegraphics[scale=.3]{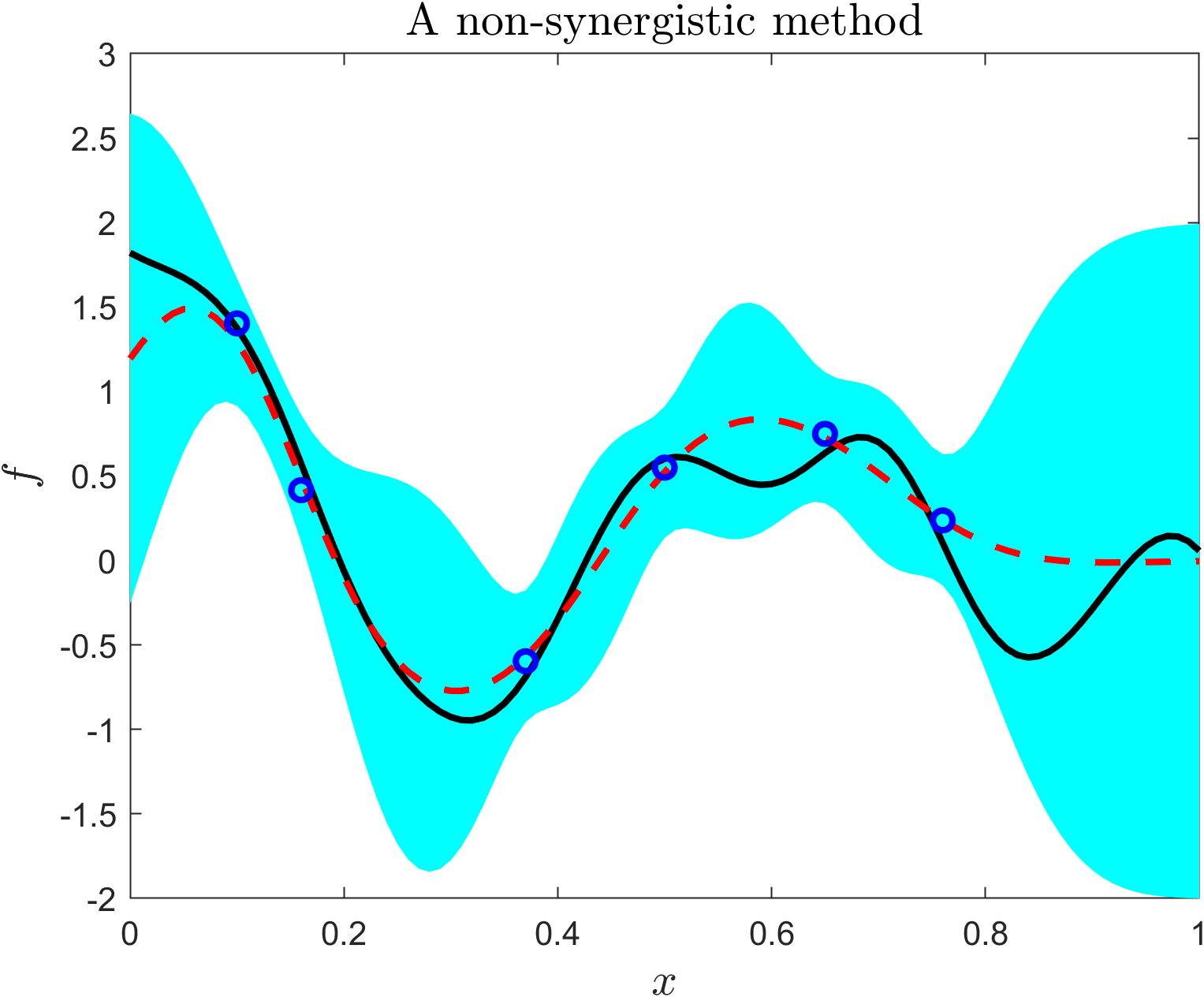}
    }
    \subfigure[Reconstruction of $u|_{t=1}$.]{
        \includegraphics[scale=.3]{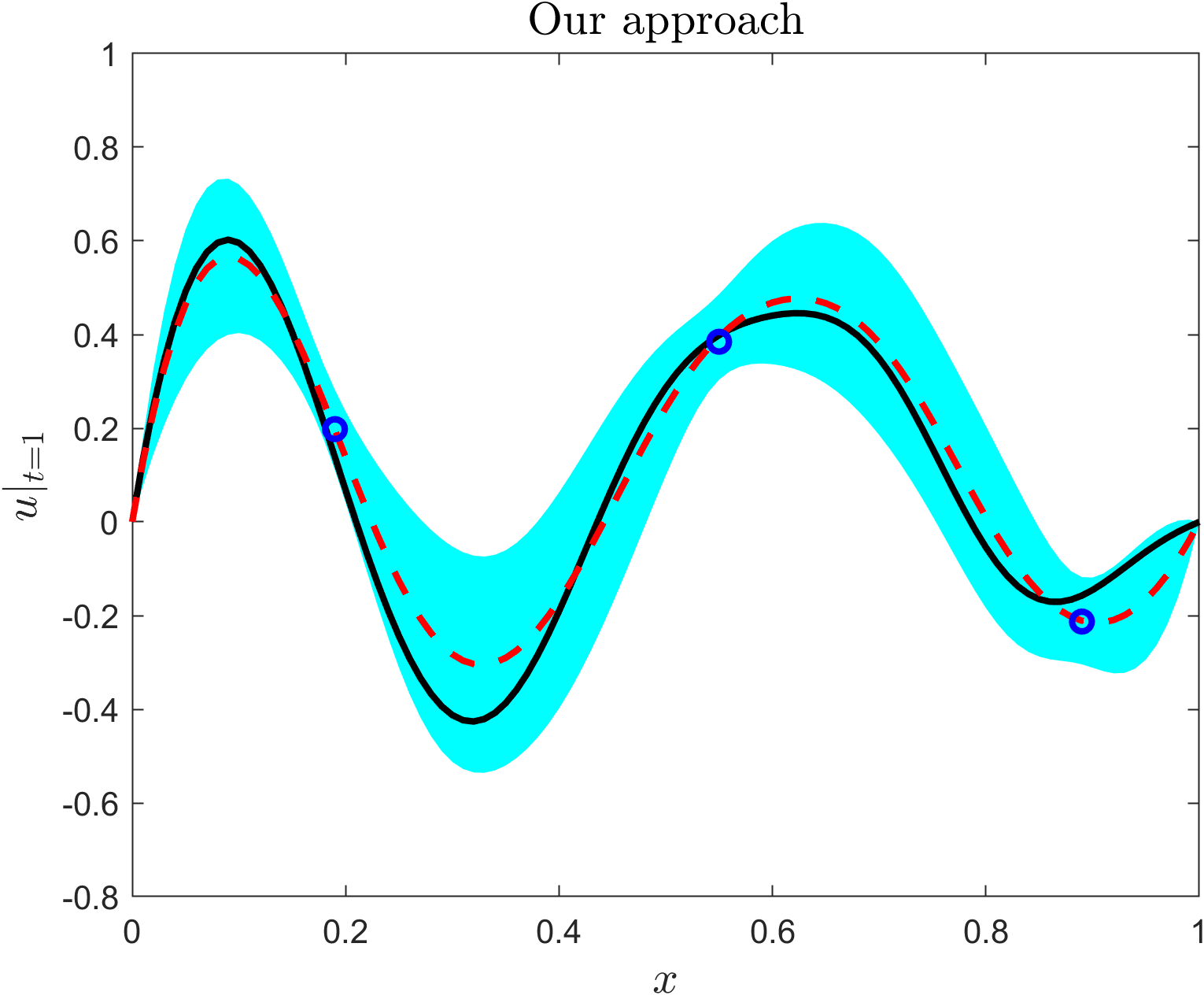}
        \includegraphics[scale=.3]{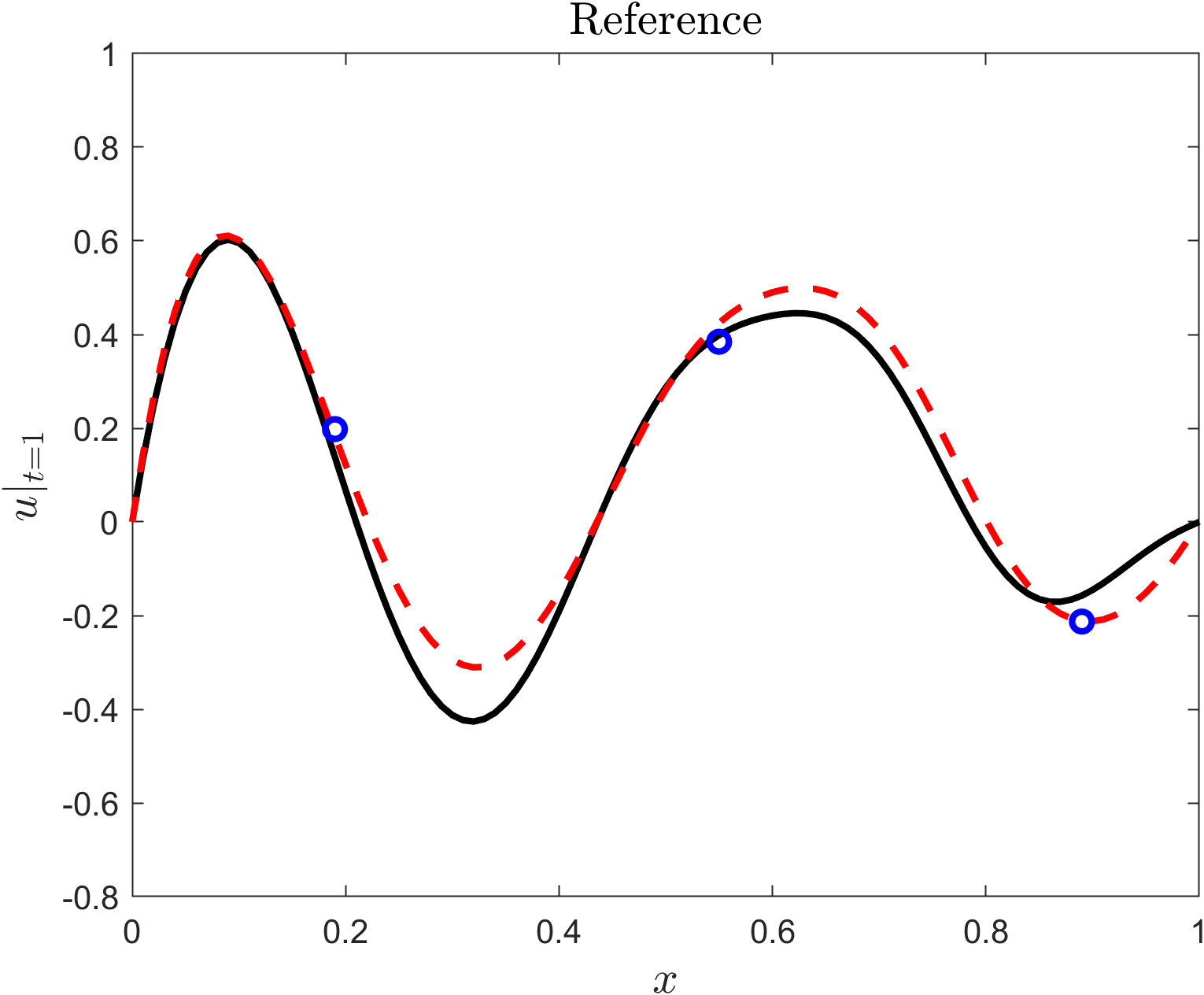}
        \includegraphics[scale=.3]{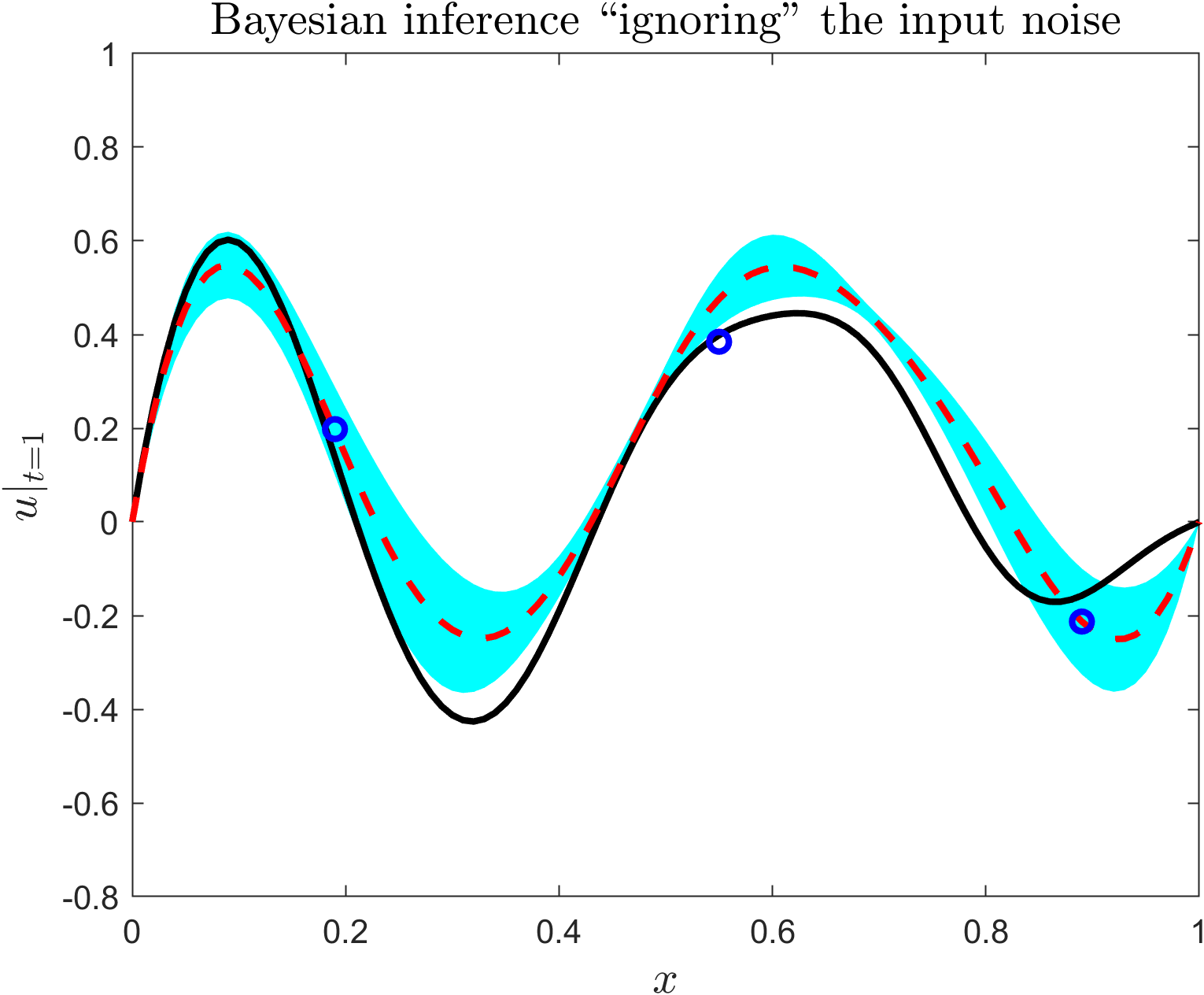}
        \includegraphics[scale=.3]{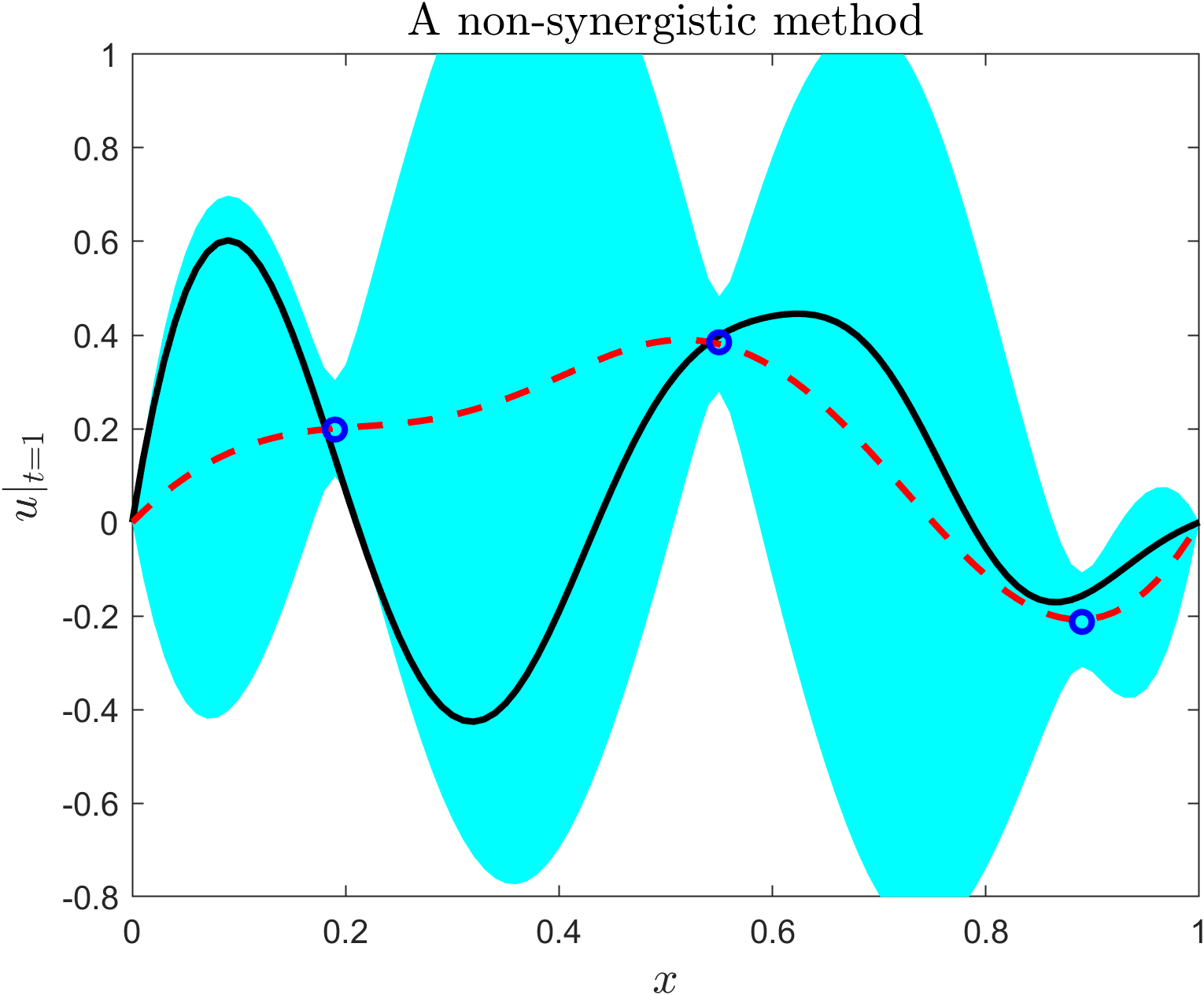}
    }
    \caption{The homogeneous reaction-diffusion equation: Reconstructing $f$ and $u|_{t=1}$ from their sparse and noisy measurements, in which a pretrained DeepONet serves as an equation-free surrogate that encodes the physics. From left to right we present results from our approach with correctly specified input noise, the reference method (a MAP estimate), our approach but misspecifying the scale of the input noise as a relatively small value ($0.01$), and a non-synergistic method.}
    \label{fig:example_3_1}
\end{figure}

In this section we consider a simplified case of Sec.~\ref{sec:example_3} where the diffusion term is a known constant across the field:
\begin{equation}\label{eq:reaction_diffusion2}
    \frac{\partial u}{\partial t} = D \frac{\partial^2 u}{\partial x^2} + \kappa u^2 + f(x), x\in [0, 1], t\in[0, 1],
\end{equation}
where $D=0.01$ and $\kappa=0.01$ and $f$ is the source term. We train a vanilla DeepONet \cite{lu2021learning} with clean and sufficient data to learn the solution operator from $f$ to $u|_{t=1}$, and then use the pretrained NO to reconstruct $f$ and $u|_{t=1}$ from their sparse and noisy measurements. Specifically, we assume six measurements of $f$ and three measurements of $u|_{t=1}$ are available. They are randomly sampled from $x\in[0, 1]$ and corrupted by additive Gaussian noises with scales $0.2$ and $0.05$, respectively.

Results are presented in Fig.~\ref{fig:example_3_1}. We can see that that our approach is able to provide accurate and trustworthy reconstructions of the input and output functions of the NO. Comparing our approach with the MAP estimate, we observe the necessity of UQ in reconstructing functions with pretrained NOs when noisy data are encountered. Similar to results in Sec.~\ref{sec:example_2}, misspecifying/ignoring the input noise yields significantly worse and unreliable predictions in inferring both the input and output functions. We also test the non-synergistic method in this case, and larger predicted uncertainties and errors demonstrate that reconstructions of the input and output functions are performed synergistically with our approach. Besides, as shown in Fig.~\ref{fig:example_3_1}, the predicted uncertainty of $f$ around $x=0.9$ from our approach yields a relatively small value compared to the nearby region even though no measurement of $f$ is available near $x=0.9$. This is caused by the existence of a measurement of $u|_{t=1}$ around $x=0.9$ and therefore showcases the synergistic learning.

\end{document}